\documentclass{article}

\PassOptionsToPackage{sort&compress,numbers}{natbib}
    \usepackage[final]{neurips_2021}

\usepackage[utf8]{inputenc} % allow utf-8 input
\usepackage[T1]{fontenc}    % use 8-bit T1 fonts
\usepackage{hyperref}       % hyperlinks
\usepackage{url}            % simple URL typesetting
\usepackage{booktabs}       % professional-quality tables
\usepackage{amsfonts}       % blackboard math symbols
\usepackage{nicefrac}       % compact symbols for 1/2, etc.
\usepackage{microtype}      % microtypography
\usepackage{xcolor}         % colors

\usepackage{graphicx}
\usepackage{subcaption}
\usepackage{multirow,array,makecell}
\usepackage{float}
\usepackage{wrapfig}
\usepackage{amsthm}
\usepackage{amsmath}
\usepackage{amssymb}
\usepackage[htt]{hyphenat}
\usepackage[linesnumbered,ruled,vlined]{algorithm2e}

\usepackage{color}

\definecolor{darkergreen}{RGB}{21, 152, 56}
\definecolor{red2}{RGB}{252, 54, 65}
\newcommand\redp[1]{\textcolor{red2}{(#1)}}
\newcommand\greenp[1]{\textcolor{darkergreen}{(#1)}}

\definecolor{darkred}{rgb}{0.6, 0.0, 0.0}

\newcommand{\kj}[1]{\textcolor{red}{#1}}
\newcommand{\sh}[1]{\textcolor{orange}{#1}}
\definecolor{ao}{rgb}{0.0, 0.5, 0.0}

\definecolor{pp}{rgb}{0.6,0.0,0.6}

\newcommand\data[1]{{\normalfont \texttt{#1}}}

\newcommand\eg{\textit{e.g.}}
\newcommand\ie{\textit{i.e.}}

\newtheorem{theorem}{Theorem}

\newtheorem{lemma}{Lemma}

\title{SWAD: Domain Generalization\\by Seeking Flat Minima}

\author{%
    Junbum Cha$^{1\dagger}$ \qquad Sanghyuk Chun$^{2}$\thanks{Equal contribution \quad $^\dagger$Part of work done while at NAVER Clova \newline
    \hspace*{0.0cm}Correspondence to: Junbum Cha <junbum.cha@kakaobrain.com>, Sungrae Park <sungrae.park@upstage.ai>
    } \qquad Kyungjae Lee$^3$\footnotemark[1] \\
    \bf Han-Cheol Cho$^4$ \quad Seunghyun Park$^4$ \quad Yunsung Lee$^5$ \quad Sungrae Park$^6$$^\dagger$\\\\
  $^1$ Kakao Brain \quad $^2$ NAVER AI Lab \quad $^3$ Chung-Ang University \\
   $^4$ NAVER Clova \quad $^5$ Korea University \quad $^6$ Upstage AI Research
}

\begin{document}

\maketitle
\begin{abstract}
\vspace{-.5em}
Domain generalization (DG) methods aim to achieve generalizability to an unseen target domain by using only training data from the source domains.
Although a variety of DG methods have been proposed, a recent study shows that under a fair evaluation protocol, called DomainBed, the simple empirical risk minimization (ERM) approach works comparable to or even outperforms previous methods.
Unfortunately, simply solving ERM on a complex, non-convex loss function can easily lead to sub-optimal generalizability by seeking sharp minima.
In this paper, we theoretically show that finding flat minima results in a smaller domain generalization gap.
% In this paper, we find theoretical relationship between flat minima and domain generalization gap.
We also propose a simple yet effective method, named Stochastic Weight Averaging Densely (SWAD), to find flat minima.
SWAD finds flatter minima and suffers less from overfitting than does the vanilla SWA by a dense and overfit-aware stochastic weight sampling strategy.
% By seeking flat minima, 
SWAD shows state-of-the-art performances on five DG benchmarks, namely \data{PACS}, \data{VLCS}, \data{OfficeHome}, \data{TerraIncognita}, and \data{DomainNet}, with consistent and large margins of +1.6\% averagely on out-of-domain accuracy.
We also compare SWAD with conventional generalization methods, such as data augmentation and consistency regularization methods, to verify that the remarkable performance improvements are originated from by seeking flat minima, not from better in-domain generalizability.
Last but not least, SWAD is readily adaptable to existing DG methods without modification; the combination of SWAD and an existing DG method further improves DG performances.
Source code is available at \url{https://github.com/khanrc/swad}.

\end{abstract}

% \newcommand\blfootnote[1]{% footnote without marker
%   \begingroup
%   \renewcommand\thefootnote{}\footnote{#1}%
%   \addtocounter{footnote}{-1}%
%   \endgroup
% }
% \blfootnote{Correspondence to: Junbum Cha <junbum.cha@navercorp.com>, Sungrae Park <sungrae.park@upstage.ai>}

%%%%% Paper Skeleton
\vspace{-.4em}
\section{Introduction}
\vspace{-.25em}
\label{s_introduction}
Independent and identically distributed (i.i.d.) condition is the underlying assumption of machine learning experiments.
However, this assumption may not hold in real-world scenarios, \ie, the training and the test data distribution may differ significantly by \emph{distribution shifts}.
For example, a self-driving car should adapt to adverse weather or day-to-night shifts~\cite{dai2018dark, michaelis2019benchmarking}.
Even in a simple image recognition scenario, systems rely on wrong cues for their prediction, \eg, geographic distribution~\cite{de2019does}, demographic statistics~\cite{yang2020towards}, texture~\cite{geirhos2019cnn_biased_towards_texture}, or backgrounds~\cite{beery2018recognition}.
Consequently, a practical system should require generalizability to distribution shift, which is yet often failed by traditional approaches.

Domain generalization (DG) aims to address \emph{domain shift} simulated by training and evaluating on different domains.
DG tasks assume that both task labels and domain labels are accessible. For example, \data{PACS} dataset~\cite{li2017pacs} has seven task labels (\eg, ``dog'', ``horse'') and four domain labels (\eg, ``photo'', ``sketch'').
Previous approaches explicitly reduced domain gaps in the latent space~\cite{muandet2013icml_DIFL,ganin2016dann,li2018cdann,bahng2019rebias,zhao2020er_entropy_regularization}, obtained well-transferable model parameters by the meta-learning framework~\cite{li2018mldg,dou2019masf,balaji2018metareg,zhang2020arm}, data augmentation~\cite{zhou2021mixstyle,shankar2018crossgrad,carlucci2019jigsaw_jigen}, or capturing causal relation~\cite{arjovsky2019irm,krueger2020vrex}.
Despite numerous previous attempts for a decade, \citet{gulrajani2020domainbed} showed that a simple empirical risk minimization (ERM) approach works comparably or even outperforms the previous attempts on diverse DG benchmarks under a fair evaluation protocol, called ``DomainBed''.

Unfortunately, although ERM showed surprising empirical success on DomainBed, simply minimizing the empirical loss on a complex and non-convex loss landscape is typically not sufficient to arrive at a good generalization~\cite{keskar2016largebatch,garipov2018fge,izmailov2018swa,foret2020sharpness}.
%%%
In particular, the connection between the generalization gap and the flatness of loss landscapes has been actively discussed under the i.i.d. condition~\cite{keskar2016largebatch,dziugaite2017computing,garipov2018fge,izmailov2018swa,jiang2019fantastic,foret2020sharpness}.
\citet{izmailov2018swa} argued that seeking flat minima will lead to robustness against the loss landscape shift between training and test datasets, while a simple ERM converges to the boundary of a wide flat minimum and achieves insufficient generalization.
In the DG scenario, because training and test loss landscapes differ more drastically due to the domain shift, we conjecture that the generalization gap between flat and sharp minima is larger than expected in the i.i.d. scenario.

\begin{table}[t]
\centering
\small
\caption{\small {\bf Comparisons with SOTA.} The proposed SWAD outperforms other state-of-the-art DG methods on five different DG benchmarks with significant gaps (+1.6pp in the average).
% our Our method outperforms every state-of-the-art method for all of the benchmark datasets.
}
\label{table:best_table}
\vspace{0.5em}
% \renewcommand{\arraystretch}{1.1}
% \begin{tabular}{l @{\extracolsep{\fill}} lllll|c} 
% \setlength{\tabcolsep}{3pt}
\begin{tabular}{@{}llllll|c@{}} 
\toprule
                     & \textbf{\data{PACS}} & \textbf{\data{VLCS}} & \textbf{\data{OfficeHome}} & \textbf{\data{TerraInc}} & \textbf{\data{DomainNet}} & \textbf{Avg}.  \\ 
                    %  & \textbf{\data{PACS}~\cite{li2017pacs}} & \textbf{\data{VLCS}~\cite{fang2013vlcs}} & \textbf{\data{OfficeHome}~\cite{venkateswara2017officehome}} & \textbf{\data{TerraInc}~\cite{beery2018terraincognita}} & \textbf{\data{DomainNet}~\cite{peng2019domainnet}} & \textbf{Avg}.  \\ 
\midrule
ERM~\cite{vapnik1998statistical} & 85.5 & 77.5 & 66.5 & 46.1 & 40.9 & 63.3\\
Best SOTA competitor & 86.6~\cite{seo2020dson}          & 78.8~\cite{sun2016coral}          & 68.7~\cite{sun2016coral}                & 48.6~\cite{nam2019sagnet}              & 43.6~\cite{balaji2018metareg,chattopadhyay2020dmg}               & 65.3           \\ 
SWAD (proposed)                 & \textbf{88.1}          & \textbf{79.1}          & \textbf{70.6}                & \textbf{50.0}              & \textbf{46.5}               & \textbf{66.9}           \\
\midrule
Previous SOTA~\cite{sun2016coral} + SWAD                 & 88.3 & 78.9 & 71.3 & 51.0 & 46.8 & 67.3 \\

% CORAL + SWAD & 88.3 & 78.9 & 71.3 & 51.0 & 46.8 & 67.3\\
\bottomrule
\end{tabular}
\vspace{-1em}
\end{table}

To show that flatter minima generalize better to unseen domains, we formulate a robust risk minimization (RRM) problem defined by the worst-case empirical risks within neighborhoods in parameter space~\cite{norton2019diametrical, foret2020sharpness}.
We theoretically show that the generalization gap of DG, \ie, the error on the target domain, is upper bounded by RRM, \ie, a flat optimal solution.
Based on our theoretical observation, we modify stochastic weight averaging (SWA) \citep{izmailov2018swa}, one of the popular existing flatness-aware solvers, by introducing a dense and overfit-aware stochastic weight sampling strategy.
% We propose Stochastic Weight Averaging Densely (SWAD), which is built upon stochastic weight averaging (SWA), one of the popular existing flatness-aware solvers.
% Specifically, SWAD introduces a dense and overfit-aware stochastic weight sampling strategy for SWA.
First, we suggest to sample weights \textbf{\textit{densely}}, \ie, for every iteration.
Also, we search the start and end iterations for averaging by considering the validation loss to \textbf{\textit{avoid overfitting}}.
We empirically show that the proposed Stochastic Weight Averaging Densely (SWAD) finds flatter minima than the vanilla SWA does, resulting in better generalization to unseen domains.

% \textbf{Contribution. }
\textbf{Contribution.}
Our main contribution is introducing flatness into DG, and showing remarkably outperforming performances against existing DG methods. As shown in Table~\ref{table:best_table}, our SWAD improves the average DG performances by 3.6pp against the ERM baseline and 1.6pp against the existing best methods. Furthermore, by combining SWAD and previous SOTA \cite{sun2016coral}, we even achieve 0.4pp improvements against the vanilla SWAD results.
We also empirically show that while popular in-domain generalization methods without considering flatness, \eg, Mixup~\cite{zhang2018mixup} or CutMix~\cite{yun2019cutmix}, are not effective to out-of-domain generalization (Table~\ref{table:generalization}), flatness-aware methods, \eg, SWA~\cite{izmailov2018swa} or SAM~\cite{foret2020sharpness}, are only effective methods to both in-domain and out-of-domain generalization.

\vspace{-.25em}
\section{A Theoretical Relationship between Flatness and Domain Generalization}
\label{s_theoretical_analysis}
\vspace{-.25em}

% \jb{ToDo: less emphasize theoretical results -- I think our current flow is not that problematic, but anyway we need to follow the reivew a little bit at least...}
% \kj{What about change the section title? something like Relationship between Flat Minima and Domain Generalization Gap}

Let $\mathcal{D}:=\left\{\mathcal{D}_{i}\right\}_{i}^{I}$ be a set of training domains, where $\mathcal{D}_{i}$ is a distribution over input space $\mathcal{X}$, and $I$ is the total number of domains. From each domain, we observe $n$ training data points which consist of input $x$ and target label $y$, $(x_{j}^{i},y_{j}^{i})_{j=1}^{n}\sim\mathcal{D}_{i}$.
We also define a set of target domain $\mathcal{T}:=\left\{\mathcal{T}_{i}\right\}_{i}^{T}$ similarly, where the number of target domains $T$ is usually set to one.
For the sake of simplicity, unlike \citet{ben2010theory},
we assume that there exists a global labeling function $h(x)$ that generates target label for multiple domains, \ie, $y_{j}^{i}=h(x_{j}^{i})$ for all $i$ and $j$.
Domain generalization (DG) aims to find a model parameter $\theta \in \Theta$ which generalizes well over both multiple training domains $\mathcal{D}$ and unseen target domain $\mathcal{T}$.
More specifically, let us consider a bounded instance loss function $\ell:\mathcal{Y}\times\mathcal{Y} \rightarrow [0, c]$,
% \footnote{While we assume that $\ell(\cdot,\cdot)$ is bounded by one, it can be generalized for any bounded loss function.}
such that $\ell(y_{1},y_{2})=0$ holds if and only if $y_{1}=y_{2}$ where $\mathcal{Y}$ is a set of labels.
For simplicity, we set $c$ to one in our proofs, but we note that $\ell(\cdot,\cdot)$ can be generalized for any bounded loss function.
Then, we can define a population loss over multiple domains by $\mathcal{E}_{\mathcal{D}}(\theta)=\frac{1}{I}\sum_{i=1}^{I}\mathbb{E}_{x^{i}\sim\mathcal{D}_{i}}[\ell(f(x^{i};\theta), y^{i}))]$, where $f(\cdot;\theta)$ is a model parameterized by $\theta$.
Formally, the goal of DG is to find a model which minimizes both $\mathcal{E}_{\mathcal{D}}(\theta)$ and $\mathcal{E}_{\mathcal{T}}(\theta)$ by only minimizing an empirical risk $\hat{\mathcal{E}}_{\mathcal{D}}(\theta):=\frac{1}{In}\sum_{i=1}^{I}\sum_{j=1}^{n}\ell(f(x^{i};\theta), y^{i}))$ over training domains $\mathcal D$.

%In practice, ERM, \ie, $\arg\min_\theta \hat{\mathcal{E}}_{\mathcal{D}}(\theta)$, is non-convex and has multiple local minima whose solution parameters are similar in terms of the training loss $\hat{\mathcal{E}}_{\mathcal{D}}(\theta)$, but significantly different in terms of the generalization performances $\mathcal{E}_{\mathcal{D}}(\theta)$ and $\mathcal{E}_{\mathcal{T}}(\theta)$.
In practice, ERM, \ie, $\arg\min_\theta \hat{\mathcal{E}}_{\mathcal{D}}(\theta)$, can have multiple solutions that provide similar values of the training losses but significantly different generalizability on $\mathcal{E}_{\mathcal{D}}(\theta)$ and $\mathcal{E}_{\mathcal{T}}(\theta)$.
% In practice, $\arg\min_\theta \hat{\mathcal{E}}_{\mathcal{D}}(\theta)$ is a non-convex problem and has multiple local minimum which have similar training loss, $\hat{\mathcal{E}}_{\mathcal{D}}(\theta)$, but whose true loss ar significantly different in terms of the generalization performances, $\mathcal{E}_{\mathcal{D}}(\theta)$ and $\mathcal{E}_{\mathcal{T}}(\theta)$.
Unfortunately, the typical optimization methods, such as SGD and Adam~\cite{kingma2015adam}, often lead sub-optimal generalizability as finding sharp and narrow minima even under the i.i.d. assumption~\cite{keskar2016largebatch,dziugaite2017computing,garipov2018fge,izmailov2018swa,jiang2019fantastic,foret2020sharpness}.
In the DG scenario, the generalization gap between empirical loss and target domain loss becomes even worse
% than i.i.d. assumption 
due to domain shift.
Here, we provide a theoretical interpretation of the relationship between finding a flat minimum and minimizing the domain generalization gap, inspired by previous studies~\cite{keskar2016largebatch,dziugaite2017computing,garipov2018fge,izmailov2018swa,jiang2019fantastic,foret2020sharpness}.

\begin{wrapfigure}{r}{0.41\linewidth}
    \centering
%    \vspace{-1em}
    \includegraphics[width=\linewidth]{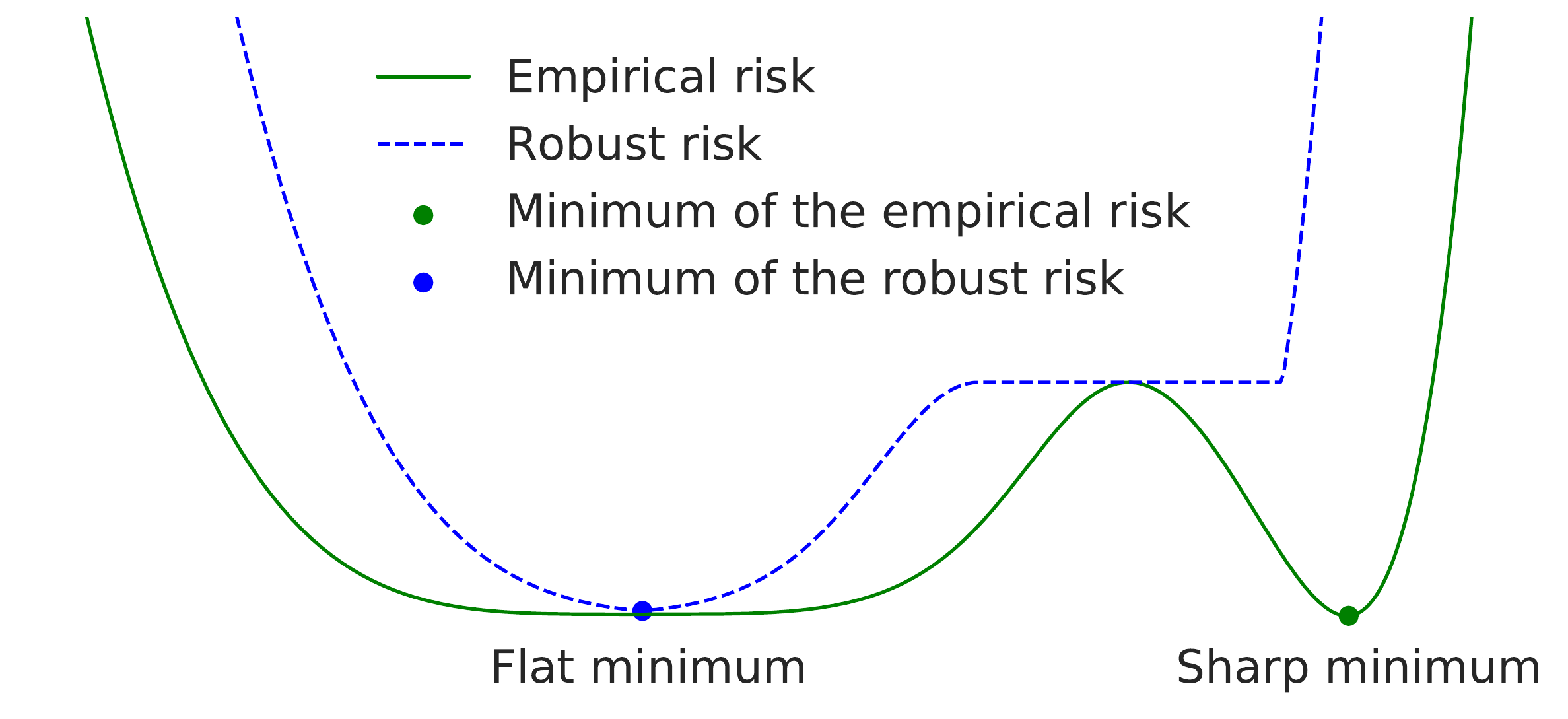}
    \caption{\small {\bf Robust risk minimization (RRM) and flat minima.} With proper $\gamma$, RRM will find flat minima.}
    \label{fig:rrm_and_flatoptima}
    \vspace{-1.8em}
\end{wrapfigure}
We consider a robust empirical loss function defined by the worst-case loss within neighborhoods in the parameter space as $\hat{\mathcal{E}}_{\mathcal{D}}^{\gamma}(\theta):=\max_{\|\Delta\|\leq\gamma} \hat{\mathcal{E}}_{\mathcal{D}}(\theta+\Delta)$, where $\|\cdot\|$ denotes the L2 norm and $\gamma$ is a radius which defines neighborhoods of $\theta$.
Intuitively, if $\gamma$ is sufficiently larger than the ``radius'' of a sharp optimum $\theta_s$ of $\hat{\mathcal{E}}_{\mathcal{D}}(\theta)$, $\theta_s$ is no longer an optimum of $\hat{\mathcal{E}}_{\mathcal{D}}^{\gamma}(\theta)$ as well as its neighborhoods within the $\gamma$-ball.
On the other hand, if an optimum $\theta_f$ has larger ``radius'' than $\gamma$, there exists a local optimum within $\gamma$-ball -- See Figure~\ref{fig:rrm_and_flatoptima}.
Hence, solving the robust risk minimization (RRM), \ie, $\arg\min_\theta \hat{\mathcal{E}}_{\mathcal{D}}^{\gamma}(\theta)$, will find a near solution of a flat optimum showing better generalizability~\cite{norton2019diametrical, foret2020sharpness}.
However, as domain shift worsen the generalization gap by breaking the i.i.d. assumption, it is not trivial that RRM will find an optimum with better DG performance.
% as the i.i.d. scenario.
To answer the question,
% we provide a theorem, showing that finding a flat minimum by RRM, $\hat{\mathcal{E}}_{\mathcal{D}}^{\gamma}(\theta)$, bounds the loss for the unseen target domain $\mathcal T$, $\mathcal{E}_{\mathcal{T}}(\theta)$ (Theorem~\ref{thm:gen_bnd_dg_final}).
we first show the generalization bound between $\hat{\mathcal{E}}^{\gamma}_{\mathcal{D}}$ and $\mathcal{E}_{\mathcal{T}}$ as follows:
\begin{theorem}\label{thm:gen_bnd_dg}
Consider a set of $N$ covers $\{\Theta_{k}\}_{k=1}^{N}$ such that the parameter space $\Theta \subset \cup_{k}^{N} \Theta_{k}$ where $diam(\Theta):=\sup_{\theta,\theta'\in \Theta}\|\theta-\theta'\|_{2}$, $N:=\left\lceil\left(diam(\Theta)/\gamma\right)^{d}\right\rceil$ and $d$ is dimension of $\Theta$.
Let $v_{k}$ be a VC dimension of each $\Theta_{k}$.
Then, for any $\theta\in\Theta$, the following bound holds with probability at least $1-\delta$,
\begin{equation}
\label{eq:thm1_bound}
\mathcal{E}_{\mathcal{T}}(\theta) < \hat{\mathcal{E}}_{\mathcal{D}}^{\gamma}(\theta) +\frac{1}{2I}\sum_{i=1}^{I}\mathbf{Div}(\mathcal{D}_{i},\mathcal{T})+ \max_{k\in[1,N]} \sqrt{\frac{v_{k}\ln\left(m/v_{k}\right)+\ln(N/\delta)}{m}},
\end{equation}
where $m = nI$ is the number of the training samples and $\mathbf{Div}(\mathcal{D}_{i},\mathcal{T}):=2\sup_{A}|\mathbb{P}_{\mathcal{D}_{i}}(A)-\mathbb{P}_{\mathcal{T}}(A)|$ is a divergence between two distributions.
\end{theorem}
Proof can be done similarly as \cite{ben2010theory} and \cite{norton2019diametrical}.
% \kj{Note that proof can be done by using the similar techniques in \cite{ben2010theory} and \cite{norton2019diametrical}.}
In Theorem~\ref{thm:gen_bnd_dg}, the test loss $\mathcal{E}_{\mathcal{T}}(\theta)$ is bounded by three terms: (1) the robust empirical loss $\hat{\mathcal{E}}_{\mathcal{D}}^{\gamma}(\theta)$, (2) the discrepancy between training distribution and test distribution, \ie, the quantity of domain shift, and (3) a confidence bound related to the radius $\gamma$ and the number of the training samples $m$.
Our theorem is similar to \citet{ben2010theory}, while our theorem does not have the term related to the difference in labeling functions across the domains. It is because we simply assume there is no difference between labeling functions for each domain for simplicity. If one assumes a different labeling function, the dissimilarity term can be derived easily because it is independent and compatible with our main proof.
More details of Theorem~\ref{thm:gen_bnd_dg}, including proof and discussions on the confidence bound, are in Appendix C.1 and C.2.
% Here, since increasing $\gamma$ decreases $N$, \ie, grows $|\Theta_k|$, $v_k$, the VC dimension of each cover $\Theta_k$ is increased.
% Note that the relationship between $\gamma$ and the confidence bound is the same as that of \citet{foret2020sharpness}.

From Theorem~\ref{thm:gen_bnd_dg}, one can conjure that minimizing the robust empirical loss is directly related to the generalization performances on the target distribution.
We show that the domain generalization gap on the target domain $\mathcal T$ by the optimal solution of RRM, $\hat{\theta}^{\gamma}$, is upper bounded as follows:
\begin{theorem}\label{thm:gen_bnd_dg_final}
Let $\hat{\theta}^{\gamma}$ denote the optimal solution of the RRM, \ie, $\hat{\theta}^{\gamma}:=\arg\min_{\theta}\hat{\mathcal{E}}^{\gamma}_{\mathcal{D}}(\theta)$, and let $v$ be a VC dimension of the parameter space $\Theta$.
Then, the gap between the optimal test loss, $\min_{\theta'}\mathcal{E}_{\mathcal{T}}\left(\theta'\right)$, and the test loss of $\hat{\theta}^{\gamma}$, $\mathcal{E}_{\mathcal{T}}(\hat{\theta}^{\gamma})$, has the following bound with probability at least $1-\delta$.
\begin{align}
\begin{split}
    \mathcal{E}_{\mathcal{T}}(\hat{\theta}^{\gamma}) - \min_{\theta'}\mathcal{E}_{\mathcal{T}}&\left(\theta'\right) \quad \leq \quad \hat{\mathcal{E}}_{\mathcal{D}}^{\gamma}(\hat{\theta}^{\gamma}) - \min_{\theta''}\hat{\mathcal{E}}_{\mathcal{D}}(\theta'') + \frac{1}{I}\sum_{i=1}^{I}\mathbf{Div}(\mathcal{D}_{i},\mathcal{T})\\
    &+ \max_{k\in[1,N]} \sqrt{\frac{v_{k}\ln\left(m/v_{k}\right)+\ln\left(2N/\delta\right)}{m}} + \sqrt{\frac{v\ln\left(m/v\right) + \ln\left(2/\delta\right)}{m}}
\end{split}
\end{align}
\end{theorem}
Proof is in Appendix C.3. It implies that if we find the optimal solution of the RRM (\ie, $\hat{\theta}^{\gamma}$), then the generalization gap in the test domain (\ie, $\mathcal{E}_{\mathcal{T}}(\hat{\theta}^{\gamma}) - \min_{\theta'}\mathcal{E}_{\mathcal{T}}\left(\theta'\right)$) is upper bounded by the gap between the RRM and ERM (\ie,
$\hat{\mathcal{E}}_{\mathcal{D}}^{\gamma}(\hat{\theta}^{\gamma}) - \min_{\theta''}\hat{\mathcal{E}}_{\mathcal{D}}(\theta'')$).
Other terms in Theorem~\ref{thm:gen_bnd_dg_final} are the discrepancy between the train domains $\mathcal D$ and the target domain $\mathcal T$, and the confidence bounds caused by sample means.
We remark that if we choose a proper $\gamma$, the optimal solution of the RRM will find a point near a flat optimum of ERM as shown in Figure~\ref{fig:rrm_and_flatoptima}.
Hence, Theorem~\ref{thm:gen_bnd_dg_final} and the intuition from Figure~\ref{fig:rrm_and_flatoptima} imply that seeking a flat minimum of ERM will lead to a better domain generalization gap.

\section{SWAD: Domain Generalization by Seeking Flat Minima}
\label{s_method}

We have shown that flat minima will bring a better domain generalization. 
In this section, we 
% \jb{first look at existing flatness-aware solvers,}
propose Stochastic Weight Averaging Densely (SWAD) algorithm, and provide empirical quantitative and qualitative analyses on SWAD and flatness to understand why SWAD works better than ERM.

\subsection{A baseline method: stochastic weight averaging}

Since the importance of flatness in loss landscapes has emerged~\cite{keskar2016largebatch,dziugaite2017computing,garipov2018fge,izmailov2018swa,jiang2019fantastic,foret2020sharpness}, several methods have been proposed to find flat minima~\cite{foret2020sharpness,chaudhari2019entropy,izmailov2018swa}. 
We select stochastic weight averaging (SWA)~\cite{izmailov2018swa} as a baseline, which finds flat minima by a weight ensemble approach.
% We adopt stochastic weight averaging (SWA)~\cite{izmailov2018swa} as a our baseline, which is a widely used flatness-aware solver. It finds flat minima by a simple weight ensemble approach. 
% Stochastic weight averaging (SWA)~\cite{izmailov2018swa}, one of the popular approaches, finds flat minima by a simple weight ensemble approach.
% SWA~\cite{izmailov2018swa} is a simple weight ensemble approach to find flat minima. 
More specifically, SWA updates a pretrained model (namely, a model trained with sufficiently enough training epochs, $K_0$) with a cyclical~\cite{smith2017cyclical} or high constant learning rate scheduling. SWA gathers model parameters for every $K$ epochs during the update and averages them for the model ensemble.
SWA finds an ensembled solution of different local optima found by a sufficiently large learning rate to escape a local minimum. \citet{izmailov2018swa} empirically showed that SWA finds flatter minima than ERM. We also considered sharpness-aware minimization (SAM)~\cite{foret2020sharpness}, which is another popular flatness-aware solver, but SWA finds flatter minima than SAM (See Figure~\ref{fig:flatness}).
We illustrate an overview of SWA in Figure~\ref{fig:swa_overview}.

\subsection{Dense and overfit-aware stochastic weight sampling strategy}
% \vspace{-1em}

\begin{figure}[t]

\centering

% \subcaptionbox{Comparison in loss surface view\label{fig:diff}}{
%     \includegraphics[width=0.27\textwidth]{figures/swa_vs_swad.pdf}
% (a) The extent of flat region depends on the degree of shaking defined as flat. Thanks to constant learning rate and densely averaging, SWAD catches the center of the wider flat region accurately. Relatively, SWA cannot provide enough robustness to domain shifts because it finds a solution in the narrow flat region.
% }\quad
\begin{subfigure}{0.4\linewidth}
    \includegraphics[width=\linewidth]{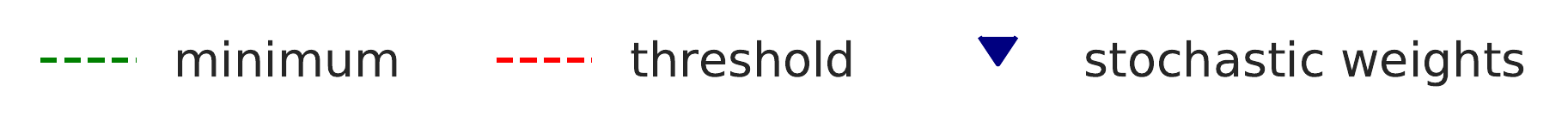}
\end{subfigure}\\
\subcaptionbox{\small SWA\label{fig:swa_overview}}{%
  \includegraphics[width=0.4\textwidth]{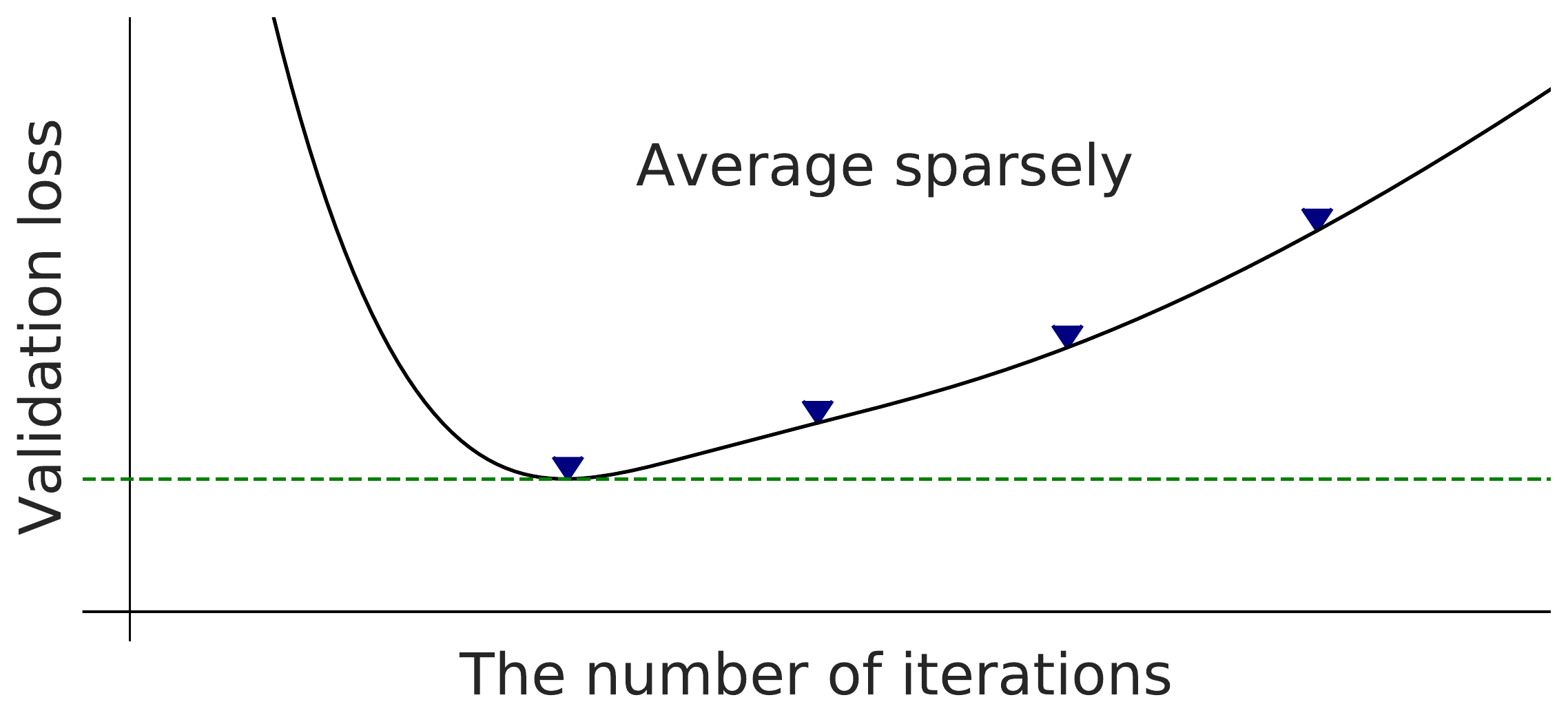}%
}
\qquad \quad
\subcaptionbox{\small SWAD\label{fig:swad_overview} (proposed)}{%
  \includegraphics[width=0.4\textwidth]{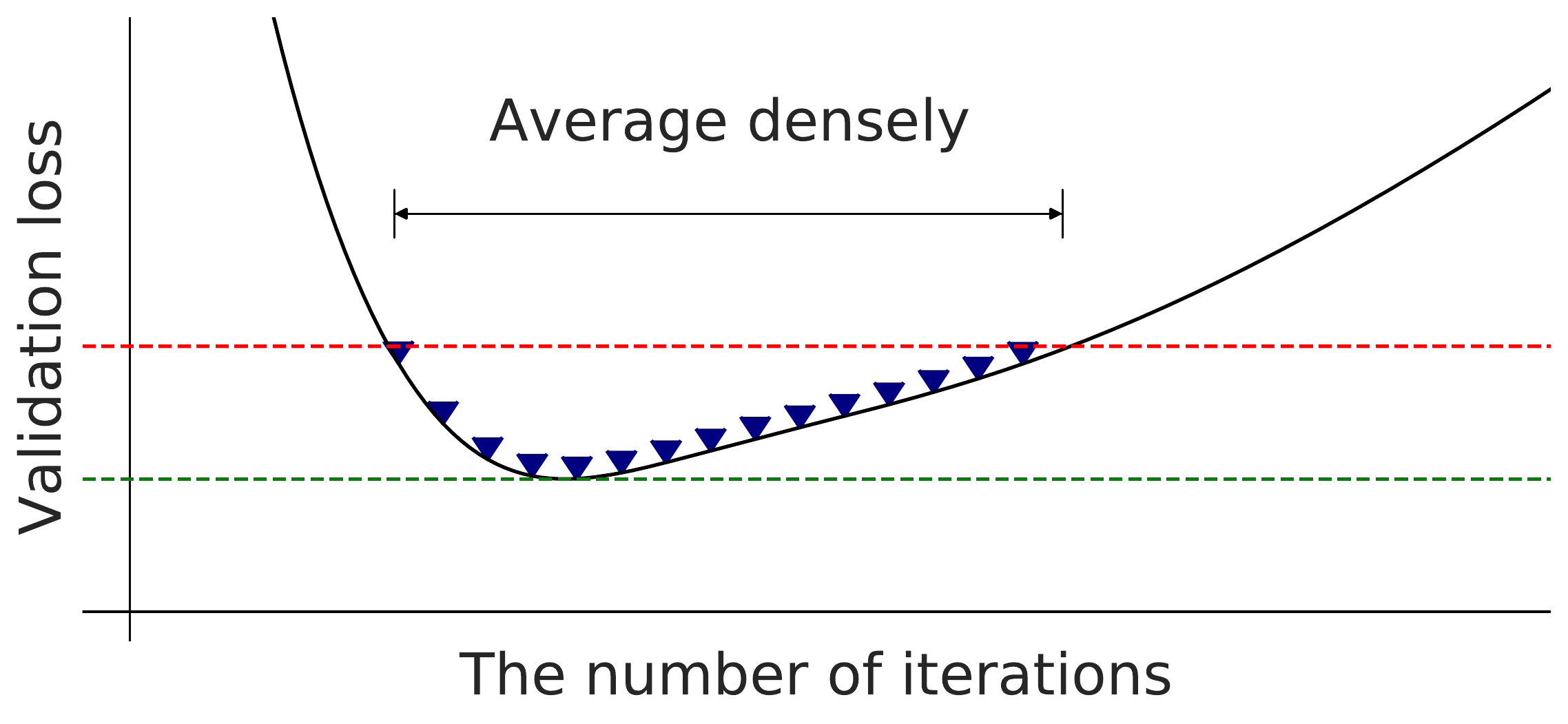}%
}
\vspace{-.5em}
\caption{\small {\bf Comparison between SWA and SWAD.} (a) SWA collects stochastic weights for every $K$ epochs from the pre-defined $K_0$ epochs to the final epoch.
% in an interval of certain epoch. 
(b) Our SWAD collects stochastic weights \textbf{\textit{densely}}, \ie, for every iteration, to obtain sufficiently many weights. SWAD collects the weights from the start iteration $t_s$ to the end iteration $t_e$, where $t_s$ and $t_e$ are obtained by monitoring the validation loss (\textbf{\textit{overfit-aware}} scheduling).
% By the proposed dense and overfit-aware gathering strategy, SWAD can collect sufficiently many stochastic weights while avoiding overfitting.
% (b) SWAD computes a threshold from the minimum validation loss, chooses a range from the threshold, and averages all of the weight in the range.}
}
\label{fig:swa_swad_comparison}
\vspace{-1.5em}
\end{figure}

Despite its advantages, directly applying SWA to DG task has two problems.
First, SWA averages a few weights (usually less than ten) by sampling weights for every $K$ epochs, results in an inaccurate approximation of flat minima on a high-dimensional parameter space (\eg, 23M for ResNet-50~\cite{he2016_cvpr_resnet}).
% \footnote{\citet{izmailov2018swa} also observed similar phenomenon: averaging 10 weights outperforms 5 weights.}
Furthermore, a common DG benchmark protocol uses relatively small training epochs (\eg,
% DomainBed~\cite{gulrajani2020domainbed} typically trained with less than two epochs, 
\citet{gulrajani2020domainbed} trained with less than two epochs for \data{DomainNet} benchmark), resulting in insufficient stochastic weights for SWA.
% not a sufficient to gather stochastic weights for SWA.
From this motivation, we propose a \textbf{\textit{``dense''}} sampling strategy for gathering sufficiently enough stochastic weights.

In addition, widely used DG datasets, such as \data{PACS} ($\approx$ 10K images, 7 classes) and \data{VLCS} ($\approx$ 11K images, 5 classes), are relatively smaller than large-scale datasets, such as ImageNet~\cite{russakovsky2015imagenet} ($\approx$ 1.2M images, 1K classes). In this case, we observe that a simple ERM approach is rapidly reached to a local optimum only within a few epochs, and easily suffers from the overfitting issue, \ie, the validation loss is increased after a few training epochs.
% In addition, since domain generalization benchmarks, such as \data{PACS} ($\approx$ 10K images, 4 domains, 7 classes), \data{VLCS} ($\approx$ 11K images, 4 domains, 5 classes), \data{OfficeHome} ($\approx$ 15K images, 4 domains, 65 classes), and \data{TerraIncognita} ($\approx$ 25K images, 4 domains, 10 classes), are relatively smaller datasets than large-scale datasets, such as ImageNet~\cite{russakovsky2015imagenet} ($\approx$ 1.2M images, 1K classes), we observe that a simple ERM approach is rapidly reached to a local optimum only within a few epochs, and easily suffers from the overfitting issue, \ie, the validation loss is increased after a few training epochs.
It implies that directly applying the vanilla SWA will suffer from the overfitting issue by averaging sub-optimal solutions (\ie, overfitted parameters).
Hence, we need an \textbf{\textit{``overfit-aware''}} sampling scheduling to omit the sub-optimal solutions for SWA.

The main idea of Stochastic Weight Averaging Densely (SWAD) is a dense and overfit-aware stochastic weight gathering strategy.
First, instead of collecting weights for every $K$ epochs, SWAD collects weights for every iteration. This dense sampling strategy easily collects sufficiently many weights than the sparse one.
% (for every $K$ epochs).
We also employ overfit-aware sampling scheduling by considering traces of the validation loss.
Instead of sampling weights from $K_0$ pretraining epochs to the final epoch, we search the start iteration (when the validation loss achieves a local optimum for the first time) and the end iteration (when the validation loss is no longer decreased, but keep increasing).
More specifically, we introduce three parameters: an optimum patient parameter $N_s$, an overfitting patient parameter $N_e$, and the tolerance rate $r$ for searching the start iteration $t_s$ and the end iteration $t_e$.
First, we search $t_s$ which satisfies $\min_{i\in[0, \ldots, N_s-1]} \mathcal E_\text{val}^{(t_s + i)}=\mathcal E_\text{val}^{(t_s)}$, where $\mathcal{E}_\text{val}^{(i)}$ denotes the validation loss at iteration $i$.
Simply, $t_s$ is the first iteration where the loss value is no longer decreased during $N_s$ iterations.
Then, we find $t_e$ satisfying $\min_{i \in [0, 1, \ldots, N_e - 1]} \mathcal E_\text{val}^{(t_e + i)} > r \mathcal E_\text{val}^{(t_s)}$.
In other words, $t_e$ is the first iteration where the validation loss values exceed the tolerance $r$ during $N_e$ iterations.

We illustrate the overview of SWAD and the comparison of SWAD to SWA in Figure~\ref{fig:swa_swad_comparison}. Detailed pseudo code is provided in Appendix B.4.
We compare SWAD with other possible SWA strategies in \S\ref{label:s__ablation} and show that our design choice works better for DG tasks.

\subsection{Empirical analysis of SWAD and flatness}
\label{subsection:s__empirical_analysis_of_swad_and_flatness}

\begin{figure}[t]
    \centering
    % \vspace{-0.5em}
    \setlength{\tabcolsep}{0pt}
    \newcommand\figwidth{.24}
    \begin{subfigure}{0.55\linewidth}
        \includegraphics[width=\linewidth]{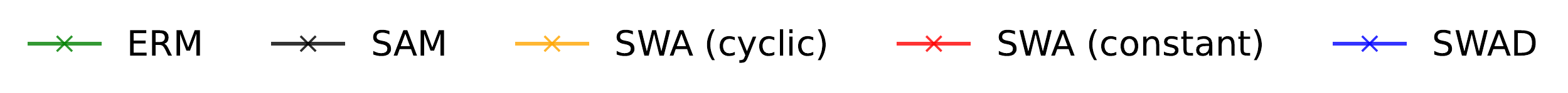}
    \end{subfigure}\\
    %
    % \begin{subfigure}{1.0\linewidth}
        \centering
        \begin{subfigure}{0.3\linewidth}
            \setlength{\abovecaptionskip}{3pt}
            \includegraphics[width=\linewidth]{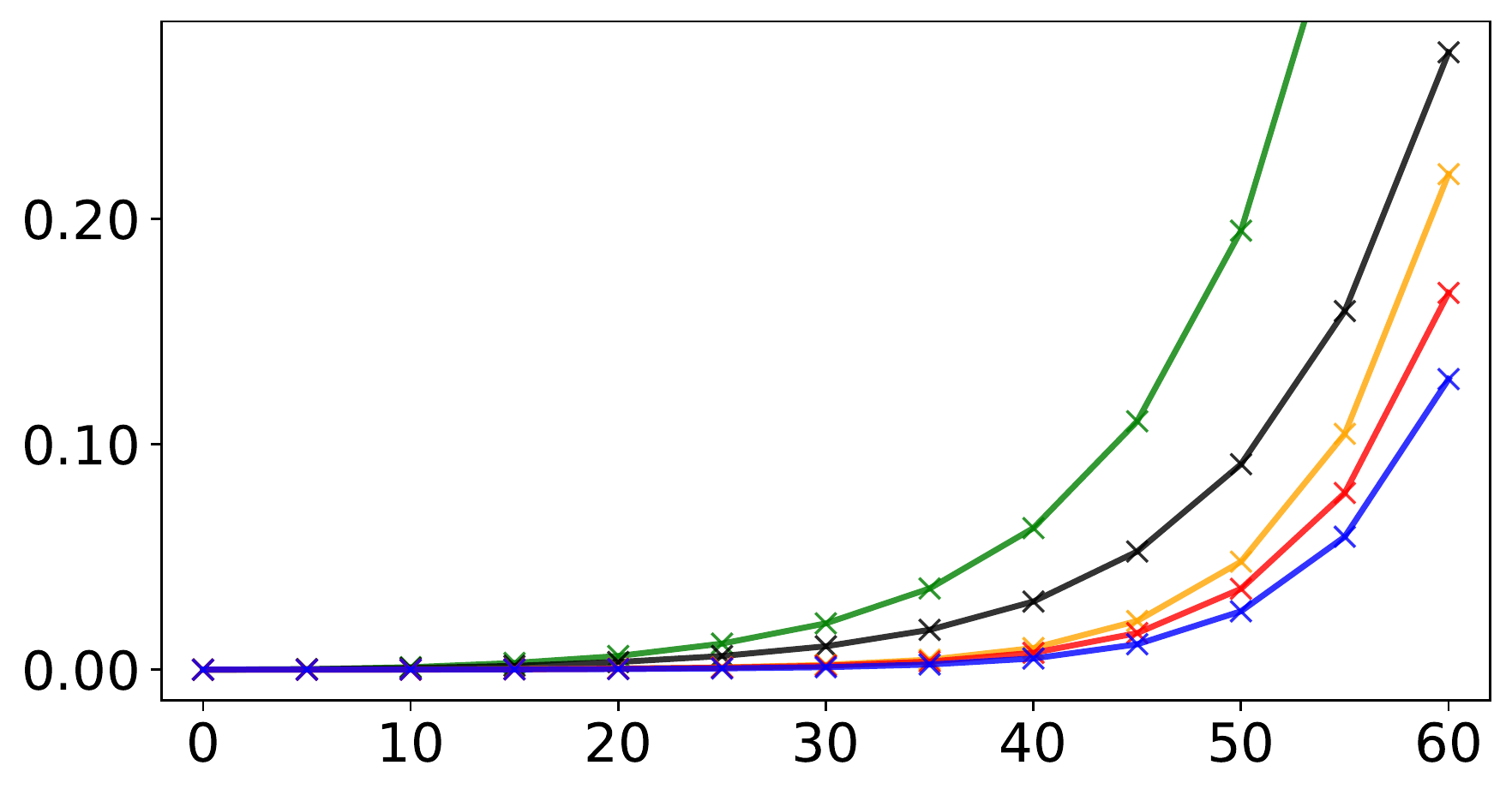}
            \caption{Average train flatness}
        \end{subfigure}
        \quad
        \begin{subfigure}{0.3\linewidth}
            \setlength{\abovecaptionskip}{3pt}
            \includegraphics[width=\linewidth]{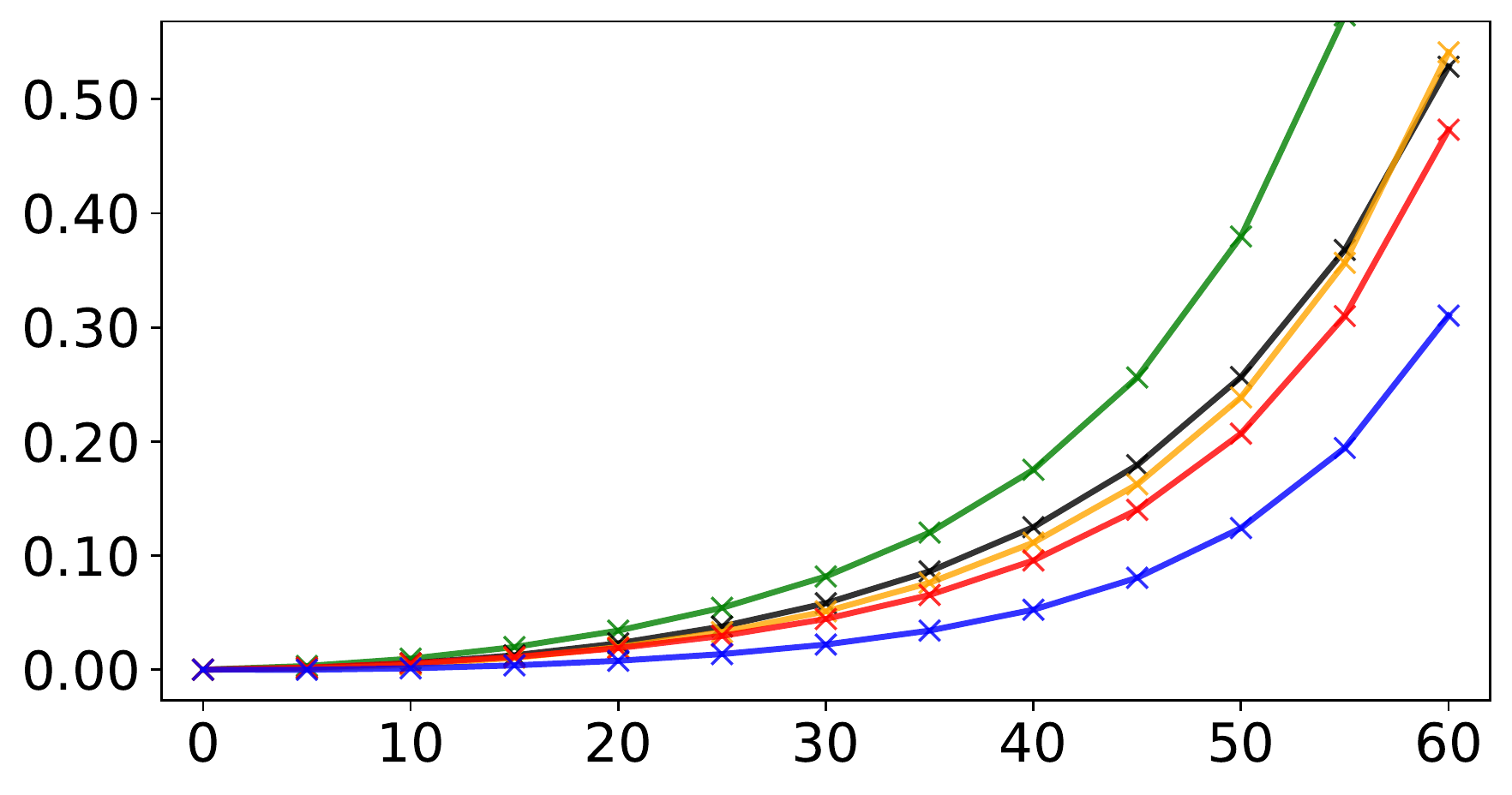}
            \caption{Average test flatness}
        \end{subfigure}
    %     \caption{\small Averaged flatness on \data{PACS} dataset. Left and right figures indicate averaged train and test flatness, respectively.}
    % \end{subfigure}
    \\\vspace{0.4em}
    \begin{subfigure}{1.0\linewidth}
        \begin{tabular*}{\textwidth}{c @{\extracolsep{\fill}} cccc}
            \rotatebox[origin=c]{90}{Train}
            &
            \raisebox{-0.45\height}{\includegraphics[width=\figwidth\textwidth]{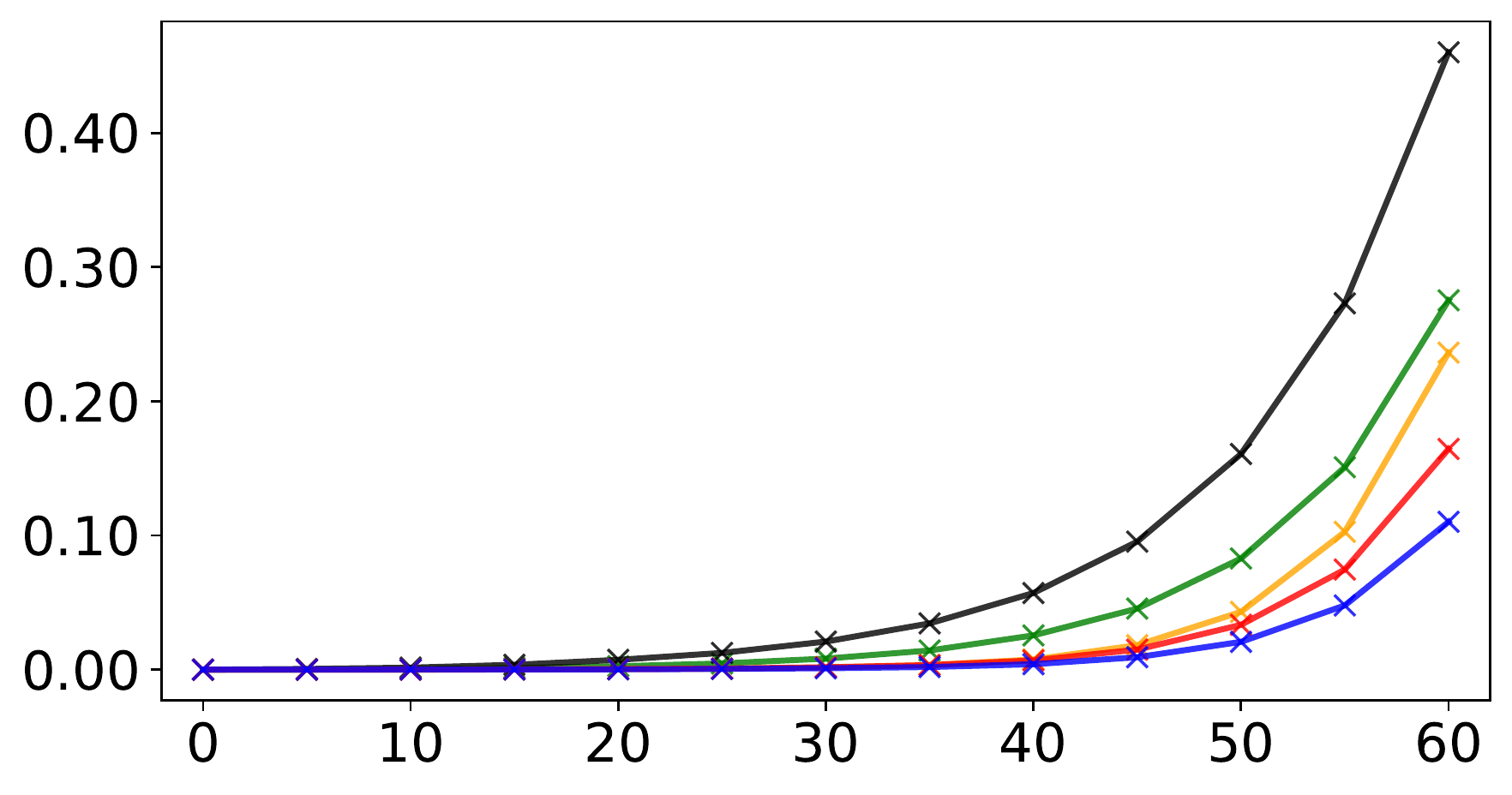}}
            &
            \raisebox{-0.45\height}{\includegraphics[width=\figwidth\textwidth]{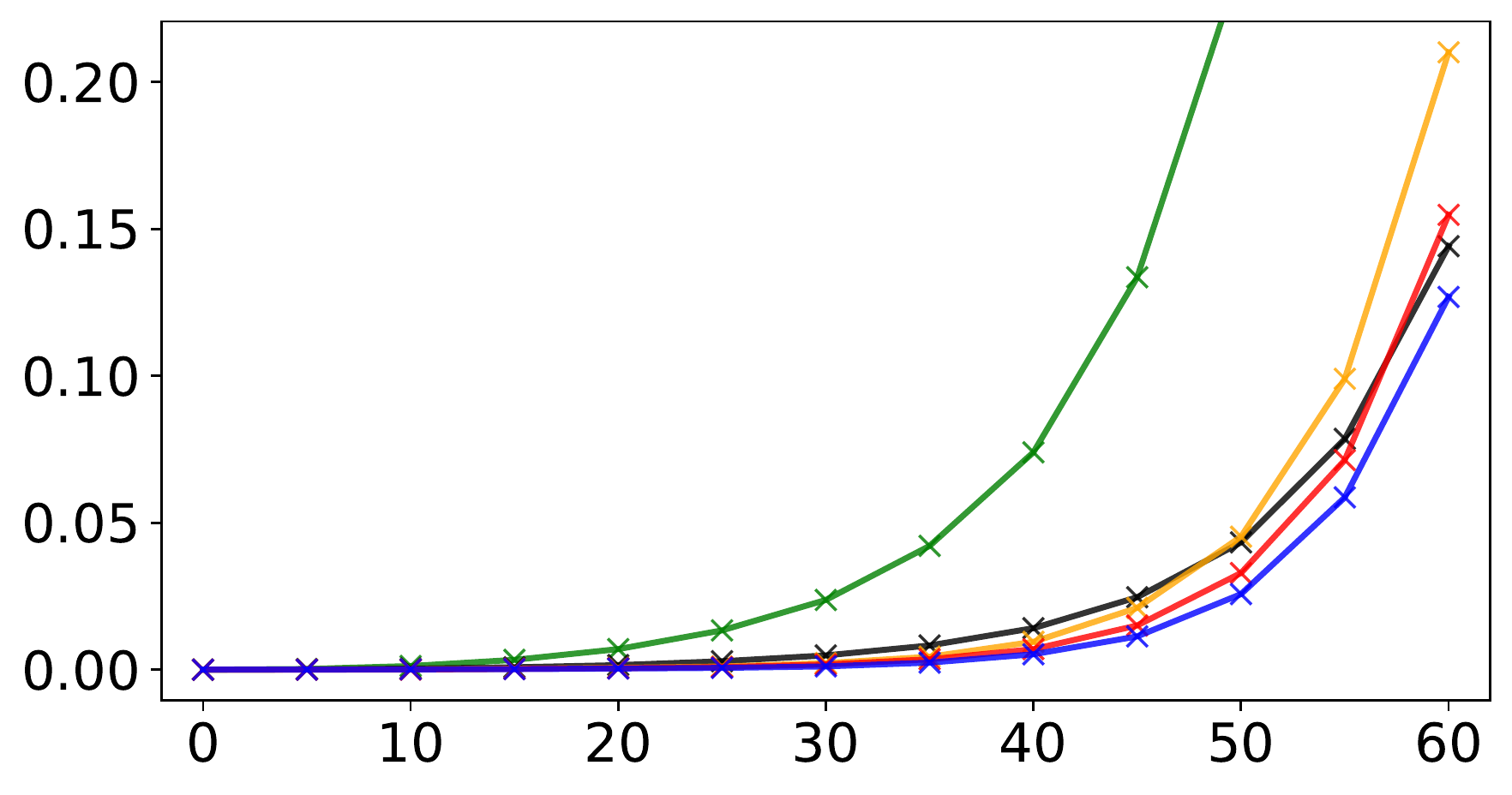}}
            &
            \raisebox{-0.45\height}{\includegraphics[width=\figwidth\textwidth]{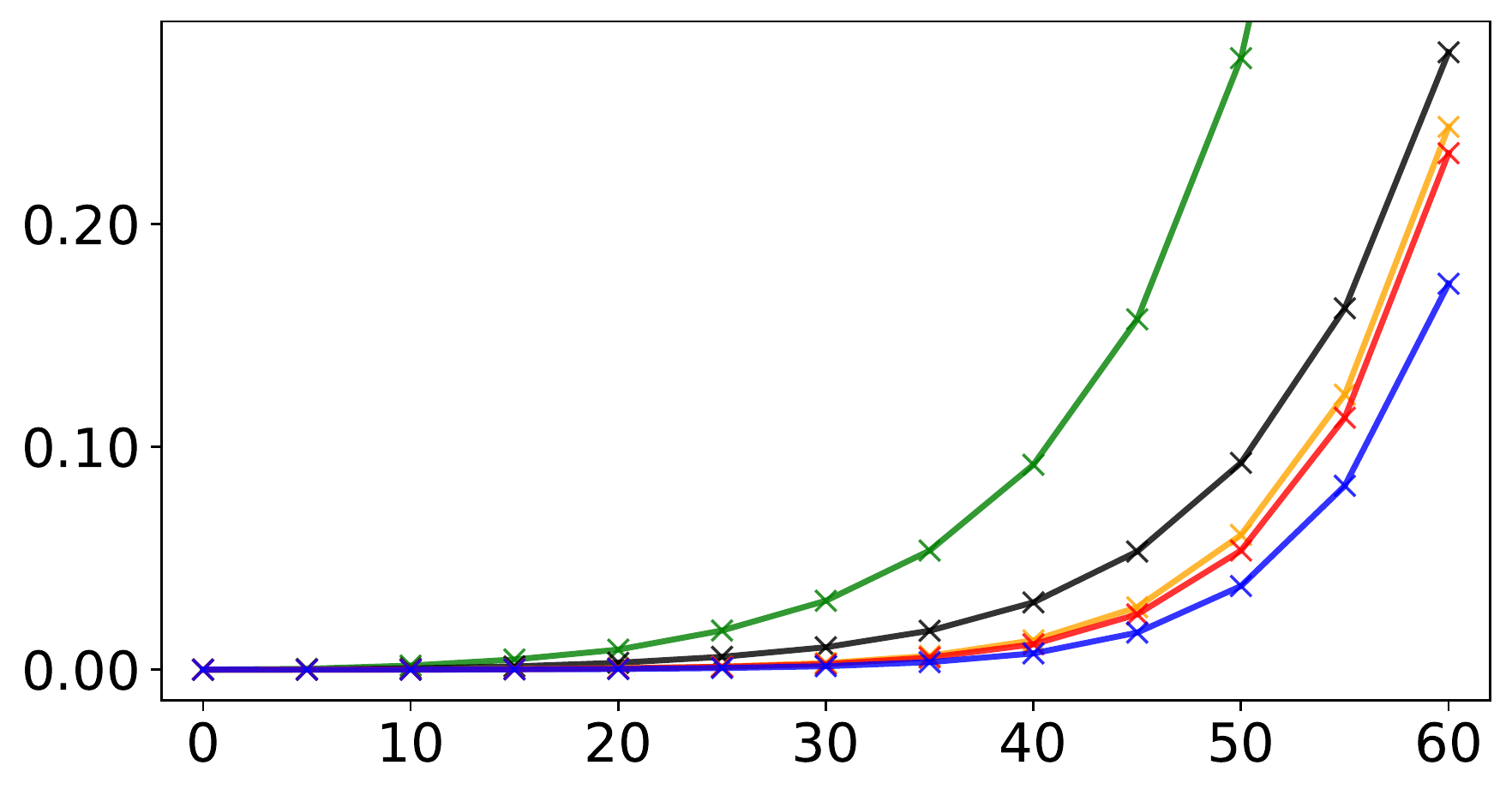}}
            &
            \raisebox{-0.45\height}{\includegraphics[width=\figwidth\textwidth]{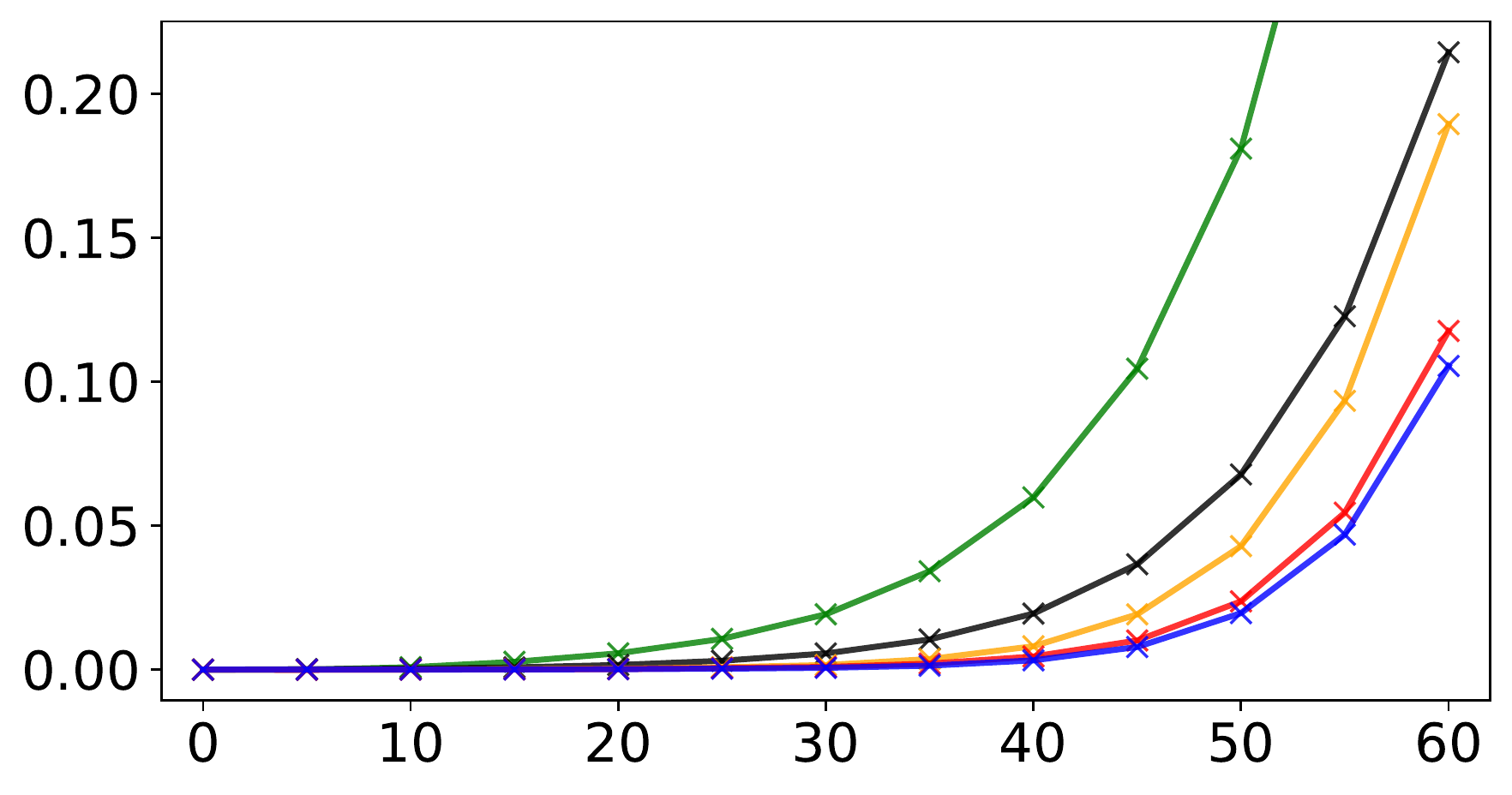}}
        \\
            \rotatebox[origin=c]{90}{Test}
            &
            \raisebox{-0.5\height}{\includegraphics[width=\figwidth\textwidth]{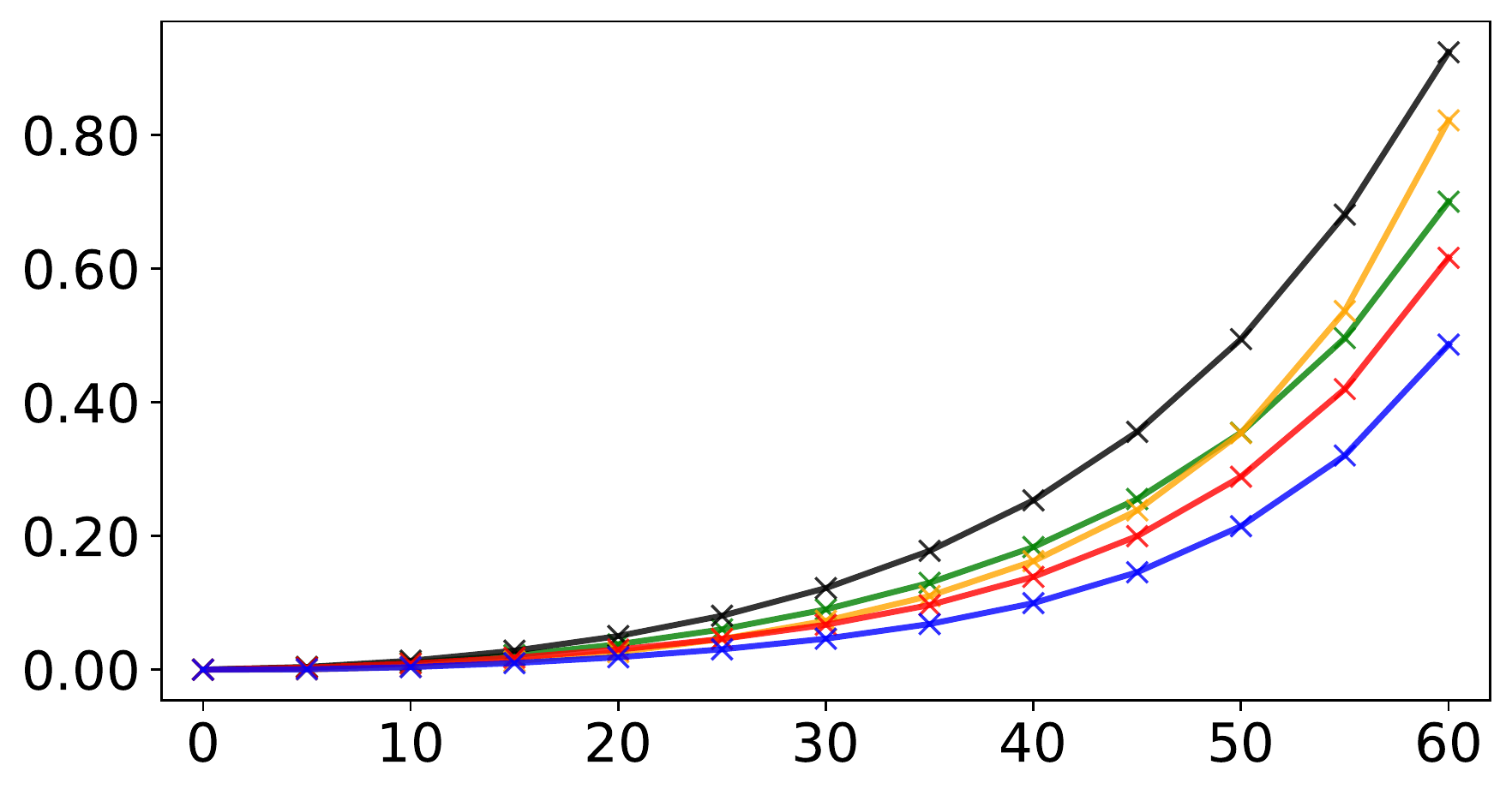}}
            &
            \raisebox{-0.5\height}{\includegraphics[width=\figwidth\textwidth]{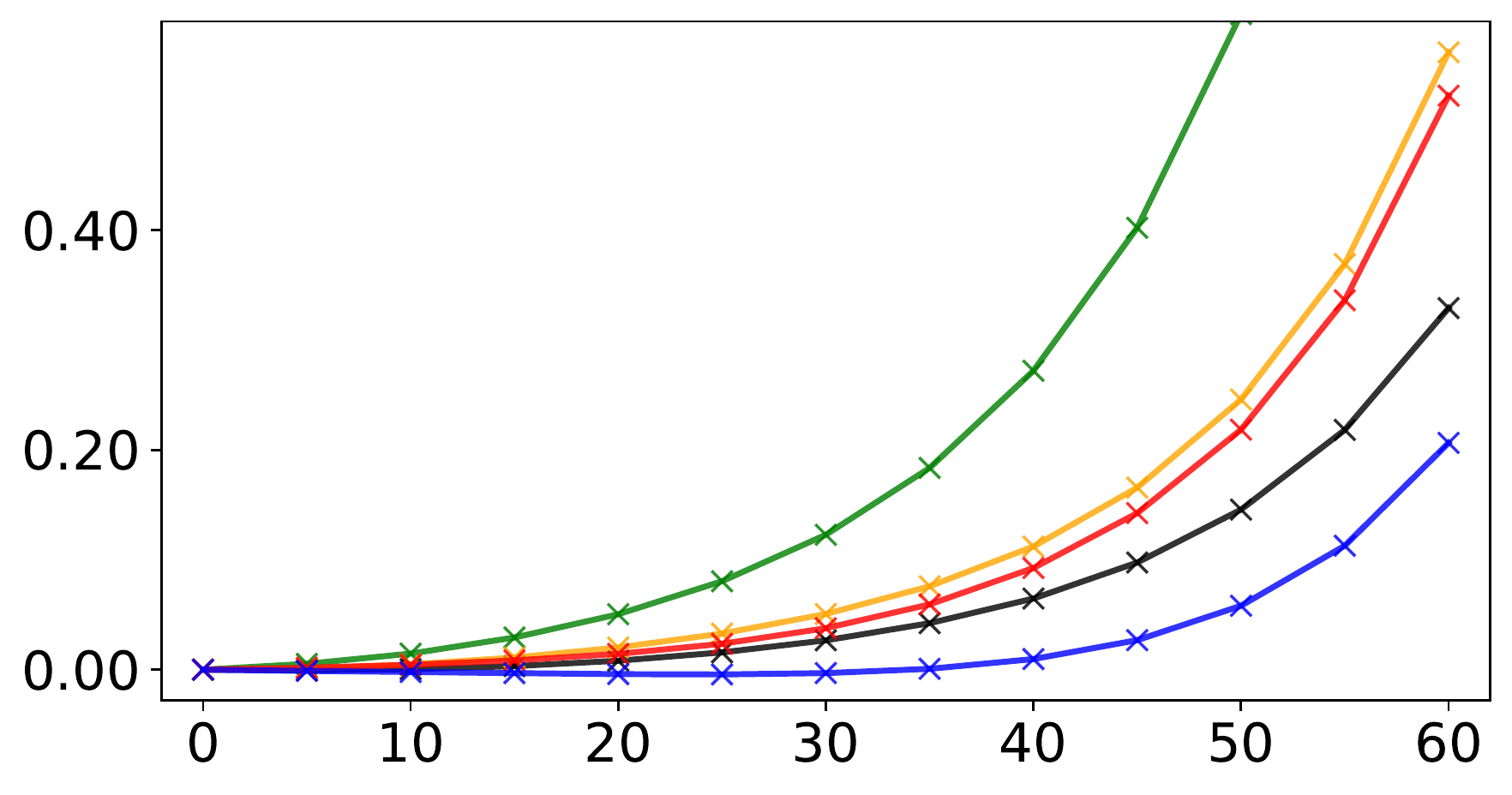}}
            &
            \raisebox{-0.5\height}{\includegraphics[width=\figwidth\textwidth]{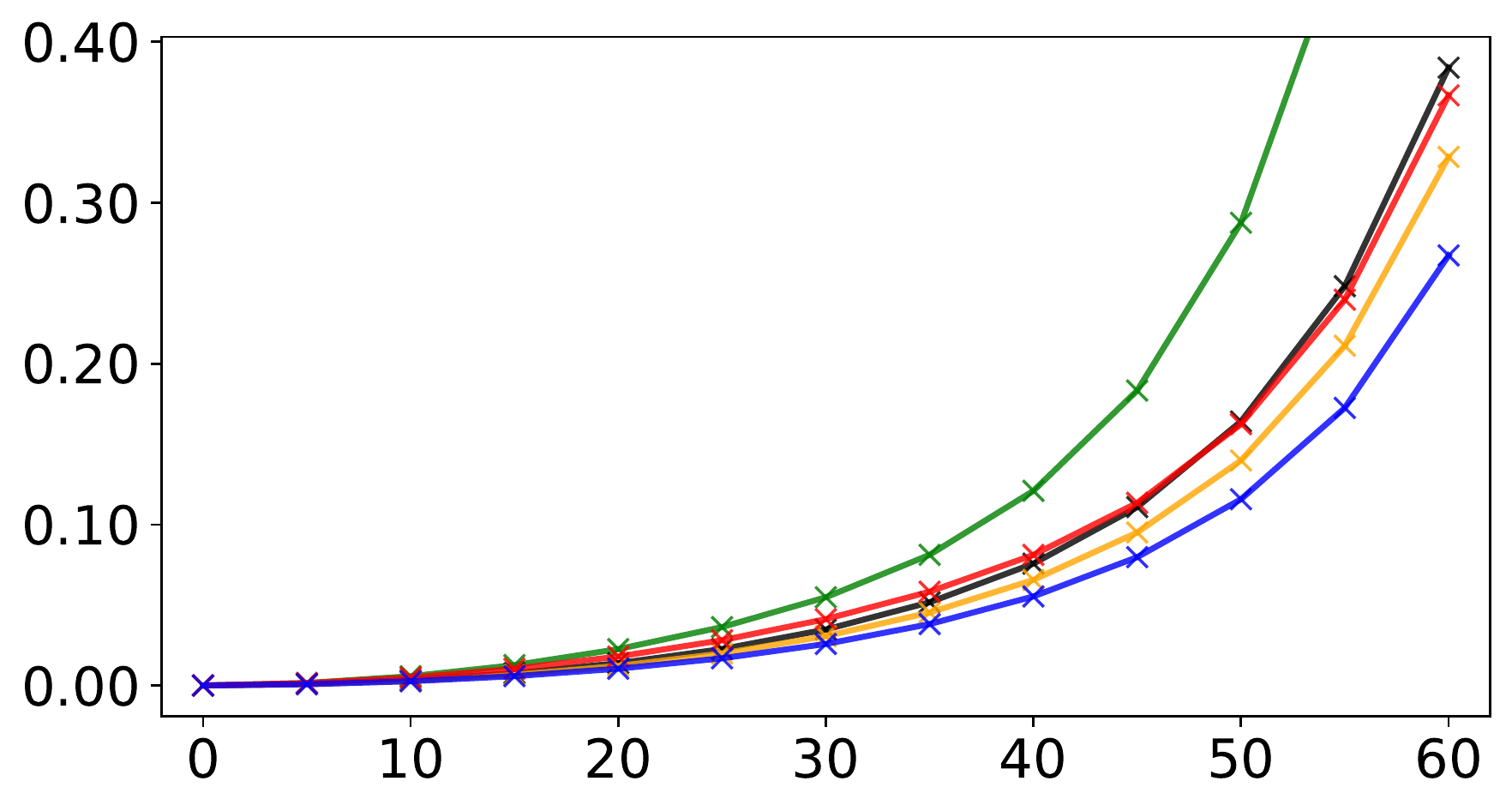}}
            &
            \raisebox{-0.5\height}{\includegraphics[width=\figwidth\textwidth]{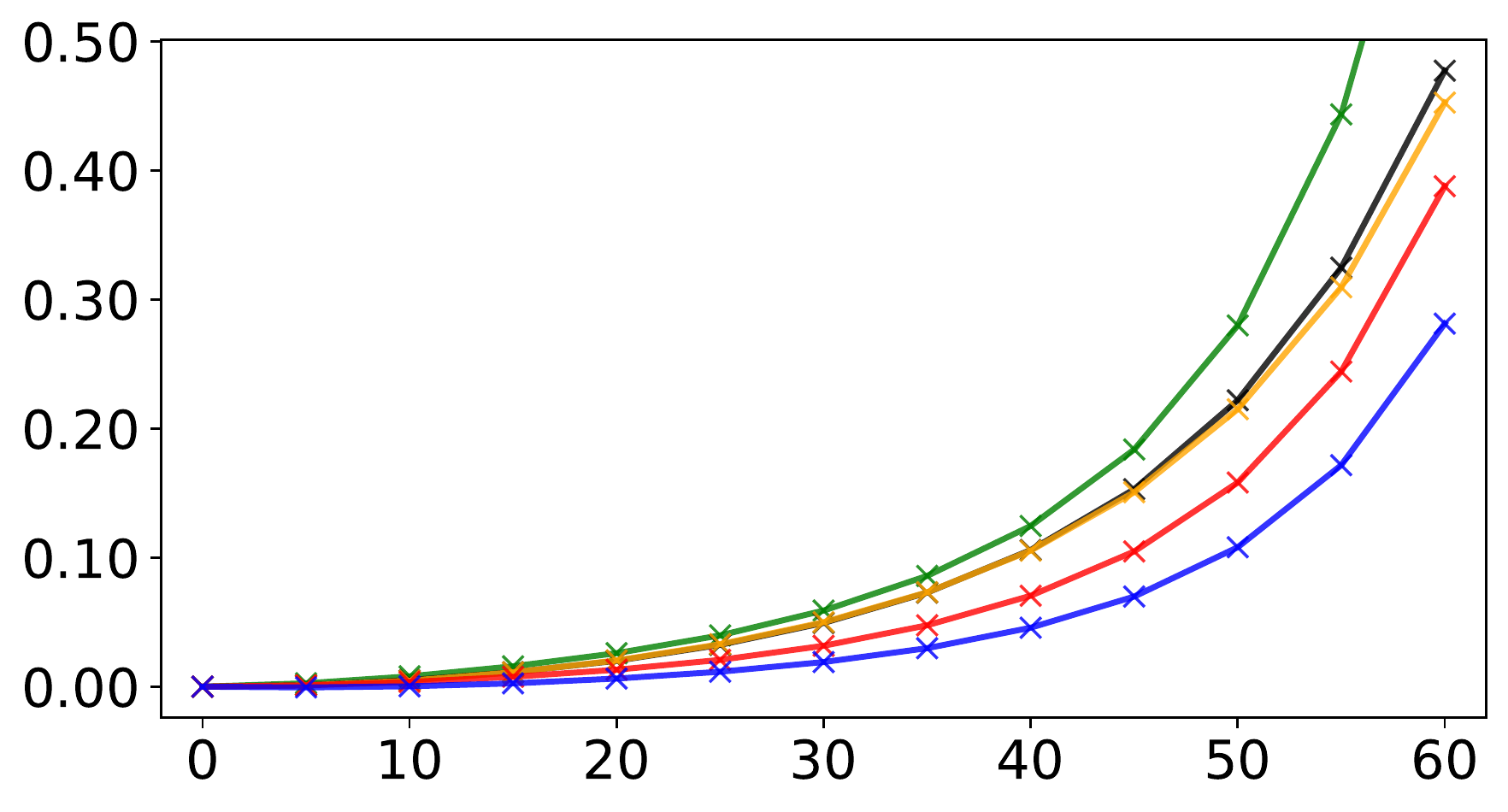}}
        \\
            &
            {\small \textit{Art painting} (test)}
            &
            {\small \textit{Cartoon} (test)}
            &
            {\small \textit{Photo} (test)}
            &
            {\small \textit{Sketch} (test)}
        \end{tabular*}
        \setlength{\abovecaptionskip}{3pt}
        \caption{\small Flatness for each target domain}
    \end{subfigure}
    \caption{\small \textbf{Local flatness comparisons.} We plot the local flatness via loss gap, \ie, $\mathcal{F}_\gamma(\theta)=\mathbb {E}_{\| \theta' \| = \|\theta\| + \gamma} [\mathcal{E}(\theta')-\mathcal{E}(\theta)]$, of ERM, SAM, SWA, and SWAD by varying radius $\gamma$ on different domains of \data{PACS} dataset. For each figure, Y-axis indicates the flatness $\mathcal{F}_\gamma(\theta)$ and X-axis indicates the radius $\gamma$. We measure the train flatness $\mathcal{F_\gamma^D}(\theta)$ on seen domains and the test flatness $\mathcal{F_\gamma^T}(\theta)$ on unseen domain. Each point is computed by Monte-Carlo approximation with 100 random samples. This comparisons show SWAD finds flatter minima than not only ERM but also SAM and SWA.
    %, in both train and test domains.
    }
    \label{fig:flatness}
\vspace{-1em}
\end{figure}

Here, we analyze solutions found by SWAD in terms of flatness. We first verify that the SWAD solution is flatter than those of ERM, SWA, and SAM. Our loss surface visualization shows that the SWAD solution is located on the center of the flat region, while ERM finds a boundary solution.
Finally, we show that the sharp boundary solutions by ERM are not generalized well, resulting in sensitivity to the model selection.
All following empirical analyses are conducted on \data{PACS} dataset, validating by all four domains (art painting, cartoon, photo, and sketch).

\paragraph{Local flatness anaylsis.}
% We verify that the proposed SWAD finds flatter minima than ERM and SWA.

To begin with, we quantify the local flatness of a model parameter $\theta$ by assuming that flat minima will have smaller changes of loss value within its neighborhoods than sharp minima.
For the given model parameter $\theta$, we compute the expected loss value changes between $\theta$ and parameters on the sphere surrounding $\theta$ with radius $\gamma$, \ie, $\mathcal{F}_\gamma (\theta)=\mathbb {E}_{\| \theta' \| = \|\theta\| + \gamma} [\mathcal{E}(\theta')-\mathcal{E}(\theta)]$.
In practice, $\mathcal{F}_\gamma(\theta)$ is approximated by Monte-Carlo sampling with 100 samples.
Note that the proposed local flatness $\mathcal{F}_\gamma(\theta)$ is computationally efficient than measuring curvature using the Hessian-based quantities. Also, $\mathcal{F}_\gamma(\theta)$ has an unbiased finite sample estimator, while the worst-case loss value, \ie, $\max_{\| \theta' \| = \|\theta\| + \gamma}[\mathcal{E}(\theta')-\mathcal{E}(\theta)]$ has no unbiased finite sample estimator.

In Figure~\ref{fig:flatness}, we compare $\mathcal{F}_\gamma(\theta)$ of ERM, SAM, SWA with cyclic learning rate, SWA with constant learning rate, and SWAD by varying radius $\gamma$.
% The solution found by SWA and SWAD shows lower local flatness quantities than ERM in all experiments.
SAM and SWA find the solutions with lower local flatness than ERM on average.
SWAD finds the most flat minimum in every experiment.
% We also observe that our dense and overfit-aware sampling strategy, \ie, SWAD, finds flatter optima than SWA and SAM does.

\begin{figure}[tb]
\begin{center}
\setlength{\tabcolsep}{0pt}
\newcommand\figwidth{.22}
\begin{tabular*}{0.92\textwidth}{c @{\extracolsep{\fill}} cccc}
    \rotatebox[origin=c]{90}{Train}
    &
    \raisebox{-0.45\height}{\includegraphics[width=\figwidth\textwidth]{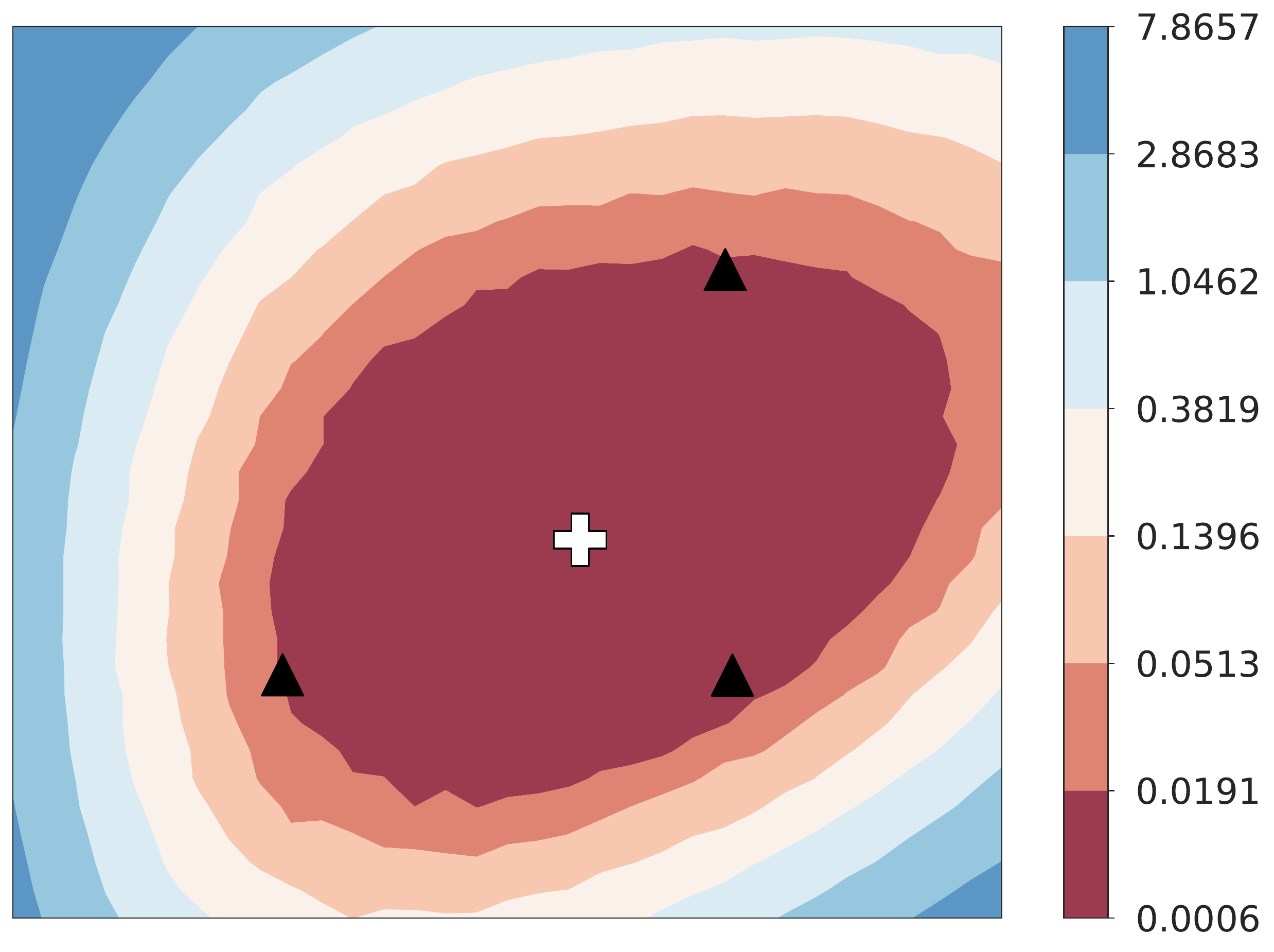}}
    &
    \raisebox{-0.45\height}{\includegraphics[width=\figwidth\textwidth]{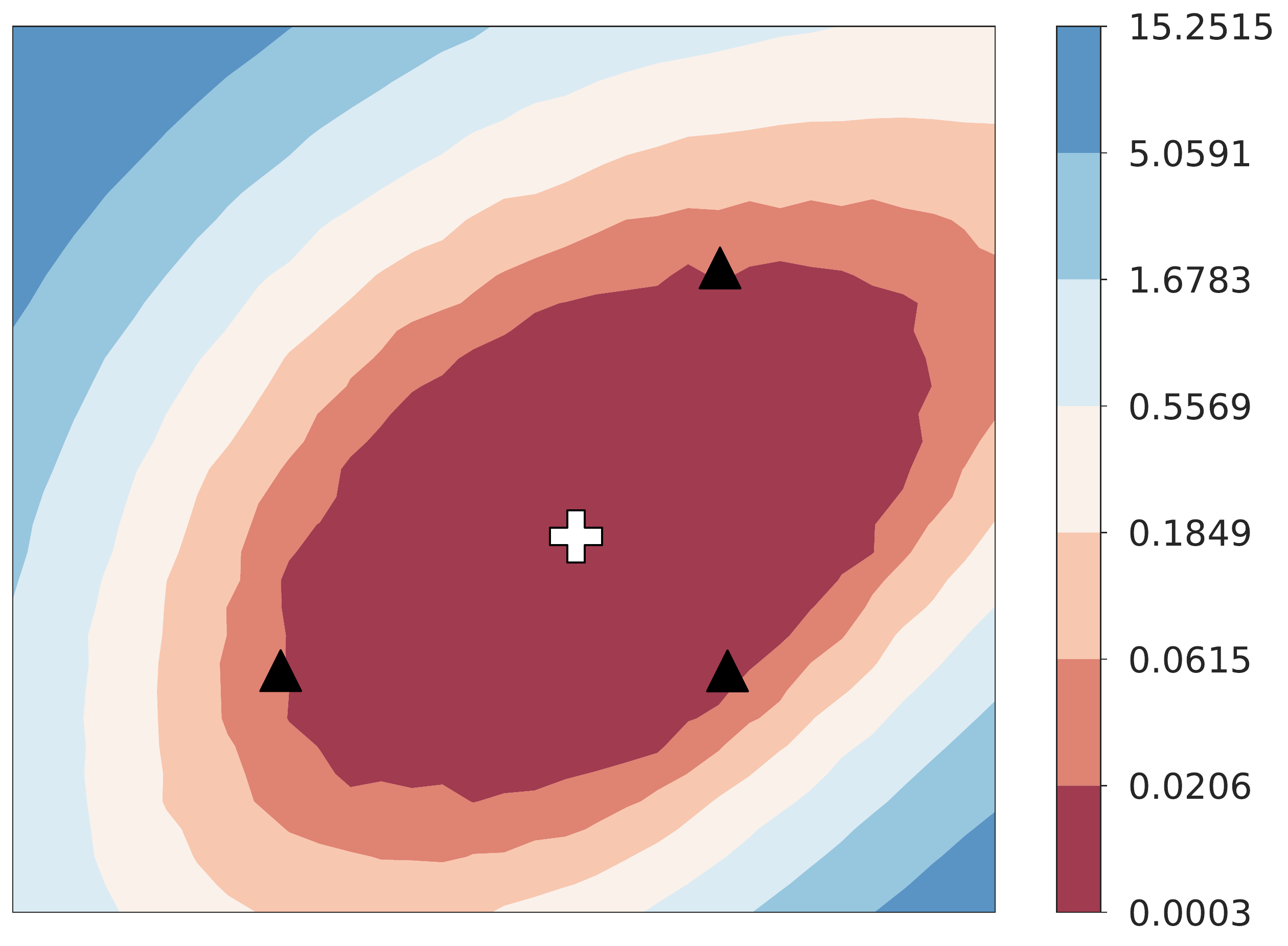}}
    &
    \raisebox{-0.45\height}{\includegraphics[width=\figwidth\textwidth]{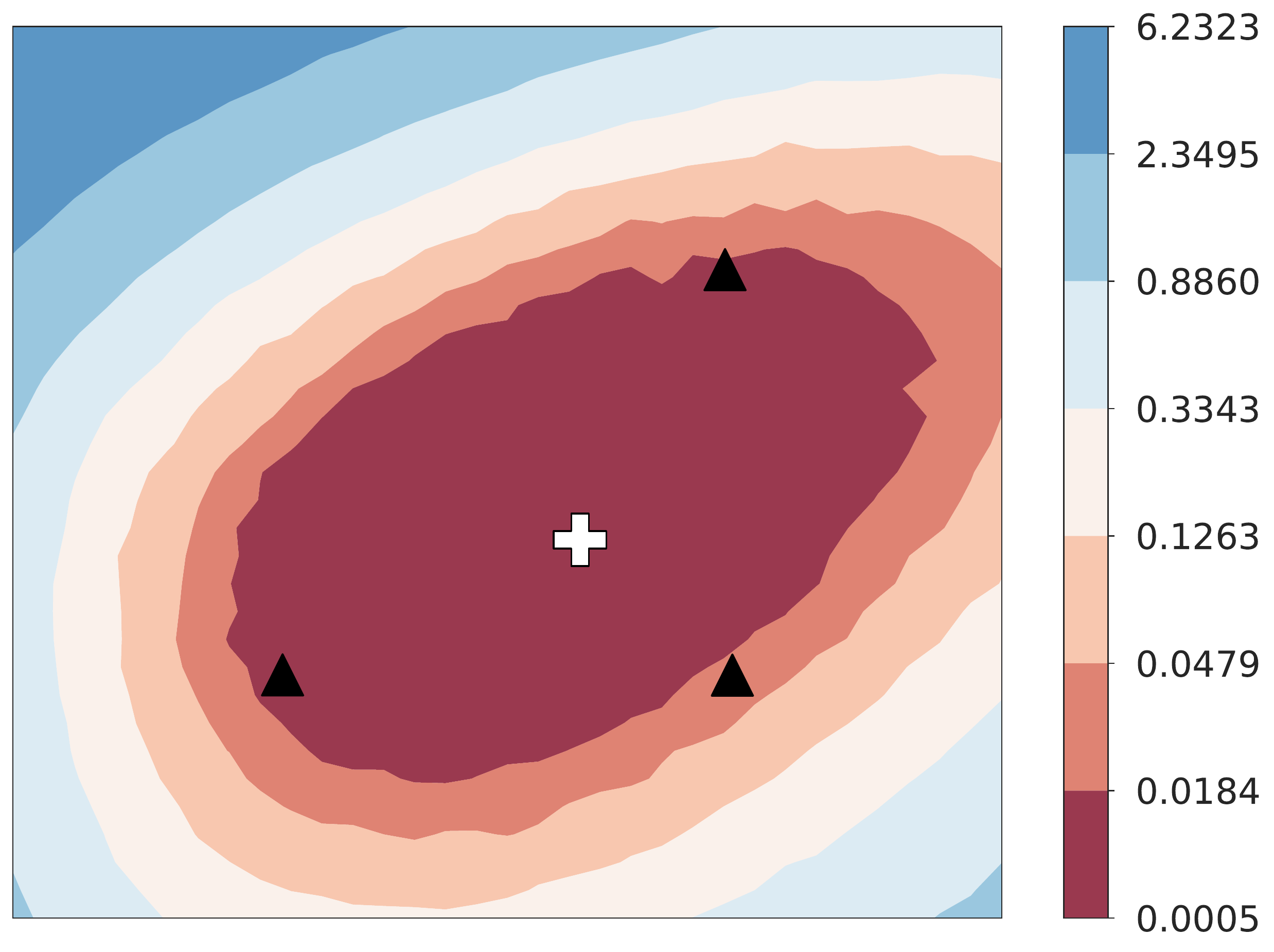}}
    &
    \raisebox{-0.45\height}{\includegraphics[width=\figwidth\textwidth]{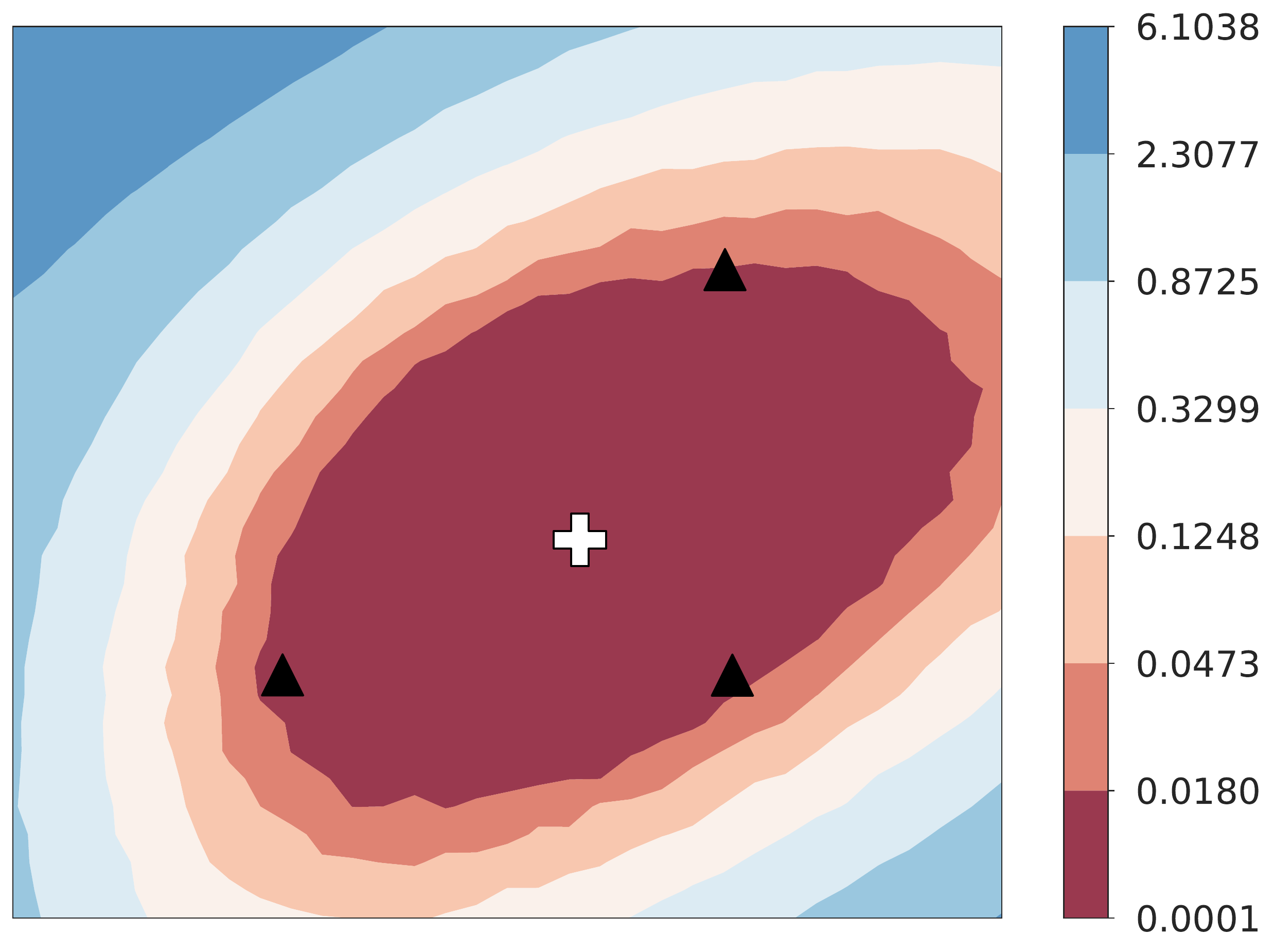}}
\\
    \rotatebox[origin=c]{90}{Test}
    &
    \raisebox{-0.5\height}{\includegraphics[width=\figwidth\textwidth]{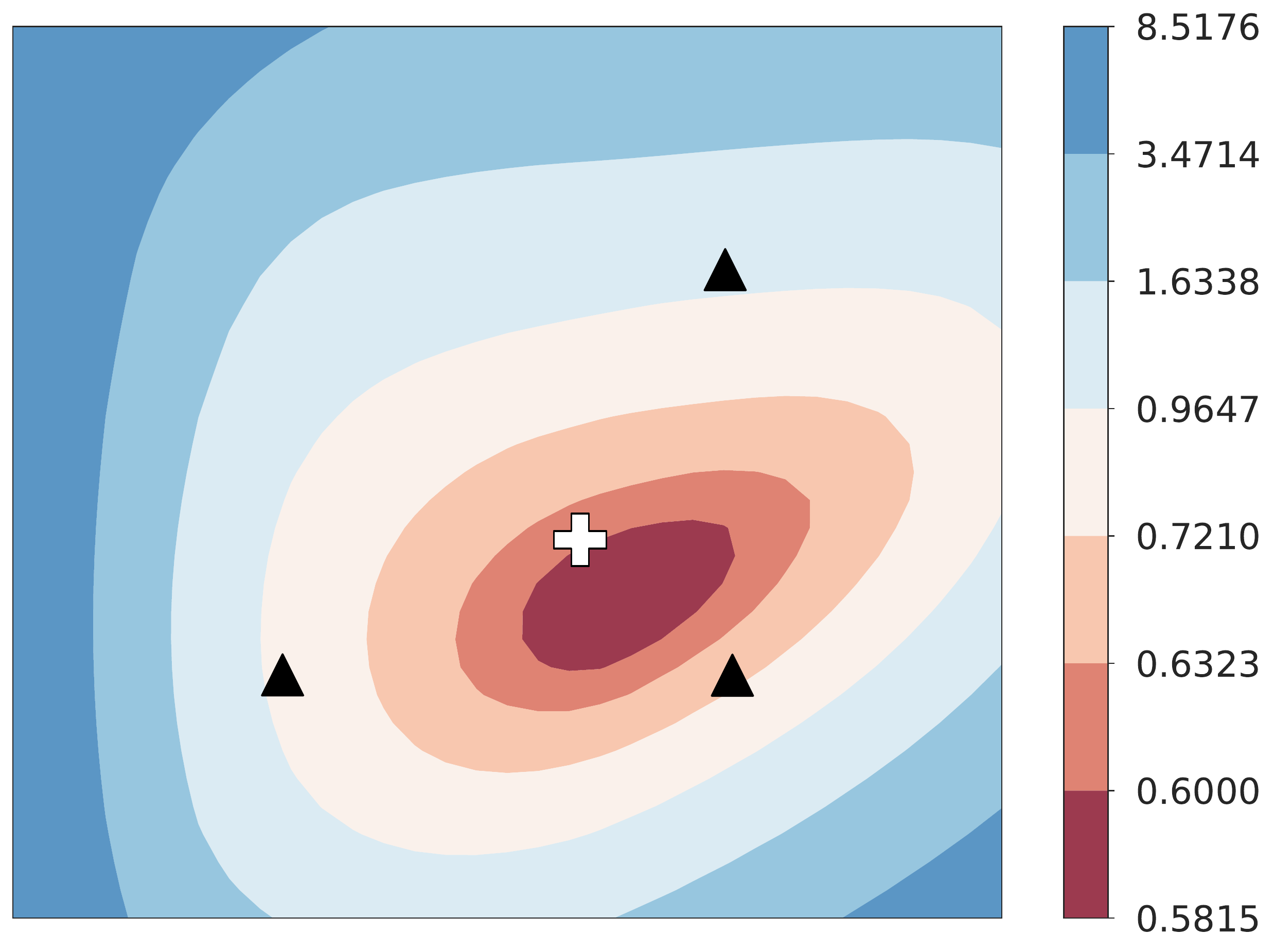}}
    &
    \raisebox{-0.5\height}{\includegraphics[width=\figwidth\textwidth]{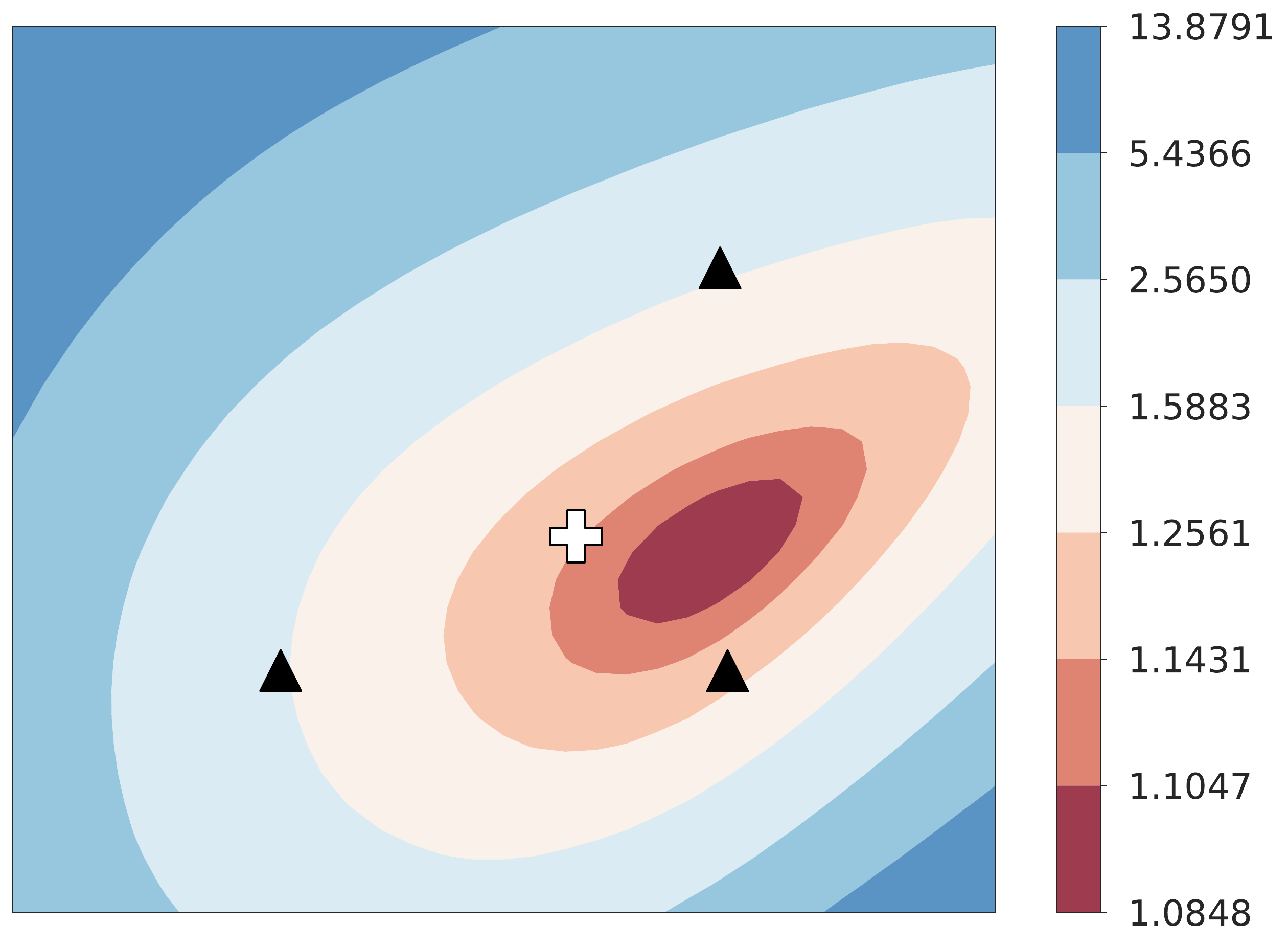}}
    &
    \raisebox{-0.5\height}{\includegraphics[width=\figwidth\textwidth]{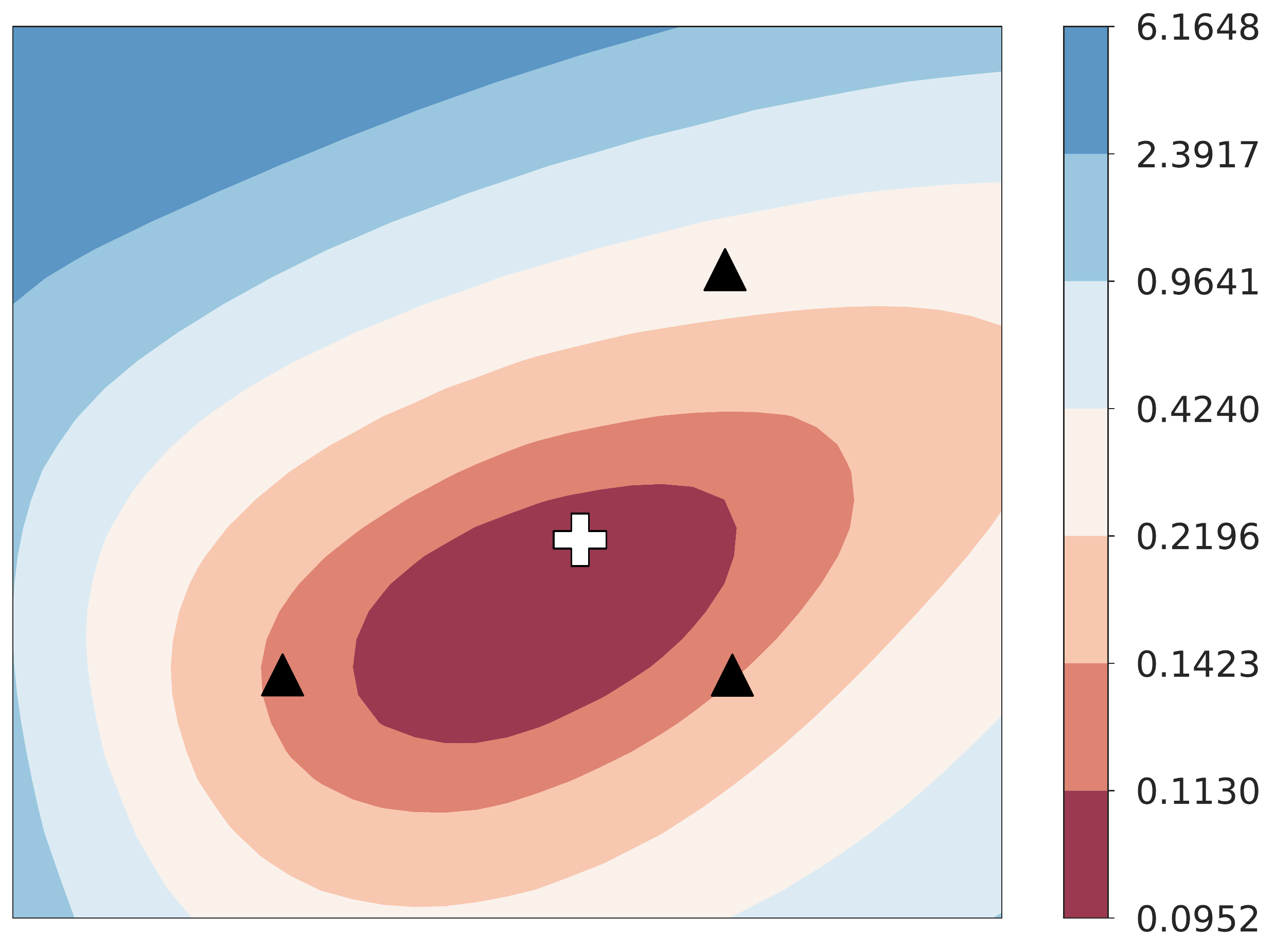}}
    &
    \raisebox{-0.5\height}{\includegraphics[width=\figwidth\textwidth]{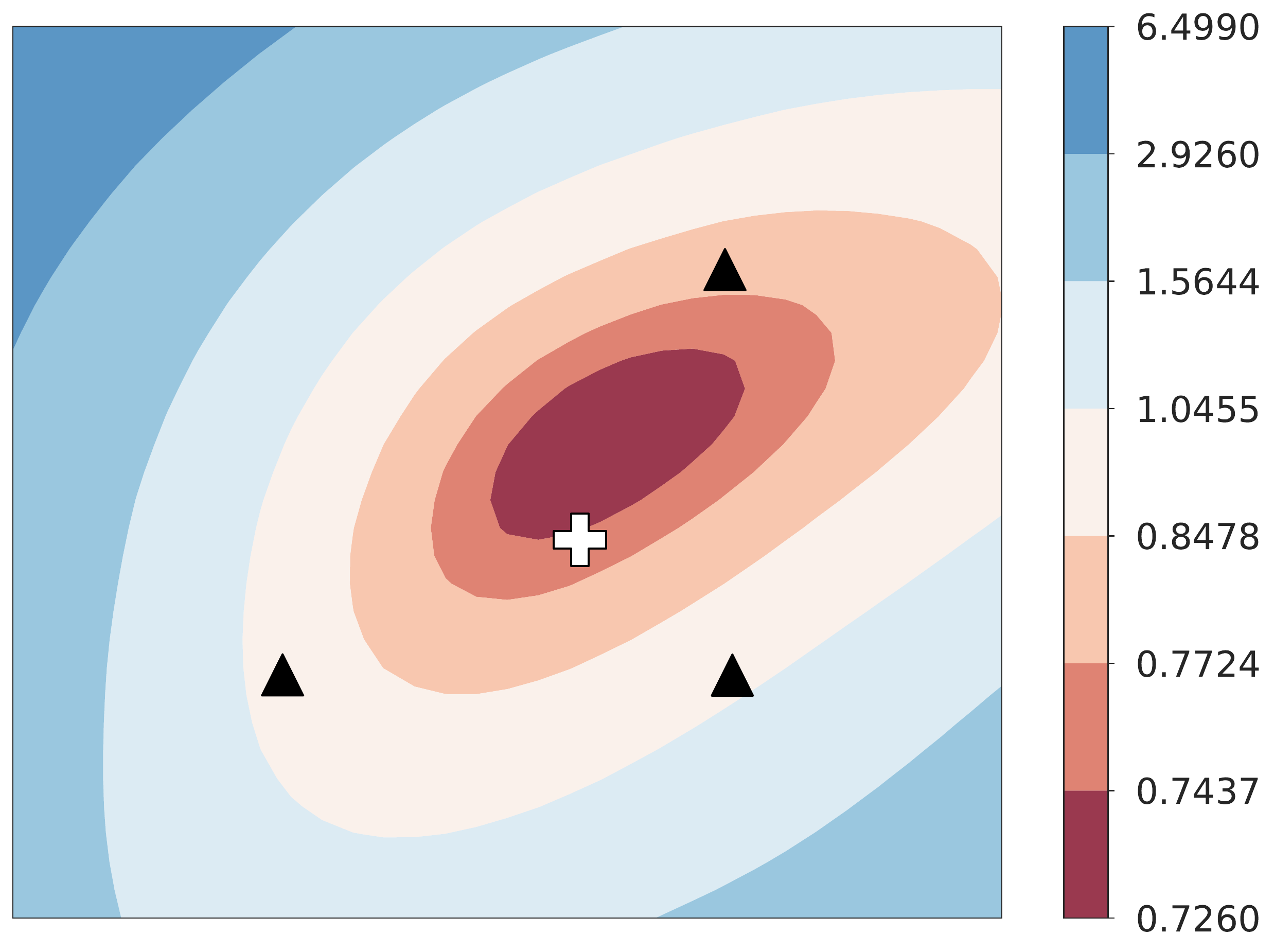}}
\\
    &
    {\small \textit{Art painting} (test)}
    &
    {\small \textit{Cartoon} (test)}
    &
    {\small \textit{Photo} (test)}
    &
    {\small \textit{Sketch} (test)}
\end{tabular*}
\end{center}
\vspace{-0.5em}
\caption{\small \textbf{Loss surfaces on model parameters in \data{PACS} dataset for each target domain.} 
The three triangles indicate model weights chosen at the end of training phase with equal intervals. Each plane is defined by the three weights and losses upon the plane are visualized with contours. The center cross mark is averaged point of the three weights. The first and second rows show the averaged training loss and the test loss surfaces, respectively.
% \sr{coloring the center yellow?}
}
\vspace{-0.5em}
\label{fig:loss_space}
\end{figure}

\begin{figure}
\centering
    \begin{subfigure}{0.5\linewidth}
        \includegraphics[width=\linewidth]{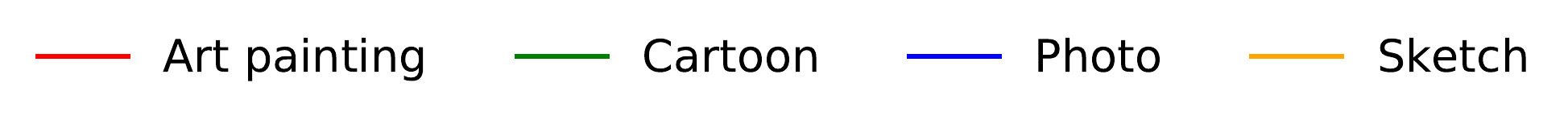}
    \end{subfigure}\\
    \begin{subfigure}{0.25\linewidth}
        \includegraphics[width=\linewidth]{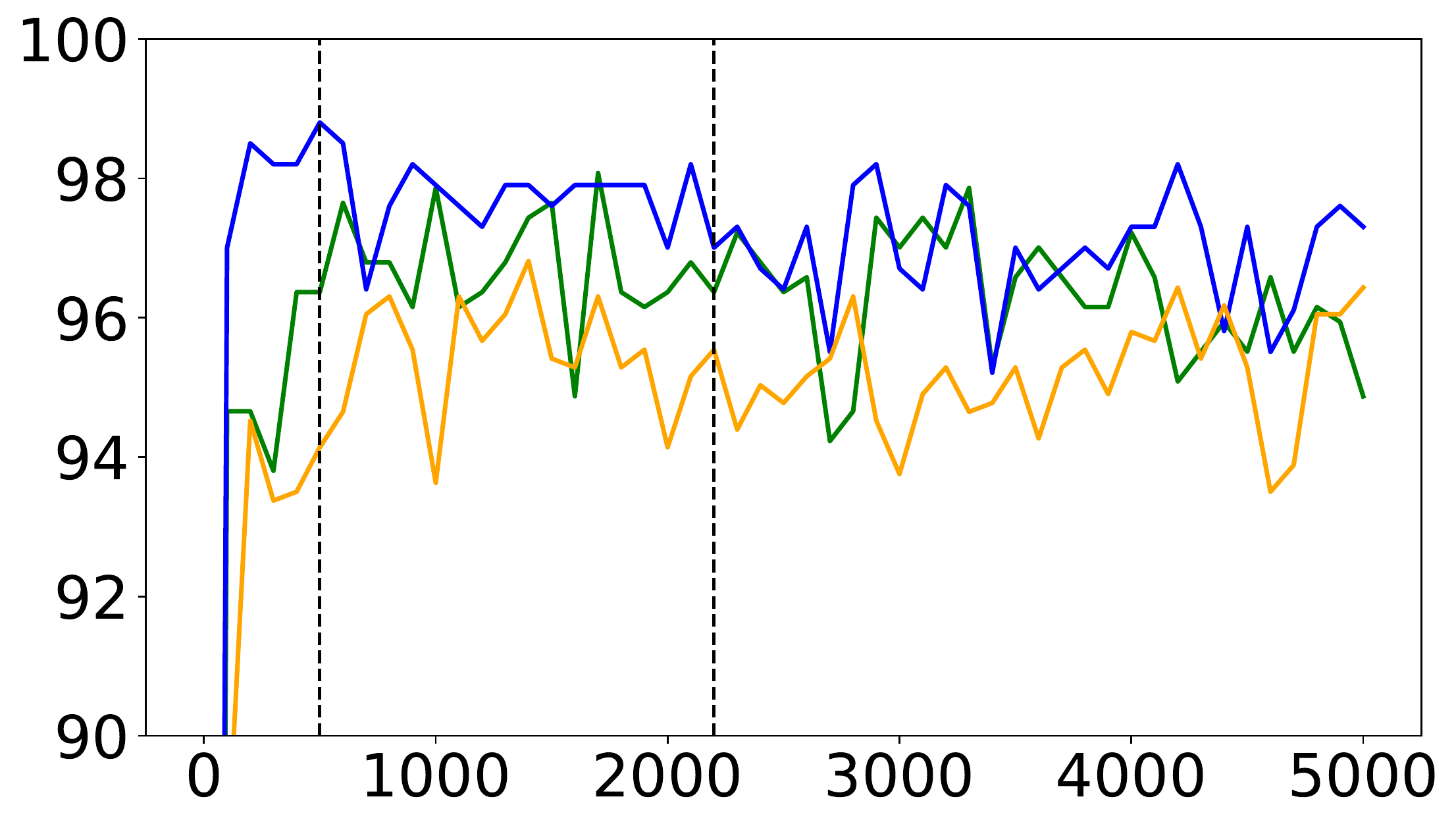}%
        \caption{\small \textit{Art painting} (test)}
    \end{subfigure}%
    \begin{subfigure}{0.25\linewidth}
        \includegraphics[width=\linewidth]{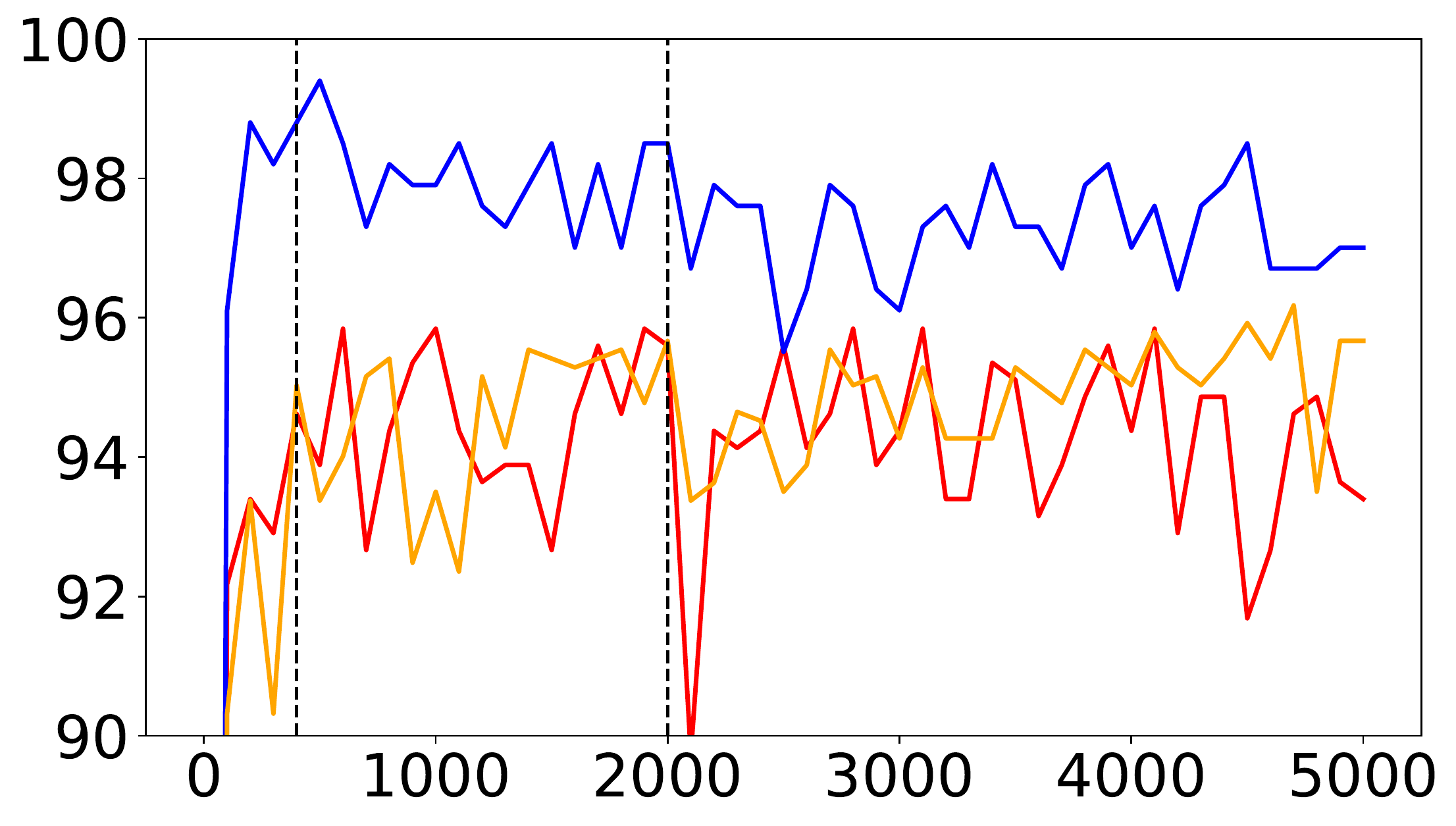}%
        \caption{\small \textit{Cartoon} (test)}
    \end{subfigure}%
    \begin{subfigure}{0.25\linewidth}
        \includegraphics[width=\linewidth]{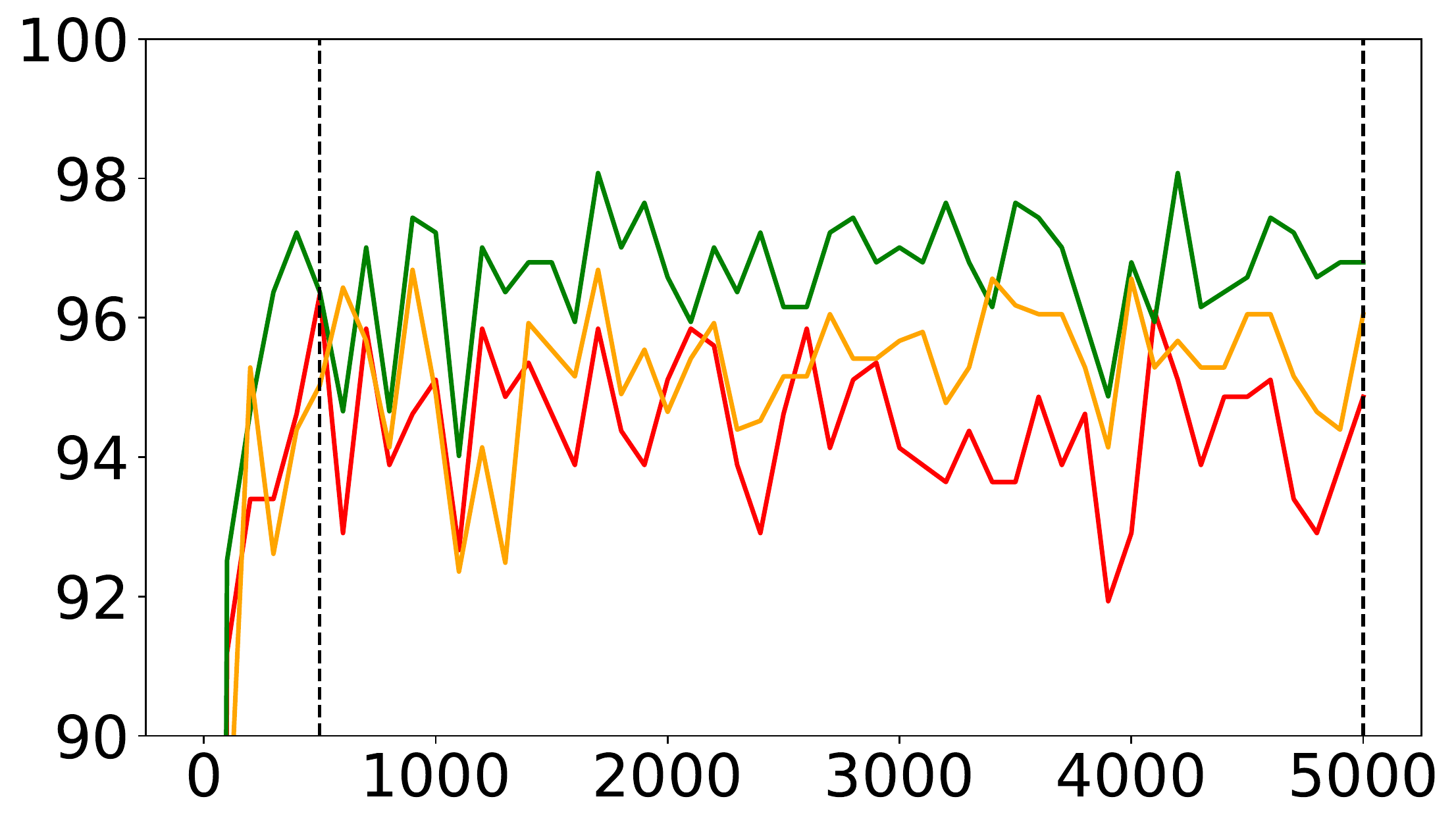}%
        \caption{\small \textit{Photo} (test)}
    \end{subfigure}%
    \begin{subfigure}{0.25\linewidth}
        \includegraphics[width=\linewidth]{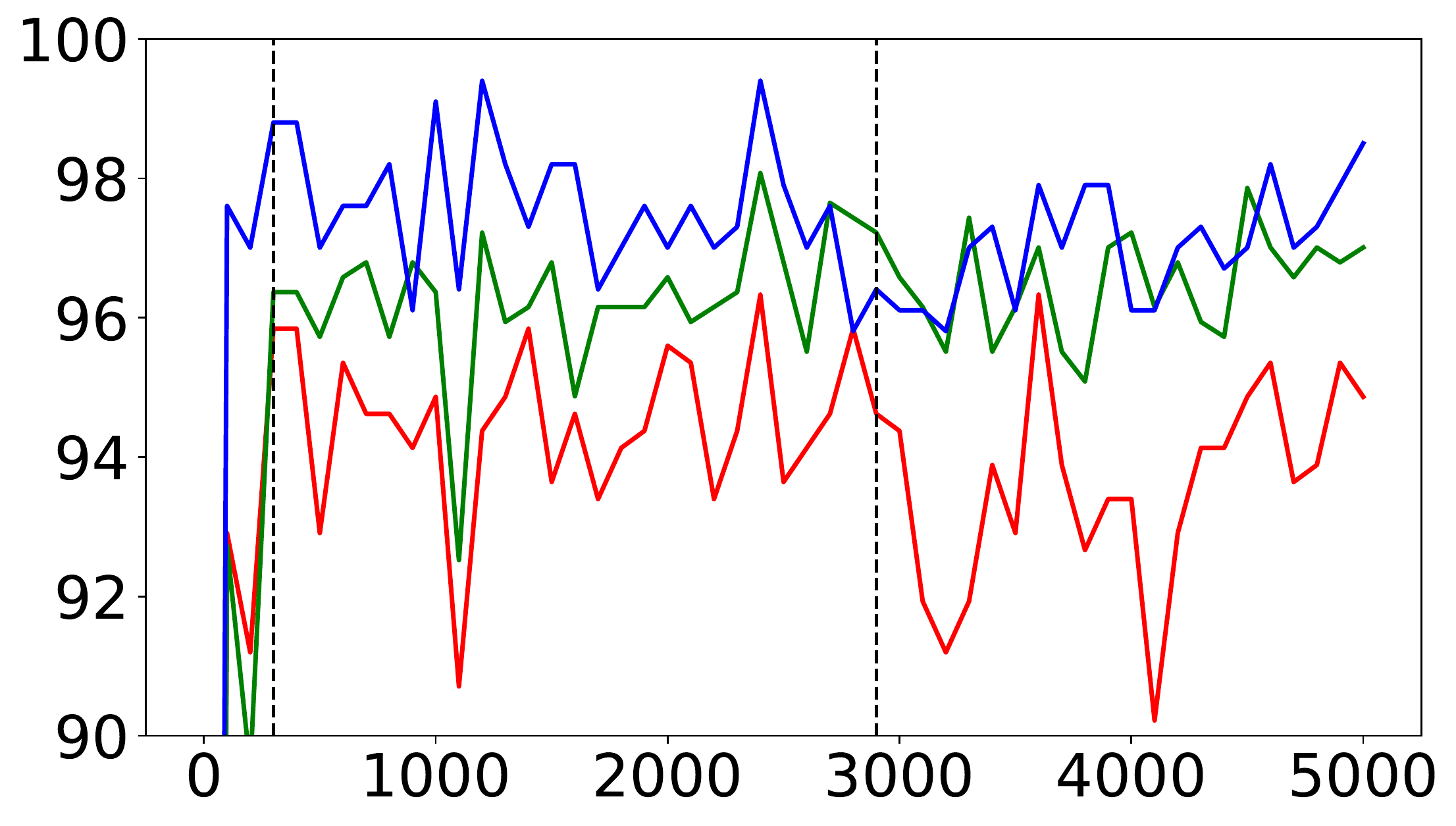}%
        \caption{\small \textit{Sketch} (test)}
    \end{subfigure}%
    \caption{\small \textbf{Validation accuracies for in-domains.} The X- and Y-axis indicate the training iterations and accuracy, respectively, about the validation domains (legend) and the test domain (caption). The vertical dot lines represent start and end iterations, $t_s$ and $t_e$, identified by the overfit-aware sampling strategy of SWAD.}
    \label{fig:accfluc}
\end{figure}

\paragraph{Loss surface visualization.}
We visualize the loss landscapes by choosing three model weights on the optimization trajectory ($\theta_1, \theta_2, \theta_3$)\footnote{We choose weights at iteration 2500, 3500, 4500 during the training.}, and computing the loss values by linear combinations of $\theta_1, \theta_2, \theta_3$\footnote{Each point is defined by two axes $u$ and $v$ computed by $u=\theta_2 - \theta_1$ and $v=\frac{(\theta_3 - \theta_1)- \langle \theta_3 - \theta_1, \theta_2 - \theta_1 \rangle}{\|\theta_2 - \theta_1 \|^2 \cdot (\theta_2 - \theta_1)}$.} as \cite{izmailov2018swa}.
More details are in Appendix B.5.
In Figure~\ref{fig:loss_space}, we observe that for all cases, ERM solutions are located at the boundary of a flat minimum of training loss, resulting in poor generalizability in test domains, that is aligned with our theoretical analysis and empirical flatness analysis.
Since ERM solutions are located on the boundary of a flat loss surface, we observe that ERM solutions are very sensitive to model selection.
In Figure~\ref{fig:accfluc}, we illustrate the validation accuracies for each train-test domain combination of \data{PACS} by ERM, over training iterations (one epoch is equivalent to 83 iterations).
We first observe that ERM rapidly reaches the best accuracy within only a few training epochs, namely less than 6 epochs.
Furthermore, the ERM validation accuracies fluctuate a lot, and the final performance is very sensitive to the model selection criterion.

On the other hand, we observe that SWA solutions are located on the center of the training loss surfaces as well as of the test loss surfaces (Figure~\ref{fig:loss_space}). Also, our overfit-aware stochastic weight gathering strategy (denoted as the vertical dot lines in Figure~\ref{fig:accfluc}) prevents the ensembled weight from overfitting and makes SWAD model selection-free.

\section{Experiments}
\label{s_empirical_results}
% In this section, we demonstrate the effectiveness of our SWAD in various domain generalization benchmarks.

\subsection{Evaluation protocols}
\label{section:s__evaluation_protocols}
% \begin{table}[tb]
%     \centering
%     \small
%     \caption{\small {\bf Domain generalization benchmarks.} We describe the details of each benchmark.}
%     \label{table:dataset}
%     \begin{tabular}{l @{\extracolsep{\fill}} rrr}
%     \toprule
%         Dataset & \# examples & \# classes & domains \\
%         \midrule
%         \data{PACS}~\cite{li2017pacs} & 9,991 & 7 & \textit{Photo}, \textit{Art painting}, \textit{Cartoon}, \textit{Sketch} \\
%         \data{VLCS}~\cite{fang2013vlcs} & 10,729 & 5 & \textit{VOC2007}, \textit{LabelMe}, \textit{Caltech101}, \textit{SUN09} \\
%         \data{OfficeHome}~\cite{venkateswara2017officehome} & 15,588 & 65 & \textit{Art}, \textit{Clipart}, \textit{Product}, \textit{Real} \\
%         \data{TerraIncognita}~\cite{beery2018terraincognita} & 24,788 & 10 & \textit{L100}, \textit{L38}, \textit{L43}, \textit{L46} \\
%         \data{DomainNet}~\cite{peng2019domainnet} & 586,575 & 345 & \textit{Clipart}, \textit{Infographic}, \textit{Painting}, \textit{Quickdraw}, \textit{Real}, \textit{Sketch}\\ \bottomrule
%     \end{tabular}
% \end{table}

% \paragraph{Dataset and optimization protocol.}
\textbf{Dataset and optimization protocol.}
Following \citet{gulrajani2020domainbed}, we exhaustively evaluate our method and comparison methods on various benchmarks: \data{PACS}~\cite{li2017pacs} (9,991 images, 7 classes, and 4 domains), \data{VLCS}~\cite{fang2013vlcs} (10,729 images, 5 classes, and 4 domains), \data{OfficeHome}~\cite{venkateswara2017officehome} (15,588 images, 65 classes, and 4 domains), \data{TerraIncognita}~\cite{beery2018terraincognita} (24,788 images, 10 classes, and 4 domains), and \data{DomainNet}~\cite{peng2019domainnet} (586,575 images, 345 classes, and 6 domains). 

\newcolumntype{x}[1]{>{\centering\arraybackslash\hspace{0pt}}p{#1}}

\begin{table}[t]
\centering
\small
\caption{\small {\bf Comparison with domain generalization methods and SWAD.} Out-of-domain accuracies on five domain generalization benchmarks are shown. We highlight the \textbf{best results} and the \underline{second best results}. Note that ERM (reproduced), Mixstyle are reproduced numbers, and other numbers are from the original literature and \citet{gulrajani2020domainbed} (denoted with $\dagger$).
Our experiments are repeated three times.
%Our reproduced numbers and results are average of three different runs. 
}
\label{table:full_table}
\vspace{.5em}
\begin{tabular}{lccccc|c}
\toprule
\textbf{Algorithm}        & \data{PACS} & \data{VLCS}             & \data{OfficeHome}       & \data{TerraInc}   & \data{DomainNet}        & \textbf{Avg.}              \\
% \textbf{Algorithm}        & \data{PACS}~\cite{li2017pacs}            & \data{VLCS}~\cite{fang2013vlcs}             & \data{OfficeHome}~\cite{venkateswara2017officehome}       & \data{TerraInc}~\cite{beery2018terraincognita}   & \data{DomainNet}~\cite{peng2019domainnet}        & \textbf{Avg.}              \\
\midrule
MASF~\cite{dou2019masf}     & 82.7          & -             & -                   & -                 & -                  & -              \\
DMG~\cite{chattopadhyay2020dmg}      & 83.4          & -             & -                   & -                 & 43.6               & -              \\
MetaReg~\cite{balaji2018metareg}  & 83.6          & -             & -                   & -                 & 43.6               & -              \\
ER~\cite{zhao2020er_entropy_regularization}       & 85.3          & -             & -                   & -                 & -                  & -              \\
pAdaIN~\cite{nuriel2020padain}   & 85.4          & -             & -                   & -                 & -                  & -              \\
EISNet~\cite{wang2020eisnet}   & 85.8          & -             & -                   & -                 & -                  & -              \\
DSON~\cite{seo2020dson}     & \underline{86.6}          & -             & -                   & -                 & -                  & -              \\
ERM$^\dagger$~\cite{vapnik1998statistical}      & 85.5          & 77.5          & 66.5                & 46.1              & 40.9               & 63.3           \\
ERM (reproduced)     & 84.2          & 77.3          & 67.6                & 47.8              & \underline{44.0}               & 64.2           \\
IRM$^\dagger$~\cite{arjovsky2019irm}      & 83.5          & 78.6          & 64.3                & 47.6              & 33.9               & 61.6           \\
GroupDRO$^\dagger$~\cite{Sagawa2020GroupDRO} & 84.4          & 76.7          & 66.0                & 43.2              & 33.3               & 60.7           \\
I-Mixup$^\dagger$~\cite{xu2020interdomain_mixup_aaai,yan2020interdomain_mixup,wang2020interdomain_mixup_icassp_dg}    & 84.6          & 77.4          & 68.1                & 47.9              & 39.2               & 63.4           \\
MLDG$^\dagger$~\cite{li2018mldg}     & 84.9          & 77.2          & 66.8                & 47.8              & 41.2               & 63.6           \\
CORAL$^\dagger$~\cite{sun2016coral}    & 86.2          & \underline{78.8}          & \underline{68.7}                & 47.7              & 41.5               & 64.5           \\
MMD$^\dagger$~\cite{li2018mmd}      & 84.7          & 77.5          & 66.4                & 42.2              & 23.4               & 58.8           \\
DANN$^\dagger$~\cite{ganin2016dann}     & 83.7          & 78.6          & 65.9                & 46.7              & 38.3               & 62.6           \\
CDANN$^\dagger$~\cite{li2018cdann}    & 82.6          & 77.5          & 65.7                & 45.8              & 38.3               & 62.0           \\
MTL$^\dagger$~\cite{blanchard2021mtl_marginal_transfer_learning}      & 84.6          & 77.2          & 66.4                & 45.6              & 40.6               & 62.9           \\
SagNet$^\dagger$~\cite{nam2019sagnet}   & 86.3          & 77.8          & 68.1                & \underline{48.6}              & 40.3               & 64.2           \\
ARM$^\dagger$~\cite{zhang2020arm}      & 85.1          & 77.6          & 64.8                & 45.5              & 35.5               & 61.7           \\
VREx$^\dagger$~\cite{krueger2020vrex}     & 84.9          & 78.3          & 66.4                & 46.4              & 33.6               & 61.9           \\
RSC$^\dagger$~\cite{huang2020rsc}      & 85.2          & 77.1          & 65.5                & 46.6              & 38.9               & 62.7           \\
Mixstyle~\cite{zhou2021mixstyle} & 85.2          & 77.9          & 60.4                & 44.0              & 34.0               & 60.3           \\ \midrule
% SAM      & 85.8          & \textbf{79.4} & 69.6                & 43.3              & 44.3               & 64.5           \\ \midrule
SWAD (ours)    & \textbf{88.1} & \textbf{79.1}          & \textbf{70.6}       & \textbf{50.0}     & \textbf{46.5}      & \textbf{66.9} \\ \addlinespace[-0.15cm]
& \scriptsize{($\pm0.1$)} & \scriptsize{($\pm0.1$)} & \scriptsize{($\pm0.2$)} & \scriptsize{($\pm0.3$)} & \scriptsize{($\pm0.1$)} &\\
\bottomrule
\end{tabular}
\end{table}

For a fair comparison, we follow training and evaluation protocol by \citet{gulrajani2020domainbed}, including the dataset splits, hyperparameter (HP) search and model selection (while SWAD does not need it) on the validation set, and optimizer HP, except the HP search space and the number of iterations for \data{DomainNet}.
We use a reduced HP search space to reduce the computational costs.
We also tripled the number of iterations for \data{DomainNet} from 5,000 to 15,000 because we observe that 5,000 is not sufficient to convergence. We re-evaluate ERM with 15,000 iterations, and observe 3.1pp average performance improvement ($40.9\% \rightarrow 44.0\%$) in \data{DomainNet}.
For training, we choose a domain as the target domain and use the remaining domains as the training domain where 20\% samples are used for validation and model selection.
ImageNet~\cite{russakovsky2015imagenet} trained ResNet-50~\cite{he2016_cvpr_resnet} is employed as the initial weight, and optimized by Adam~\cite{kingma2015adam} optimizer with a learning rate of 5e-5. We construct a mini-batch containing all domains where each domain has 32 images.
We set SWAD HPs $N_s$ to 3, $N_e$ to 6, and $r$ to 1.2 for \data{VLCS} and 1.3 for the others by HP search on the validation sets.
% Note that unlike previous methods, SWAD does not need a model selection criterion, but automatically select the ensemble weight by the given hyperparameters.
% In addition, we observe that the number of iterations for \data{DomainNet} used by \citet{gulrajani2020domainbed} (5,000) is not sufficient. Instead, we train the methods with 15,000 iterations for \data{DomainNet}. We re-evaluate a ERM method with 15,000 iterations and observe 0.9pp average improvement ($63.3\% \rightarrow 64.2\%$).
% Except the number of iterations on \data{DomainNet}, we follow the optimization setting by \citet{gulrajani2020domainbed}: 
Additional implementation details, such as other HPs, are given in Appendix B.

\textbf{Evaluation metrics.}
% \paragraph{Evaluation metrics.}
% In the all following experiments, w
We report out-of-domain accuracies for each domain and their average, \ie, a model is trained and validated on training domains and evaluated on the unseen target domain. Each out-of-domain performance is an average of three different runs with different train-validation splits.
% We report the full table including standard error in Appendix.

\subsection{Main results}
\label{label:s__main_results}

\begin{wraptable}{r}{.42\linewidth}
\vspace{-1.5em}
\caption{\small \textbf{Comparison between generalization methods on \data{PACS}.} The scores are averaged over all settings using different target domains. \greenp{$\uparrow$} and \redp{$\downarrow$} indicate statistically significant improvement and degradation from ERM.}
\vspace{-0.5em}
%\resizebox{\linewidth}{!} {
\centering
\setlength{\tabcolsep}{3pt}
\begin{tabular}{@{}lcc@{}} 
\toprule
                & \multicolumn{1}{l}{Out-of-domain} & \multicolumn{1}{l}{In-domain}  \\
\midrule
ERM             & 85.3\scriptsize{$\pm0.4$} & 96.6\scriptsize{$\pm0.0$}                      \\
EMA             & 85.5{\scriptsize$\pm0.4$}(-) & 97.0{\scriptsize$\pm0.1$}\greenp{$\uparrow$}                      \\
SAM             & 85.5{\scriptsize$\pm0.1$}(-) & 97.4{\scriptsize$\pm0.1$}\greenp{$\uparrow$}                     \\
Mixup           & 84.8{\scriptsize$\pm0.3$}(-) & 97.3{\scriptsize$\pm0.1$}\greenp{$\uparrow$}                      \\
CutMix          & 83.8{\scriptsize$\pm0.4$}\redp{$\downarrow$} & 97.6{\scriptsize$\pm0.1$}\greenp{$\uparrow$}                      \\
% SAM             & 84.9 \scriptsize{$\pm0.4$}                          & 97.3 \scriptsize{$\pm0.2$}                      \\
VAT             & 85.4{\scriptsize$\pm0.6$}(-)  & 96.9{\scriptsize$\pm0.2$}\greenp{$\uparrow$}                      \\
$\Pi$-model     & 83.5{\scriptsize$\pm0.5$}\redp{$\downarrow$}  & 96.8{\scriptsize$\pm0.2$}\greenp{$\uparrow$}                      \\
\midrule
SWA      & 85.9{\scriptsize$\pm0.1$}\greenp{$\uparrow$}        & 97.1{\scriptsize$\pm0.1$}\greenp{$\uparrow$}                      \\
% SWA$_{\text{fit-on-val}}$ & 86.2 \scriptsize{$\pm0.2$}                          & 97.5 \scriptsize{$\pm0.2$}                      \\
SWAD   & \textbf{87.1}{\scriptsize$\pm0.2$}\greenp{$\uparrow$} & \textbf{97.7}{\scriptsize$\pm0.1$}\greenp{$\uparrow$}             \\
\bottomrule
\end{tabular}
%}
\vspace{-1em}
\label{table:generalization}
\end{wraptable}
\textbf{Comparison with domain generalization methods.}
% \paragraph{Comparison with domain generalization methods.}
We report the full out-of-domain performances on five DG benchmarks in Table~\ref{table:full_table}.
The full tables including out-of-domain accuracies for each domain are in Appendix E.
In all experiments, our SWAD achieves significant performance gain against ERM as well as the previous best results: +2.6pp in \data{PACS}, +0.3pp in \data{VLCS}, +1.4pp in \data{TerraIncognita}, +1.9pp in \data{OfficeHome}, and +2.9pp in \data{DomainNet} comparing to the previous best results.
We observe that SWAD provides two practical advantages comparing to previous methods. First, SWAD does not need any modification on training objectives or model architecture, \ie, it is universally applicable to any other methods.
As an example, we show that SWAD actually improves the performances of other DG methods, such as CORAL~\cite{sun2016coral} in Table~\ref{table:swad_combination}.
Moreover, as we discussed before, SWAD is free to the model selection, resulting in stable performances (\ie, small standard errors) on various benchmarks.
Note that we only compare results with ResNet-50 backbone for a fair comparison.
We describe the implementation details of each comparison method and the hyperparameter search protocol in Appendix B.

\textbf{Comparison with conventional generalization methods.}
We also compare SWAD with other conventional generalization methods to show that the remarkable domain generalization gaps by SWAD is not achieved by better generalization, but by seeking flat minima.
The comparison methods include flatness-aware optimization methods, such as SAM~\cite{foret2020sharpness},
ensemble methods, such as EMA~\cite{polyak1992ema},
data augmentation methods, such as Mixup~\cite{zhang2018mixup} and CutMix~\cite{yun2019cutmix}, and consistency regularization methods, such as VAT~\cite{miyato2018vat} and $\Pi$-model~\cite{LaineA17iclr_pi_model}.
We also split in-domain datasets into training (60\%), validation (20\%), and test (20\%) splits, while no in-domain test set used for Table~\ref{table:full_table}.
Every experiment is repeated three times.

The results are shown in Table~\ref{table:generalization}.
We observe that all conventional methods helps in-domain generalization, \ie, performing better than ERM on in-domain test set. However, their out-of-domain performances are similar to or even worse than ERM.
For example, CutMix and $\Pi$-model improve in-domain performances by 1.0pp and 0.2pp but degrade out-of-domain performances by 1.5pp and 1.8pp. 
SAM, another method for seeking flat minima, slightly increases both in-domain and out-of-domain performances but the out-of-domain performance is not statistically significant.
%We will discuss more out-of-domain accuracies by SAM in other benchmarks later.
We will discuss performances of SAM in other benchmarks later.
In contrast, the vanilla SWA and our SWAD significantly improve both in-domain and out-of-domain performances. SWAD improves the performances by SWA with statistically significantly gaps: 1.2pp on the out-of-domain and 0.6pp on the in-domain. Further comparison between SWA and SWAD is provided in \S\ref{label:s__ablation}.

\begin{table}[b]
\centering
\small
\caption{\small {\bf Combination of SWAD and other methods.} The scores are averaged over every target domain case. The performances of ERM, CORAL, and SAM are optimized by HP searches of DomainBed. In contrast, for the SWAD combination cases, 
%each algorithm and SWAD use HPs from default and ERM + SWAD, respectively.
% \jb{each algorithm uses default HPs and SWAD adopts HPs found in ERM + SWAD, without additional HP search.}
CORAL and SAM use default HPs without additional HP search.
% SWAD is simply applied without any HP search.
We additionally compare SWAD to SWA$_\text{w/ const}$.
Note that ERM + SWAD is same as ``SWAD'' in Table~\ref{table:full_table}.
% We additionally report more results by SWA variants in Appendix D.1.
%Note that the methods combined with SWAD show better performance than the hyperparameter searched baseline results, even though they use the default hyperparameters. \jb{In DomainNet column, ERM/CORAL use 5000 steps, while the others use 15000 steps.}
}
\label{table:swad_combination}
\vspace{0.2cm}
\renewcommand{\arraystretch}{1.1}
\begin{tabular}{lccccc|c} 
\toprule
                & \data{PACS} & \data{VLCS} & \data{OfficeHome} & \data{TerraInc} & \data{DomainNet} & Avg. ($\Delta$)  \\ \midrule
ERM             & 85.5 \scriptsize$\pm0.2$ & 77.5 \scriptsize$\pm0.4$ & 66.5 \scriptsize$\pm0.3$ & 46.1 \scriptsize$\pm1.8$ & 40.9 \scriptsize$\pm0.1$ & 63.3  \\
ERM + SWA$_\text{w/ const}$                                    & 86.9 \scriptsize$\pm0.2$ & 76.6 \scriptsize$\pm0.1$ & 69.3 \scriptsize$\pm0.3$       & 49.2 \scriptsize$\pm1.2$     & 45.9 \scriptsize$\pm0.0$      & 65.6 (+2.3) \\
ERM + SWAD      & 88.1 \scriptsize$\pm0.1$ & 79.1 \scriptsize$\pm0.1$ & 70.6 \scriptsize$\pm0.2$ & 50.0 \scriptsize$\pm0.3$ & 46.5 \scriptsize$\pm0.1$ & 66.9 (+3.6)  \\
\midrule
CORAL           & 86.2 \scriptsize$\pm0.3$ & 78.8 \scriptsize$\pm0.6$ & 68.7 \scriptsize$\pm0.3$ & 47.6 \scriptsize$\pm1.0$ & 41.5 \scriptsize$\pm0.1$ & 64.5  \\
CORAL + SWAD    & 88.3 \scriptsize$\pm0.1$ & 78.9 \scriptsize$\pm0.1$ & 71.3 \scriptsize$\pm0.1$ & 51.0 \scriptsize$\pm0.1$ & 46.8 \scriptsize$\pm0.0$ & 67.3 (+2.8) \\
\midrule
SAM             & 85.8 \scriptsize$\pm0.2$ & 79.4 \scriptsize$\pm0.1$ & 69.6 \scriptsize$\pm0.1$ & 43.3 \scriptsize$\pm0.7$ & 44.3 \scriptsize$\pm0.0$ & 64.5  \\
SAM + SWAD      & 87.1 \scriptsize$\pm0.2$ & 78.5 \scriptsize$\pm0.2$ & 69.9 \scriptsize$\pm0.1$ & 45.3 \scriptsize$\pm0.9$ & 46.5 \scriptsize$\pm0.1$ & 65.5 (+1.0)     \\
\bottomrule
\end{tabular}
\end{table}
\textbf{Combinations with other methods.}
Since SWAD does not require any modification on training procedures and model architectures, SWAD is universally applicable to any other methods. Here, we combine SWAD with ERM, CORAL~\cite{sun2016coral}, and SAM~\cite{foret2020sharpness}. Results are shown in Table~\ref{table:swad_combination}.
% When comparing baselines, 
Both CORAL and SAM solely show better performances than ERM with +1.2pp average out-of-domain accuracy gap.
Note that SAM is not a DG method but a sharpness-aware optimization method to find flat minima.
It supports our theoretical motivation: DG can be achieved by seeking flat minima.

By applying SWAD on the baselines, the performances are consistently improved by 3.6pp on ERM, 2.8pp on CORAL, and 1.0pp on SAM. Interestingly, CORAL + SWAD show the best performances with both incorporating different advantages of utilizing domain labels and seeking flat minima.
We also observe that SAM + SWAD shows worse performance than ERM + SWAD, while SAM performs better than ERM.
We conjecture that it is because the objective control by SAM restricts the model parameter diversity durinig training, reducing the diversity for SWA ensemble.
However, applying SWAD on SAM still leads to better performances than the sole SAM. The results demonstrate that the application of SWAD on other baselines is a simple yet effective method for DG.

% \vspace{-.5em}
\subsection{Ablation study}
\label{label:s__ablation}
\vspace{-.5em}

\begin{table}[h]
\centering
\small
\setlength{\tabcolsep}{3.0pt}
\renewcommand{\arraystretch}{1.2}
\vspace{-0.1cm}
\caption{\small \textbf{Ablation studies of the stochastic weights selection strategies on \data{PACS} and \data{VLCS}.} 
%The scores are averaged over every target domain. 
In the configuration, 
``$t_s$'', ``$t_e$'', ``lr'', and ``interval'' indicate start and end iterations of sampling, a learning rate schedule, and a stochastic weight sampling interval, respectively. 
%4000 and 5000 indicate the iteration numbers where 5000 is the total number of iterations. 
``Opt'' and ``Overfit'' indicate the start and end iterations identified by our overfit-aware sampling strategy, and ``Val'' means the start and end iterations whose averaging shows the best accuracy on the validation set. ``Cyclic'' and ``Const'' represent cyclic and constant learning rate schedules.
All experiments are repeated three times.
%That is, SWAD$_\text{fit-on-val}$ fits the weight gathering range to provide the best validation accuracy.
%the moments detected to be optimized and overfitted, respectively.
%In the lr column, Cyclic and Const indicate cyclic and constant learning rate schedules, respectively. 
%The intervals indicate iteration gaps between the stochastic weights. 
%100 and 1 in the interval column are the stochastic weight selection intervals. 
%When using cyclic learning rate, cycle length is same as the interval. 
% For example, SWA$_\text{cyclic}$ averages from 4000 to 5000 steps with 100 interval using cyclic learning rate. The interval of 100 corresponds to 1.6 epoch for \data{PACS} and 1.5 epoch for \data{VLCS}.
}
% \vspace{0.2cm}
\vspace{.5em}
\begin{tabular}{@{}l|cccc|ccc|ccc@{}} 
\toprule
                        & \multicolumn{4}{c|}{Configuration}   & \multicolumn{3}{c|}{Out-of-domain} & \multicolumn{3}{c}{In-domain}  \\
                         & $t_s$ & $t_e$     & lr    & interval & \data{PACS} & \data{VLCS} & Avg.                & \data{PACS} & \data{VLCS} & Avg.             \\
\midrule
SWA$_\text{w/ cyclic}$              & 4000  & 5000   & Cyclic & 100    & 85.9 \scriptsize{$\pm0.1$} & 76.6 \scriptsize{$\pm0.1$} & 81.2        & 97.1 \scriptsize{$\pm0.1$} & 85.0 \scriptsize{$\pm0.2$} & 91.0             \\
SWA$_\text{w/ const}$               & 4000  & 5000   & Const & 100    & 86.5 \scriptsize{$\pm0.3$} & 76.7 \scriptsize{$\pm0.2$} & 81.6        & 97.3 \scriptsize{$\pm0.1$} & 85.0 \scriptsize{$\pm0.2$} & 91.1             \\
SWAD$_\text{w/o Dense}$                & Opt   & Overfit & Const & 100    & 86.5 \scriptsize{$\pm0.4$} & 78.0 \scriptsize{$\pm0.7$} & 82.2        & 97.6 \scriptsize{$\pm0.1$} & 85.8 \scriptsize{$\pm0.4$} & 91.7             \\
SWAD$_\text{w/o Opt-Overfit}$       & 4000  & 5000   & Const & 1     & 86.6 \scriptsize{$\pm0.6$} & 76.9 \scriptsize{$\pm0.3$} & 81.7        & 97.5 \scriptsize{$\pm0.1$} & 85.2 \scriptsize{$\pm0.1$} & 91.3             \\
SWAD$_\text{w/o Overfit}$              & Opt   & 5000   & Const & 1     & \textbf{87.1} \scriptsize{$\pm0.3$} & 77.6 \scriptsize{$\pm0.1$} & 82.4        & \textbf{97.7} \scriptsize{$\pm0.1$} & 85.8 \scriptsize{$\pm0.3$} & 91.8             \\
SWAD$_\text{fit-on-val}$      & Val   & Val     & Const & 1   & 86.2 \scriptsize{$\pm0.2$}         & 78.6 \scriptsize{$\pm0.1$}         & 82.4          & 97.5 \scriptsize{$\pm0.2$}         & 85.8 \scriptsize{$\pm0.3$}         & 91.7           \\
SWAD (proposed)                        & Opt   & Overfit & Const & 1     & \textbf{87.1} \scriptsize{$\pm0.2$} & \textbf{78.9} \scriptsize{$\pm0.2$} & \textbf{83.0}        & \textbf{97.7} \scriptsize{$\pm0.1$} & \textbf{86.1} \scriptsize{$\pm0.5$} & \textbf{91.9}             \\
\bottomrule
\end{tabular}
\label{table:ablation}
\vspace{-0.5em}
\end{table}

Table~\ref{table:ablation} provides ablative studies on the starting and ending iterations for averaging, the learning rate schedule, and the sampling interval. SWA$_\text{w/ cyclic}$ (SWA in Table~\ref{table:generalization}) and SWA$_\text{w/ constant}$ are vanilla SWAs with fixed sampling positions. 
We also report SWAD by eliminating three factors: the dense sampling strategy, and searching the start iteration, searching the end iteration.
The dense sampling strategy lets SWAD estimate a more accurate approximation of flat minima: showing 0.8pp degeneration in the average out-of-domain accuracy (SWAD$_\text{w/o Dense}$).
When we take an average from $t_s$ to the final iteration, the out-of-domain performance degrades by 0.6pp (SWAD$_\text{w/o Overfit}$). Similarly, a fixed scheduling without the overfit-aware scheduling only shows very marginal improvements from the vanilla SWA (SWAD$_\text{w/o Opt-Overfit}$).
We also evaluate SWAD$_\text{fit-on-val}$ that uses the range achieving the best performances on the validation set, but it becomes overfitted to the validation, results in lower performances than SWAD.
The results demonstrate the benefits of combining ``dense'' and ``overfit-aware'' sampling strategies of SWAD.

\subsection{Exploring the other applications: ImageNet robustness}
\vspace{-.5em}
\begin{table}[h]
\centering
\small
% \caption{\small {\bf Comparison in other robustness benchmarks.} The numbers in ImageNet-C column indicate mCE and the other numbers indicate accuracy. The first column and the others indicate in-domain validation accuracy and the benchmark results, respectively.
% }
\caption{\small {\bf ImageNet robustness benchmarks.} We show the ImageNet generalization performances on ImageNet-C, background challenge (BGC), and ImageNet-R.}
\vspace{.5em}
\renewcommand{\arraystretch}{1.1}
\label{table:exploring_other_applications}
\begin{tabular}{lcccc} 
\toprule
                        % & \multicolumn{4}{c|}{Configuration}   & \multicolumn{3}{c|}{Out-of-domain} & \multicolumn{3}{c}{In-domain}  \\
% & \multicolumn{1}{c|}{In-domain} & \multicolumn{3}{c}{Robustness benchmarks} \\
\textbf{Method}       & ImageNet (\%) ↑ & ImageNet-C (mCE) ↓ & BGC (\%) ↑    & ImageNet-R (\%) ↑  \\
\midrule
ERM         & 76.5            & 57.6               & 8.7           & 36.7               \\
SWA         & 76.9            & 56.8               & 10.9          & 37.5               \\
SWAD (ours) & \textbf{77.0}   & \textbf{55.7}      & \textbf{11.8} & \textbf{38.8}      \\
\bottomrule
\end{tabular}
\end{table}

Since SWAD does not rely on domain labels, it can be applied to other robustness tasks not containing domain labels.
Table \ref{table:exploring_other_applications} show the generalizability of SWAD on ImageNet \cite{russakovsky2015imagenet} and its shifted benchmarks, namely, ImageNet-C~\cite{hendrycks2018benchmarking_robustness_imagenet_c}, ImageNet-R~\cite{hendrycks2020many_face_of_robustness}, and background challenge (BGC)~\cite{xiao2020background_challenge_bgc}.
SWAD consistently improves robustness performances against the ERM baseline and the SWA baseline.
These results support that our method is robustly and widely applicable to improve both in-domain and out-of-domain generalizability. The detailed setup is provided in Appendix B.6.

% SWAD can be applied to any other robustness tasks, thanks to its domain-free property (not utilizing domain label). We verify that on the three ImageNet robustness benchmarks, such as ImageNet-C~\cite{hendrycks2018benchmarking_robustness_imagenet_c}, ImageNet-A~\cite{hendrycks2020many_face_of_robustness}, and background challenge (BGC)~\cite{xiao2020background_challenge_bgc}. 
% In every experiment, SWAD consistently improves robustness performance: 1.9 mean corruption error (mCE) in ImageNet-C, 2.1pp accuracy in ImageNet-R, and 3.1pp accuracy in BGC, compared with ERM baseline. When compared with SWA, there are performance improvements of 0.9 mCE, 1.3pp, and 0.9pp, respectively. These results support that our method is robustly and widely applicable to improve robustness. The detailed experiment setup and results are provided in Appendix.

% \section{Related Work}
% \label{s_related_work}
% \input{2.related_work}

\section{Discussion and Limitations}
\label{s_discussion}
Despite many benefits from SWAD, such as the significant performance improvements, model selection-free property, working plug-and-play manner for various methods, there are some potential limitations.
Here, we discuss the limitations of SWAD for further improvements.

\textbf{Confidence error in Theorem~\ref{thm:gen_bnd_dg}.}
While the confidence error in Theorem~\ref{thm:gen_bnd_dg} tells the effect of $\gamma$ on generalization error bound, there exists a limitation in that the confidence error term shows improper behavior with respect to $\gamma$ if $\gamma$ is close to zero. The behavior we expect is that the confidence error of RRM converges to the confidence error of ERM as $\gamma$ decreases to zero, however, the current theorem does not show such tendency since the confidence bound diverges to infinity when $\gamma$ goes to zero. However, we would like to note that this limitation is not a drawback of RRM, but it is caused by the looseness of the union bound which is a mathematical technique used to derive the confidence error of RRM.
Our RRM formulation has a similarity to previous works~\cite{norton2019diametrical, foret2020sharpness} and we note that the counter-intuitive behavior of the confidence bound and $\gamma$ also appears in \citet{foret2020sharpness}.
% SWAD is an approximation -> there can be a better solution.

\textbf{SWAD is not a perfect flatness-aware optimization method.}
Note that SWAD is not a perfect and theoretically guaranteed solver for flat minima, but a heuristic approximation with empirical benefits.
% SWAD is an approximation to seek flat minima. We do not argue that SWAD is a perfect solution for seeking the flat minima. 
However, even if a better flatness-aware optimization method is proposed, our theoretical contribution still holds: showing the relationship between flat minima and DG.

\textbf{SWAD does not strongly utilize domain-specific information.}
In Theorem~\ref{thm:gen_bnd_dg_final}, the domain generalization gap is bounded by three factors: flat minima, domain discrepancy, and confidence bound. Most of the existing approaches focus on domain discrepancy, reducing the difference between the source domains and the target domain by domain invariant learning~\cite{muandet2013icml_DIFL,ganin2016dann,li2018cdann,bahng2019rebias,zhao2020er_entropy_regularization}.
SWAD focuses on the first factor, the flat minima. While the domain labels are used to construct a mini-batch, SWAD does not strongly utilize domain-specific information. It implies that if one can consider both flatness and domain discrepancy, better domain generalization can be achievable.
Table~\ref{table:swad_combination} gives us a clue: the combination of CORAL (utilizing domain-specific information) and SWAD (seeking flat minima) shows the best performance among all comparison methods.
As a future research direction, we encourage studying a method that can achieve both flat optima and small domain discrepancy.

% \jb{\textbf{Domain generalization in the wild.}
% Even though handling distribution shift is very important in the real world, most of domain generalization algorithm have limitations to use in the wild; they usually restrict model architecture, or painfully slow down training or inference. Utilizing domain label also becomes another constraint, as domain labels often do not exist in the real world. However, SWAD is free from the limitations: SWAD is domain-free, does not impose any constraints on model architecture, and does not slow down inference speed. That is, SWAD is widely, easily, and practically applicable solution in the wild.}

\section{Concluding Remarks}
\label{s_conclusion}
% Contributions:
% 1. We introduce the concept of flatness into domain generalization
% 2. We propose SWAD to capture flatter minima
% 3. SWAD achieves state-of-the-art performances with large margin
% 4. We show even better results by minimizing both robust risk and domain divergence

% \jb{ToDo: Update conclusion also before finalize paper.}

In this paper, we theoretically and empirically demonstrate that domain generalization (DG) is achievable by seeking flat minima.
We propose SWAD that captures flatter minima than the vanilla SWA does.
The extensive experiments on five DG benchmarks show superior performances of SWAD compared with existing DG methods.
In addition, combinations of SWAD and existing DG methods even show better performances than the vanilla SWAD.
We theoretically and empirically observe that seeking flat minima can achieve better generalizability to both in-domain and out-of-domain, while strong in-domain generalization methods without consideration of flatness, \eg, Mixup or CutMix, cannot guarantee to achieve out-of-domain generalizability in both theory and practice.
This study first brings the concept of flatness into DG tasks, and shows strong empirical performances not only in DG but also in ImageNet benchmarks.
% As our theoretical observation supports, seeking flat minima can achieve better DG, while strong in-domain generalization without consideration of flatness, \eg, Mixup or CutMix, does not have any guarantee.
% Comparisons with conventional generalization methods, such as Mixup or CutMix, supports that the significant improvement by SWAD is from seeking flat minima, not from better in-domain generalization performances.
% It also supports our theoretical result that the DG gap is bounded to robust risk and divergence between training and target domains.
% It should be noted that this study is very first work to introduce the concept of flatness into the DG field.
We hope that this study promotes a new research direction of seeking flat minima for domain generalization and other robustness tasks.
% We believe this study is a foundation work in a new research direction of seeking flat minima for domain generalization.
% We believe this study open a new direction, seeking flat minima, in DG field, and hope that this study will serve as cornerstone 

% In addition, we show that combinations of SWAD and other DG methods can achieve state-of-the-art performances on the DG benchmarks.

% \section*{Acknowledgement}
\acksection
NAVER Smart Machine Learning (NSML)~\cite{nsml} and Kakao Brain Cloud platform have been used in experiments. This work was supported by IITP grant funded by the Korea government (MSIT) (No. 2021-0-01341, AI Graduate School Program, CAU).

\bibliography{7.reference}
\bibliographystyle{unsrtnat}

% \input{checklist}

%%%%%%%%%%%%%%%%%%%%%%%%%%%%%%%%%%%%%%%%%%%%%%%%%%%%%%%%%%%%

\newpage
\appendix

% \section{Potential negative societal impacts} 
\section{Potential Societal Impacts} 
In this study, we theoretically and empirically demonstrate that domain generalization (DG) is achievable by seeking flat minima, and propose SWAD to find flat minima.
With SWAD, researchers and developers can make a model robust to domain shift in a real deployment environment, without relying on a task-dependent prior, a modified objective function, or a specific model architecture.
Accordingly, SWAD has potential positive impacts by developing machines less biased towards ethical aspects, as well as potential negative impacts, \eg, improving weapon or surveillance systems under unexpected environment changes.
% Through SWAD, developers expect that the model deployed in the real environment can handle data whose distribution is different with the training distribution.
% propose a widely applicable method, Stochastic Weight Averaging Densely (SWAD), to improve the generalizability for domain shifts. 
% On the other hand, SWAD does not rely on a task-dependent prior, does not enforce a specific model architecture, and does not modify the objective function.
% In addition, SWAD does not require any task-dependent prior, model architectural modification, and specified objectives. Such properties lead to its application without any judgments on its positive or negative impact.  
% That is, it improves generalizability of any deep learning application without discriminating which application has a positive or negative impact.
% It means that one can exploit SWAD to improve the performance of an application that has negative societal impacts, \eg, weapon or surveillance systems.

\section{Implementation Details}

\subsection{Hyperparameters of SWAD}

% the running statistics of the batch normalization layers are not updated during training, and the examples in a single batch are sampled uniformly for each domain.
The evaluation protocol by \citet{gulrajani2020domainbed} is computationally too expensive; it requires about 4,142 models for every DG algorithm.
Hence, we reduce the search space of SWAD for computational efficiency; batch size and learning rate are set to 32 for each domain and 5e-5, respectively. We set dropout probability and weight decay to zero.
% Instead, SWAD uses default value for parameters, namely batch size of 32, learning rate of 5e-5, no dropout, and no weight decay. 
We only search $N_s, N_e$ and $r$. $N_s$ and $N_e$ are searched in \data{PACS} dataset, and the searched values are used for all experiments, while $r$ is searched in [1.2, 1.3] depending on dataset.
% For $N_s$ and $N_e$, we find the best values in \data{PACS} dataset and fix them. Only $r$ is searched in [1.2, 1.3] depending on dataset. 
As a result, we use $N_s=3$, $N_e=6$, and $r=1.2$ for \data{VLCS} and $r=1.3$ for the others.
% We do not use any additional regularization, \eg, weight decay or dropout. Following~\citet{gulrajani2020domainbed}, 
We initialize our model by ImageNet-pretrained ResNet-50 and batch normalization statistics are frozen during training.
% ResNet-50 is initialized with the weights pretrained on ImageNet and the statistics of batch normalization layers are frozen. 
The number of total iterations is $15,000$ for DomainNet and $5,000$ for others, which are sufficient numbers to be converged. 
Finally, we slightly modify the evaluation frequency because it should be set to small enough to detect the moments that the model is optimized and overfitted. However, too small frequency brings large evaluation overhead, thus we compromise between exactness and efficiency: $50$ for \data{VLCS}, $500$ for \data{DomainNet}, and $100$ for others.

\subsection{Hyperparameter search protocol for reproduced results}

\begin{table}[h]
\centering
\caption{\small\textbf{Hyperparameter search space comparison.} U and list indicate Uniform distribution and random choice, respectively.}
\label{table:hp_search_protocol}
\begin{tabular}{llll} 
\toprule
\textbf{Parameter} & \textbf{Default value} & \textbf{DomainBed}            & \textbf{Ours}                  \\
\midrule
batch size         & 32                     & $2^\text{U(3,5.5)}$           & 32                             \\
learning rate      & 5e-5                & $10^\text{U(-5,-3.5)}$        & [1e-5, 3e-5, 5e-5]  \\
ResNet dropout     & 0                      & [0.0, 0.1, 0.5] & [0.0, 0.1, 0.5]  \\
weight decay       & 0                      & $10^\text{U(-6,-2)}$          & [1e-4, 1e-6]                 \\
\bottomrule
\end{tabular}
\end{table}

We evaluate recently proposed methods, SAM~\cite{foret2020sharpness} and Mixstyle~\cite{zhou2021mixstyle}, and compare them with previous results. For a fair comparison, we follow the hyperparameter (HP) search protocol proposed by~\citet{gulrajani2020domainbed}, with a modification to reduce computational resources.
They searched HP by training a total of 58,000 models, corresponding to about 4,142 runs for each algorithm.
It is too much computational burden to train 4,142 models whenever evaluate a new algorithm.
% , causing environmental destruction by increasing CO2 emission.
Therefore, we re-design the HP search protocol efficiently and effectively. In the HP search protocol of DomainBed~\cite{gulrajani2020domainbed}, training domains and algorithm-specific parameters are included in the HP search space, and HP is found for every data split independently by random search. Instead, we do not sample training domains, use HP found in the first data split to the other splits, search algorithm-specific HP independently, and conduct grid search on the more effectively designed HP space as shown in Table~\ref{table:hp_search_protocol}. Through the proposed protocol, we find HP for an algorithm under only 396 runs. Although the number of total runs is reduced to about 10\% ($4,142 \rightarrow 396$), the results of reproduced ERM is improved 0.9pp in average ($63.3\% \rightarrow 64.2\%$). It demonstrates both the effectiveness and the efficiency of our search protocol.

\subsection{Algorithm-specific hyperparameters}

We search the algorithm-specific hyperparameters independently in \data{PACS} dataset, based on the values suggested from each paper. For Mixstyle~\cite{zhou2021mixstyle}, we insert Mixstyle block with domain label after the 1st, 2nd, and 3rd residual blocks with $\alpha=0.1$ and $p=0.5$. We train SAM~\cite{foret2020sharpness} with $\rho=0.05$, and VAT~\cite{miyato2018vat} with $\epsilon=1.0$ and $\alpha=1.0$. In $\Pi$-model~\cite{LaineA17iclr_pi_model}, $w_{max}=1$ is chosen among various $w_{max}$ values such as 1, 10, 100, and 300.
%Note that they used $w_{max}$ from 100 to 300, but the large value degrades the performance in our experiments. 
We use EMA~\cite{polyak1992ema} with $decay=0.99$, Mixup~\cite{zhang2018mixup} with $\alpha=0.2$, and CutMix~\cite{yun2019cutmix} with $\alpha=1.0$ and $p=0.5$.

% \begin{itemize}
%     \item Mixstyle
%     \item SAM
%     \item VAT
%     \item $\Pi$-model
%     \item EMA
%     \item Mixup
%     \item CutMix (beta 1.0, cutmix_prob 0.5)
% \end{itemize}

\subsection{Pseudo code}
\newcommand{\BREAK}{\textbf{break}}
\begin{algorithm}[h]
    % \SetAlgoNoLine
    \DontPrintSemicolon
    
    \caption{Stochastic Weight Averaging Densely}
    \label{alg:swa}

    \KwIn{initial weight $\theta_0$, constant learning rate $\alpha$, tolerance rate $r$, optimum patience $N_s$, overfit patience $N_e$, total number of iterations $T$, training loss $\mathcal E_\text{train}^{(i)}$, validation loss $\mathcal E_\text{val}^{(i)}$}
    \KwOut{averaged weight $\theta^{\text{SWAD}}$ from $t_s$ to $t_e$}
    
    $t_s \leftarrow 0$ \tcp*{start iteration for averaging}
    $t_e \leftarrow T$ \tcp*{end iteration for averaging}
    $l \leftarrow \text{None}$ \tcp*{loss threshold}
    \For{$i \gets 1$ \KwTo $T$}{
        $\theta_i \leftarrow \theta_{i-1} - \alpha \nabla \mathcal{E}_\text{train}^{(i-1)}$  \\
        \If{ $l = $ \textup{None}}{
            \If{$\mathcal{E}_\text{val}^{(i-N_s+1)} = \min_{0 \leq i' < N_s} \mathcal{E}_\text{val}^{(i-i')}$ }{
                $t_s \leftarrow i-N_s+1$ \\
                $l \leftarrow \frac{r}{N_s}\sum_{i'=0}^{N_s-1}\mathcal{E}_\text{val}^{(i-i')}$
            }
        }
        \ElseIf{$l < \min_{0\leq i'< N_e} \mathcal{E}_\text{val}^{(i-i')}$}{
            $t_e \leftarrow i-N_e$ \\
            \BREAK
        }
    }
    $\theta^{\text{SWAD}} \leftarrow \frac{1}{t_e - t_s + 1}\sum^{t_e}_{i'=t_s} \theta^{i'}$
%   \STATE $\theta^{\text{SWAD}} \leftarrow \text{average}(\theta_{t_s:t_e})$
\end{algorithm}

\subsection{Loss surface visualization}

Following~\citet{garipov2018fge}, we choose three model weights $\theta_1, \theta_2, \theta_3$ and define two dimensional weight plane from the weights:

\begin{align}
u=\theta_2 - \theta_1, \qquad v=\frac{(\theta_3 - \theta_1)- \langle \theta_3 - \theta_1, \theta_2 - \theta_1 \rangle}{\|\theta_2 - \theta_1 \|^2 \cdot (\theta_2 - \theta_1)},
\end{align}

% \begin{align}
% u&=\theta_2 - \theta_1, \\
% v&=\frac{(\theta_3 - \theta_1)- \langle \theta_3 - \theta_1, \theta_2 - \theta_1 \rangle}{\|\theta_2 - \theta_1 \|^2 \cdot (\theta_2 - \theta_1)},
% \end{align}

where $\hat u=u/\|u\|$ and $\hat v=v/\|v\|$ are orthonormal bases of the weight plane. Then, we build Cartesian grid near the weights on the plane. For each grid point, we calculate the weight corresponding to the point and compute loss from the weight.
The results are visualized as a contour plot, as shown in Figure 4 in the main text.

\subsection{ImageNet robustness experiments}

We investigate the extensibility of SWAD via three robustness benchmarks (Section 4.4 in the main text), namely ImageNet-C~\cite{hendrycks2018benchmarking_robustness_imagenet_c}, ImageNet-R~\cite{hendrycks2020many_face_of_robustness}, and background challenge (BGC)~\cite{xiao2020background_challenge_bgc}. ImageNet-C measures the robustness against common corruptions such as Gaussian noise, blur, or weather changes. We follow \citet{hendrycks2018benchmarking_robustness_imagenet_c} for measuring mean corruption error (mCE). The lower ImageNet-C implies that the model is robust against corruption noises.
BGC evaluates the robustness against background manipulations as well as the adversarial robustness. The BGC dataset has two groups, foreground and background. BGC manipulates images by combining the foregrounds and backgrounds, and measures whether the model predicts a consistent prediction with any manipulated image.
ImageNet-R tests the robustness against different domains. ImageNet-R collects very different domain images of ImageNet, such as art, cartoons, deviantart, graffiti, embroidery, graphics, origami, paintings, patterns, plastic objects, plush objects, sculptures, sketches, tattoos, toys, and video game renditions. Showing better performances in ImageNet-R leads to the same conclusion as other domain generalization benchmarks.

% As shown in the Table 6 in the main text, our method consistently improves both in-domain and out-of-domain robustness. Compared to the ERM baseline, SWAD improves performances by 0.5pp for in-domain ImageNet validation, 1.9 mCE for ImageNet-C, 2.1pp for ImageNet-R, and 3.1pp for BGC. Compared to SWA, our method shows subtle improvement in the in-domain ImageNet validation set (0.1pp), but it shows significant improvements on the other out-of-domain benchmarks; 0.9 mCE for ImageNet-C, 1.3pp for ImageNet-R, and 0.9pp for BGC. These results are consistent with the results of this study.

% These additional results support that our method is robustly and widely applicable to large-scale applications, such as ImageNet. Furthermore, since SWAD can be easily combined with other methods, the performance can be improved with task-specific methods like CORAL + SWAD.

\paragraph{Experiment details.}
We use ResNet-50 architecture and mostly follow standard training recipes. We use SGD optimizer with momentum of 0.9, base learning rate of 0.1 with linear scaling rule \cite{goyal2017training_imagenet_1hour} and polynomial decay, 5 epochs gradual warmup, batch size of 2048, and total epochs of 90. For SWA, the learning rate is decayed to 1/20 until $80\%$ of training (72 epochs), and the cyclic learning rate with 3 epochs cycle length is used for the left $20\%$ of training. SWAD follows the same learning rate decay until $80\%$ of training, but averages every weight from every iteration after $80\%$ of training with constant learning rate.

\section{Proof of Theorems}
\subsection{Technical Lemmas}

Consider an instance loss function $\ell(y_{1}, y_{2})$
such that $\ell:\mathcal{Y}\times\mathcal{Y} \rightarrow [0,1]$ and $\ell(y_{1},y_{2})=0$ if and only if $y_{1}=y_{2}$.
Then, we can define a functional error as $\mathcal{E}_{\mathcal{P}}(f(\cdot;\theta),h):=\mathbb{E}_{\mathcal{P}}[\ell(f(x;\theta),h(x))]$.
Note that if we set $h$ as a true label function which generates the label of inputs, $y=h(x)$,
then, it becomes a population loss $\mathcal{E}_{\mathcal{P}}(\theta)=\mathcal{E}_{\mathcal{P}}(f(\cdot;\theta),h)$.
Given two distributions, $\mathcal{P}$ and $\mathcal{Q}$, the following lemma shows that the difference between the error with $\mathcal{P}$ and the error with $\mathcal{Q}$ is bounded by the divergence between $\mathcal{P}$ and $\mathcal{Q}$.
\begin{lemma}\label{lem:bnd_dist}
$\left|\mathcal{E}_{\mathcal{P}}(h_{1},h_{2}) - \mathcal{E}_{\mathcal{Q}}(h_{1},h_{2})\right| \leq \frac{1}{2}\mathbf{Div}(\mathcal{P},\mathcal{Q})$
\end{lemma}
\begin{proof}
We employ the same technique in \citet{zhao2018multi_domain_adapt} for our loss function $\ell$.
From the Fubini's theorem, we have,
\begin{align}
    \mathbb{E}_{x\sim\mathcal{P}}[\ell(h_{1}(x),h_{2}(x))] = \int_{0}^{\infty} \mathbb{P}_{\mathcal{P}}\left(\ell(h_{1}(x),h_{2}(x)) > t\right) dt
\end{align}
By using this fact, 
% \begin{align}
%     &\left|\mathbb{E}_{x\sim\mathcal{P}}[l(h_{1}(x),h_{2}(x))]- \mathbb{E}_{x\sim\mathcal{Q}}[l(h_{1}(x),h_{2}(x))]\right|\\
%     &= \left|\int_{\mathcal{X}} l(h_{1}(x),h_{2}(x))\left[d\mathcal{P}(x) - d\mathcal{Q}(x)\right]\right|\\
%     &\leq \int_{0}^{\infty} \left|\mathbb{P}_{\mathcal{P}}\left(l(h_{1}(x),h_{2}(x)) > t\right) -  \mathbb{P}_{\mathcal{Q}}\left(l(h_{1}(x),h_{2}(x)) > t\right) \right|dt\\
%     &\leq M\sup_{t\in[0,M]} \left|\mathbb{P}_{\mathcal{P}}\left(l(h_{1}(x),h_{2}(x)) > t\right) -  \mathbb{P}_{\mathcal{Q}}\left(l(h_{1}(x),h_{2}(x)) > t\right) \right|\\
%     &\leq M\sup_{h_{1},h_{2}}\sup_{t\in[0,M]} \left|\mathbb{P}_{\mathcal{P}}\left(l(h_{1}(x),h_{2}(x)) > t\right) -  \mathbb{P}_{\mathcal{Q}}\left(l(h_{1}(x),h_{2}(x)) > t\right) \right|\\
%     &\leq M\sup_{\bar{h}\in\bar{\mathcal{H}}}\left|\mathbb{P}_{\mathcal{P}}\left(\bar{h}(x)=1\right) -  \mathbb{P}_{\mathcal{Q}}\left(\bar{h}(x)=1\right) \right|\\
%     &\leq M\sup_{A}\left|\mathbb{P}_{\mathcal{P}}\left(A\right) -  \mathbb{P}_{\mathcal{Q}}\left(A\right) \right|
% \end{align}
\begin{align}
    &\left|\mathbb{E}_{x\sim\mathcal{P}}[\ell(h_{1}(x),h_{2}(x))]- \mathbb{E}_{x\sim\mathcal{Q}}[\ell(h_{1}(x),h_{2}(x))]\right|\\
    &= \left|\int_{0}^{\infty} \mathbb{P}_{\mathcal{P}}\left(\ell(h_{1}(x),h_{2}(x)) > t\right) dt - \int_{0}^{\infty} \mathbb{P}_{\mathcal{Q}}\left(\ell(h_{1}(x),h_{2}(x)) > t\right) dt\right|\\
    &\leq \int_{0}^{\infty} \left|\mathbb{P}_{\mathcal{P}}\left(\ell(h_{1}(x),h_{2}(x)) > t\right) -  \mathbb{P}_{\mathcal{Q}}\left(\ell(h_{1}(x),h_{2}(x)) > t\right) \right|dt\\
    &\leq M\sup_{t\in[0,M]} \left|\mathbb{P}_{\mathcal{P}}\left(\ell(h_{1}(x),h_{2}(x)) > t\right) -  \mathbb{P}_{\mathcal{Q}}\left(\ell(h_{1}(x),h_{2}(x)) > t\right) \right|\\
    &\leq M\sup_{h_{1},h_{2}}\sup_{t\in[0,M]} \left|\mathbb{P}_{\mathcal{P}}\left(\ell(h_{1}(x),h_{2}(x)) > t\right) -  \mathbb{P}_{\mathcal{Q}}\left(\ell(h_{1}(x),h_{2}(x)) > t\right) \right|\\
    &\leq M\sup_{\bar{h}\in\bar{\mathcal{H}}}\left|\mathbb{P}_{\mathcal{P}}\left(\bar{h}(x)=1\right) -  \mathbb{P}_{\mathcal{Q}}\left(\bar{h}(x)=1\right) \right|\\
    &\leq M\sup_{A}\left|\mathbb{P}_{\mathcal{P}}\left(A\right) -  \mathbb{P}_{\mathcal{Q}}\left(A\right) \right|
\end{align}
where $\bar{\mathcal{H}}:=\left\{\mathbb{I}[\ell(h(x),h'(x))>t]\middle|h,h'\in\mathcal{H}, t \in [0,M]\right\}$.
\end{proof}

\begin{lemma}\label{lem:gen_bnd_rrm}
Consider a distribution $\mathcal{S}$ on input space and global label function $f:\mathcal{X}\rightarrow\mathcal{Y}$.
Let $\left\{\Theta_{k}\subset\mathbb{R}^{d}, k=1,\cdots,N\right\}$ be a finite cover of a parameter space $\Theta$ which consists of closed balls with radius $\gamma/2$ where $N:=\left\lceil\left(diam(\Theta)/\gamma\right)^{d}\right\rceil$.
Let $\theta_{k}\in\arg\max_{\Theta_{k}\cap\Theta}\mathcal{E}_{\mathcal{S}}(\theta)$ be a local maximum in the $k$-th ball.
Let a VC dimension of $\Theta_{k}$ be $v_{k}$. Then, for any $\theta\in\Theta$,
the following bound holds with probability at least $1-\delta$.
\begin{align}
    \mathcal{E}_{S}(\theta) - \hat{\mathcal{E}}_{S}^{\gamma}(\theta) \leq \max_{k}\sqrt{\frac{\left(v_{k}\left[\ln\left(n/v_{k}\right)+1\right]+\ln\left(N/\delta\right)\right)}{2n}}
\end{align}
where $\hat{\mathcal{E}}_{S}^{\gamma}(\theta_{k})$ is an empirical robust risk with $n$ samples.
\end{lemma}
\begin{proof}
We first show that the following inequality holds for the local maximum of $N$ covers,
\begin{align}
    \mathbb{P}\left(\max_{k}\left[\mathcal{E}_{S}(\theta_{k})-\hat{\mathcal{E}}_{S}(\theta_{k})\right] > \epsilon\right)
    &\leq \sum_{k=1}^{N}\mathbb{P}\left(\mathcal{E}_{S}(\theta_{k})-\hat{\mathcal{E}}_{S}(\theta_{k}) > \epsilon\right)\\
    &\leq \sum_{k=1}^{N}\mathbb{P}\left(\sup_{\theta\in\Theta_{k}}\left[\mathcal{E}_{S}(\theta)-\hat{\mathcal{E}}_{S}(\theta)\right] > \epsilon\right)\\
    &\leq\sum_{k=1}^{N}\left(\frac{en}{v_{k}}\right)^{v_{k}}e^{-2n\epsilon^{2}}.
\end{align}
Now, we introduce a confidence error bound $\epsilon_{k}:=\sqrt{\frac{\left(v_{k}\left[\ln\left(n/v_{k}\right)+1\right]+\ln\left(N/\delta\right)\right)}{2n}}$.
Then, we set $\epsilon:=\max_{k}\epsilon_{k}$. Then, we get,
\begin{align}
    \mathbb{P}\left(\max_{k}\left[\mathcal{E}_{S}(\theta_{k})-\hat{\mathcal{L}}_{S}(\theta_{k})\right] > \epsilon\right)
    &\leq\sum_{k=1}^{N}\left(\frac{en}{v_{k}}\right)^{v_{k}}e^{-2n\epsilon^{2}}\\
    &\leq\sum_{k=1}^{N}\left(\frac{en}{v_{k}}\right)^{v_{k}}e^{-2n\epsilon_{k}^{2}}\\
    &=\sum_{k=1}^{N}\frac{\delta}{N}=\delta,
\end{align}
since $\epsilon>\sqrt{\frac{\left(v_{k}\left[\ln\left(n/v_{k}\right)+1\right]+\ln\left(N/\delta\right)\right)}{2n}}$ for all $k$.
Hence, the inequality holds with probability at least $1-\delta$.

Based on this fact, let us consider the set of events such that $\max_{k}\left[\mathcal{E}_{S}(\theta_{k})-\hat{\mathcal{E}}_{S}(\theta_{k})\right] \leq \epsilon$.
Then, for any $\theta$, there exists $k'$ such that $\theta\in\Theta_{k'}$.
Then, we get
\begin{align}
    \mathcal{E}_{S}(\theta) - \hat{\mathcal{E}}_{S}^{\gamma}(\theta) &\leq \mathcal{E}_{S}(\theta) - \hat{\mathcal{E}}_{S}(\theta_{k'})\\
    &\leq \mathcal{E}_{S}(\theta) - \mathcal{E}_{S}(\theta_{k'}) + \epsilon\\
    &\leq \mathcal{E}_{S}(\theta_{k'}) - \mathcal{E}_{S}(\theta_{k'}) + \epsilon = \epsilon,
\end{align}
where the second inequality holds since $\mathcal{E}_{S}(\theta_{k'})-\hat{\mathcal{E}}_{S}(\theta_{k'})\leq\max_{k}\left[\mathcal{E}_{S}(\theta_{k})-\hat{\mathcal{E}}_{S}(\theta_{k})\right] \leq \epsilon$ and the final inequality holds since $\theta_{k'}$ is the local maximum in $\Theta_{k'}$.
In this regards, we know that $\max_{k}\left[\mathcal{E}_{S}(\theta_{k})-\hat{\mathcal{E}}_{S}(\theta_{k})\right] \leq \epsilon$ implies $\mathcal{E}_{S}(\theta) - \hat{\mathcal{E}}_{S}^{\gamma}(\theta) \leq \epsilon$.
Consequently, $\mathcal{E}_{S}(\theta) - \hat{\mathcal{E}}_{S}^{\gamma}(\theta) \leq \epsilon$ holds with probability at least $1-\delta$.
\end{proof}

\subsection{Proof of Theorem 1}
% \begin{theorem}\label{thm:gen_bnd_dg}
% Consider a set of $N$ covers $\{\Theta_{k}\}_{k=1}^{N}$ such that the parameter space $\Theta \subset \cup_{k}^{N} \Theta_{k}$ where $diam(\Theta):=\sup_{\theta,\theta'\in \Theta}\|\theta-\theta'\|_{2}$, $N:=\left\lceil\left(diam(\Theta)/\gamma\right)^{d}\right\rceil$ and $d$ is dimension of $\Theta$.
% Let $v_{k}$ be a VC dimension of each $\Theta_{k}$.
% Then, for any $\theta\in\Theta$, the following bound holds with probability at least $1-\delta$,
% \begin{equation}
% \label{eq:thm1_bound}
% \mathcal{E}_{\mathcal{T}}(\theta) < \hat{\mathcal{E}}_{\mathcal{D}}^{\gamma}(\theta) +\frac{1}{2I}\sum_{i=1}^{I}\mathbf{Div}(\mathcal{D}_{i},\mathcal{T})+ \max_{k\in[1,N]} \sqrt{\frac{v_{k}\ln\left(m/v_{k}\right)+\ln(N/\delta)}{m}},
% \end{equation}
% where $m = nI$ is the number of the training samples and $\mathbf{Div}(\mathcal{D}_{i},\mathcal{T}):=2\sup_{A}|\mathbb{P}_{\mathcal{D}_{i}}(A)-\mathbb{P}_{\mathcal{T}}(A)|$ is a divergence between two distributions.
% \end{theorem}
\begin{proof}
The proof consists of two parts.
First, we show that the following inequality holds with high probability.
$$
\mathcal{E}_{\mathcal{T}}(\theta) \leq \hat{\mathcal{E}}_{\mathcal{S}}^{\gamma}(\theta) + \frac{1}{2}\mathbf{Div}\left(\mathcal{S},\mathcal{T}\right)+ \max_{k}\sqrt{\frac{\left(v_{k}\left[\ln\left(n/v_{k}\right)+1\right]+\ln\left(N/\delta\right)\right)}{2n}}.
$$
Then, secondly, we apply the inequality for multiple source domains.

The first part can be proven by simply combining Lemma \ref{lem:bnd_dist} and Lemma \ref{lem:gen_bnd_rrm}.
Then, we get,
\begin{align}
\mathcal{E}_{\mathcal{T}}(\theta) &\leq \mathcal{E}_{\mathcal{S}}(\theta) + \frac{1}{2}\mathbf{Div}\left(\mathcal{S},\mathcal{T}\right)\\
&\leq \hat{\mathcal{E}}^{\gamma}_{S}(\theta) + \frac{1}{2}\mathbf{Div}\left(\mathcal{S},\mathcal{T}\right) + \max_{k}\sqrt{\frac{\left(v_{k}\left[\ln\left(n/v_{k}\right)+1\right]+\ln\left(N/\delta\right)\right)}{2n}}
\end{align}
where $\mathbf{Div}\left(\mathcal{S},\mathcal{T}\right)$ is a divergence between $\mathcal{S}$ and $\mathcal{T}$.

For the second part, we set $\mathcal{D}:=\sum_{i=1}^{I}\mathcal{D}_{i}/I$ which is a mixture of source distributions.
Then, by applying $\mathcal{D}$ to the first part, we obtain the following inequality,
\begin{align}
\mathcal{E}_{\mathcal{T}}(\theta) &\leq \hat{\mathcal{E}}^{\gamma}_{\mathcal{D}}(\theta) + \frac{1}{2}\mathbf{Div}\left(\mathcal{D},\mathcal{T}\right) + \max_{k}\sqrt{\frac{\left(v_{k}\left[\ln\left(In/v_{k}\right)+1\right]+\ln\left(N/\delta\right)\right)}{2In}}\\
&\leq \hat{\mathcal{E}}^{\gamma}_{\mathcal{D}}(\theta) + \frac{1}{2I}\sum_{i=1}^{I}\mathbf{Div}\left(\mathcal{D}_{i},\mathcal{T}\right) + \max_{k}\sqrt{\frac{\left(v_{k}\left[\ln\left(In/v_{k}\right)+1\right]+\ln\left(N/\delta\right)\right)}{2In}} 
\end{align}
where the total number of training data set is $In$ and, for the second inequality, we use the fact that $\frac{1}{2}\mathbf{Div}\left(\mathcal{D},\mathcal{T}\right)\leq \frac{1}{2I}\sum_{i=1}^{I}\mathbf{Div}\left(\mathcal{D}_{i},\mathcal{T}\right)$, which has been proven in \cite{zhao2018multi_domain_adapt}.
\end{proof}

\subsection{Proof of Theorem 2}
% \begin{theorem}\label{thm:gen_bnd_dg_final}
% Let $\hat{\theta}^{\gamma}$ denote the optimal solution of the RRM, \ie, $\hat{\theta}^{\gamma}:=\arg\min_{\theta}\hat{\mathcal{E}}^{\gamma}_{\mathcal{D}}(\theta)$, and let $v$ be a VC dimension of the parameter space $\Theta$.
% Then, the gap between the optimal test loss, $\min_{\theta'}\mathcal{E}_{\mathcal{T}}\left(\theta'\right)$, and the test loss of $\hat{\theta}^{\gamma}$, $\mathcal{E}_{\mathcal{T}}(\hat{\theta}^{\gamma})$, has the following bound with probability at least $1-\delta$.
% \begin{align}
% \begin{split}
%     \mathcal{E}_{\mathcal{T}}(\hat{\theta}^{\gamma}) - \min_{\theta'}\mathcal{E}_{\mathcal{T}}&\left(\theta'\right) \quad \leq \quad \hat{\mathcal{E}}_{\mathcal{D}}^{\gamma}(\hat{\theta}^{\gamma}) - \min_{\theta''}\hat{\mathcal{E}}_{\mathcal{D}}(\theta'') + \frac{1}{I}\sum_{i=1}^{I}\mathbf{Div}(\mathcal{D}_{i},\mathcal{T})\\
%     &+ \max_{k\in[1,N]} \sqrt{\frac{v_{k}\ln\left(m/v_{k}\right)+\ln\left(2N/\delta\right)}{m}} + \sqrt{\frac{v\ln\left(m/v\right) + \ln\left(2/\delta\right)}{m}}
% \end{split}
% \end{align}
% \end{theorem}

\begin{proof}
First, let $\bar{\theta}\in\arg\max_{\theta\in\Theta}\mathcal{E}_{\mathcal{T}}(\theta)$. Then, from generalization error bound of $\mathcal{E}_{\mathcal{D}}(\bar{\theta})$, the following inequality holds with probability at most $\frac{\delta}{2}$,
\begin{align}
\hat{\mathcal{E}}_{\mathcal{D}}(\bar{\theta}) - \mathcal{E}_{\mathcal{D}}(\bar{\theta}) > \sqrt{\frac{v\ln\left(In/v\right) + \ln\left(2/\delta\right)}{In}},
\end{align}
where $v$ is a VC dimension of $\Theta$.
Furthermore, from Theorem 1,
 we have the following inequality with probability at most $\frac{\delta}{2}$,
\begin{align}
\mathcal{E}_{\mathcal{T}}(\hat{\theta}^{\gamma}) > \mathcal{E}_{\mathcal{D}}^{\gamma}(\hat{\theta}^{\gamma}) + \frac{1}{2}\mathbf{Div}(\mathcal{D},\mathcal{T}) + \max_{k\in[1,N]} \sqrt{\frac{v_{k}\ln\left(In/v_{k}\right)+\ln(2N/\delta)}{In}}.
\end{align}

Finally, let us consider the set of event such that $\hat{\mathcal{E}}_{\mathcal{D}}(\bar{\theta}) - \mathcal{E}_{\mathcal{D}}(\bar{\theta}) \leq \sqrt{\frac{v\ln\left(In/v\right) + \ln\left(2/\delta\right)}{In}}$ and $\mathcal{E}_{\mathcal{T}}(\hat{\theta}^{\gamma}) \leq \mathcal{E}_{\mathcal{D}}^{\gamma}(\hat{\theta}^{\gamma}) + \frac{1}{2}\mathbf{Div}(\mathcal{D},\mathcal{T}) + \max_{k\in[1,N]} \sqrt{\frac{v_{k}\ln\left(In/v_{k}\right)+\ln(2N/\delta)}{In}}$ whose probability is at least greater than $1-\delta$.
Then, under this set of event, we have,
\begin{align}
    \min_{\theta'}\hat{\mathcal{E}}_{\mathcal{D}}(\theta') &\leq \hat{\mathcal{E}}_{\mathcal{D}}(\bar{\theta})\leq \mathcal{E}_{\mathcal{D}}(\bar{\theta}) +\sqrt{\frac{v\ln\left(In/v\right) + \ln\left(2/\delta\right)}{In}}\\
    &\leq \mathcal{E}_{\mathcal{T}}(\bar{\theta})+ \frac{1}{2}\mathbf{Div}(\mathcal{D}, \mathcal{T}) +\sqrt{\frac{v\ln\left(In/v\right) + \ln\left(2/\delta\right)}{In}}\\
    &\leq \min_{\theta'}\mathcal{E}_{\mathcal{T}}\left(\theta'\right) + \frac{1}{2}\mathbf{Div}(\mathcal{D}, \mathcal{T}) + \sqrt{\frac{v\ln\left(In/v\right) + \ln\left(2/\delta\right)}{In}}
\end{align}
Consequently, we have,
\begin{align}
% \begin{split}
\mathcal{E}_{\mathcal{T}}(\hat{\theta}^{\gamma})& - \min_{\theta'}\mathcal{E}_{\mathcal{T}}\left(\theta'\right) \nonumber \\
&\leq \mathcal{E}_{\mathcal{D}}^{\gamma}(\hat{\theta}^{\gamma}) - \min_{\theta'}\hat{\mathcal{E}}_{\mathcal{D}}\left(\theta'\right) + \mathbf{Div}(\mathcal{D}, \mathcal{T}) + \max_{k\in[1,N]} \sqrt{\frac{v_{k}\ln\left(In/v_{k}\right)+\ln(2N/\delta)}{In}} \nonumber \\
&+ \sqrt{\frac{v\ln\left(In/v\right) + \ln\left(2/\delta\right)}{In}}\\
&\leq \mathcal{E}_{\mathcal{D}}^{\gamma}(\hat{\theta}^{\gamma}) - \min_{\theta'}\hat{\mathcal{E}}_{\mathcal{D}}\left(\theta'\right) + \frac{1}{I}\sum_{i=1}^{I}\mathbf{Div}(\mathcal{D}_{i}, \mathcal{T}) + \max_{k\in[1,N]} \sqrt{\frac{v_{k}\ln\left(In/v_{k}\right)+\ln(2N/\delta)}{In}} \nonumber \\
&+ \sqrt{\frac{v\ln\left(In/v\right) + \ln\left(2/\delta\right)}{In}}
% \end{split}
\end{align}
\end{proof}

\section{Additional Experiments}

\subsection{Comparison of flatness-aware solvers}

\begin{table}[h]
\centering
\small
\caption{\small {\bf Flatness-aware solvers comparison.} SWAs collect $10$ weights from the last $20\%$ of training.
}
\vspace{.5em}
\renewcommand{\arraystretch}{1.1}
\label{table:comparison_flatness_solvers}
\begin{tabular}{lcccccc} 
\toprule
\textbf{Algorithm} & \data{PACS} & \data{VLCS} & \data{OfficeHome} & \data{TerraInc} & \data{DomainNet} & Avg.  \\
\midrule
ERM (baseline)                         & 85.5 \scriptsize$\pm0.2$ & 77.5 \scriptsize$\pm0.4$ & 66.5 \scriptsize$\pm0.3$ & 46.1 \scriptsize$\pm1.8$ & 40.9 \scriptsize$\pm0.1$ & 63.3  \\
SAM                                    & 85.8 \scriptsize$\pm0.2$ & \textbf{79.4} \scriptsize$\pm0.1$ & 69.6 \scriptsize$\pm0.1$ & 43.3 \scriptsize$\pm0.7$ & 44.3 \scriptsize$\pm0.0$ & 64.5  \\
SWA$_\text{w/ cyclic}$                                    & 87.1 \scriptsize$\pm0.1$ & 76.5 \scriptsize$\pm0.2$ & 68.5 \scriptsize$\pm0.2$       & 49.6 \scriptsize$\pm1.0$     & 45.6 \scriptsize$\pm0.0$      & 65.5  \\
SWA$_\text{w/ const}$                                    & 86.9 \scriptsize$\pm0.2$ & 76.6 \scriptsize$\pm0.1$ & 69.3 \scriptsize$\pm0.3$       & 49.2 \scriptsize$\pm1.2$     & 45.9 \scriptsize$\pm0.0$      & 65.6  \\
SWAD                            & \textbf{88.1} \scriptsize$\pm0.1$ & 79.1 \scriptsize$\pm0.1$ & \textbf{70.6} \scriptsize$\pm0.2$       & \textbf{50.0} \scriptsize$\pm0.3$     & \textbf{46.5} \scriptsize$\pm0.1$      & \textbf{66.9}  \\
\bottomrule
\end{tabular}
\end{table}

% \begin{table}[h]
% \centering
% \small
% \caption{{\bf Comparison between SWA and SWAD.} SWA adopts cyclic learning rate schedule in the last $20\%$ of training. The cycle length is set to $2\%$ of total iterations to collect 10 weights for averaging. For each dataset, standard error is reported from three trials.
% }
% \vspace{.5em}
% \renewcommand{\arraystretch}{1.1}
% \label{table:comparison_swa_swad}
% \begin{tabular}{lcccccc} 
% \toprule
% \textbf{Algorithm} & \data{PACS} & \data{VLCS} & \data{OfficeHome} & \data{TerraInc} & \data{DomainNet} & Avg. ($\Delta$)  \\
% \midrule
% SWA                                    & 87.1 \scriptsize$\pm0.1$ & 76.5 \scriptsize$\pm0.2$ & 68.5 \scriptsize$\pm0.2$       & 49.6 \scriptsize$\pm1.0$     & 45.6 \scriptsize$\pm0.0$      & 65.5  \\
% SWAD (ours)                            & 88.1 \scriptsize$\pm0.1$ & 79.1 \scriptsize$\pm0.1$ & 70.6 \scriptsize$\pm0.2$       & 50.0 \scriptsize$\pm0.3$     & 46.5 \scriptsize$\pm0.1$      & 66.9 (+1.4)  \\
% \bottomrule
% \end{tabular}
% \end{table}

% Table~\ref{table:comparison_swa_swad} shows the improvements from dense and overfit-aware sampling strategies compared with the baseline method, SWA.

Interestingly, the average performance ranking of flatness-aware solvers is the same as the results of the local flatness test (See Figure 3 in the main text). In both experiments, SWAD performs best, followed by SWAs, SAM, and ERM.
It is another evidence of our claim that domain generalization is achievable by seeking flat minima.

On the other hand, comparing SWAs and SWAD demonstrates the effectiveness of the proposed dense and overfit-aware sampling strategy.
SWAD improves average performance up to 1.4pp, and surpasses both SWAs on every benchmark.

% that proposed dense and overfit-aware sampling strategy remarkably improves domain generalization performance. Furthermore, this experiment also shows SWAD provides robustness to dataset splitting (different trials), as result in lower standard error than SWA.

\section{Full Results}

In this section, we show detailed results of Table 2 in the main text.
$\dagger$ and $\ddagger$ indicate results from DomainBed's and our HP search protocols, respectively. Standard errors are reported from three trials, if available.
% Note that $\ddagger$ results use augmentation for validation set following DomainBed~\cite{gulrajani2020domainbed}. 
% SWAD results correpond to Table 2 in the main text, which only search HP for tolerance ratio in [1.2, 1.3] and use default HP for the others. 
% SWAD$^\ddagger$ indicate the results from the proposed HP search protocol, where search space of lr is [3e-5, 5e-5], of dropout is [0.0, 0.1], of weight decay is [1e-4, 1e-6], and of tolerance ratio is [1.2, 1.3]. 

\subsection{PACS}
\begin{table}[H]
\centering
\small
\renewcommand{\arraystretch}{1.1}
\caption{\small\textbf{Out-of-domain accuracies (\%) on} \data{PACS}\textbf{.}}
\begin{tabular}{lllll|c}
\toprule
\textbf{Algorithm} & \textbf{A} & \textbf{C} & \textbf{P} & \textbf{S} & \textbf{Avg} \\
\midrule
CDANN$^\dagger$ & 84.6 \scriptsize$\pm1.8$ & 75.5 \scriptsize$\pm0.9$ & 96.8 \scriptsize$\pm0.3$ & 73.5 \scriptsize$\pm0.6$ & 82.6 \\
MASF & 82.9 & 80.5 & 95.0 & 72.3 & 82.7 \\
DMG & 82.6 & 78.1 & 94.5 & 78.3 & 83.4 \\
IRM$^\dagger$ & 84.8 \scriptsize$\pm1.3$ & 76.4 \scriptsize$\pm1.1$ & 96.7 \scriptsize$\pm0.6$ & 76.1 \scriptsize$\pm1.0$ & 83.5 \\
MetaReg & 87.2 & 79.2 & 97.6 & 70.3 & 83.6 \\
DANN$^\dagger$ & 86.4 \scriptsize$\pm0.8$ & 77.4 \scriptsize$\pm0.8$ & 97.3 \scriptsize$\pm0.4$ & 73.5 \scriptsize$\pm2.3$ & 83.7 \\
ERM$^\ddagger$ & 85.7 \scriptsize$\pm0.6$ & 77.1 \scriptsize$\pm0.8$ & 97.4 \scriptsize$\pm0.4$ & 76.6 \scriptsize$\pm0.7$ & 84.2 \\
GroupDRO$^\dagger$ & 83.5 \scriptsize$\pm0.9$ & 79.1 \scriptsize$\pm0.6$ & 96.7 \scriptsize$\pm0.3$ & 78.3 \scriptsize$\pm2.0$ & 84.4 \\
MTL$^\dagger$ & 87.5 \scriptsize$\pm0.8$ & 77.1 \scriptsize$\pm0.5$ & 96.4 \scriptsize$\pm0.8$ & 77.3 \scriptsize$\pm1.8$ & 84.6 \\
I-Mixup & 86.1 \scriptsize$\pm0.5$ & 78.9 \scriptsize$\pm0.8$ & 97.6 \scriptsize$\pm0.1$ & 75.8 \scriptsize$\pm1.8$ & 84.6 \\
MMD$^\dagger$ & 86.1 \scriptsize$\pm1.4$ & 79.4 \scriptsize$\pm0.9$ & 96.6 \scriptsize$\pm0.2$ & 76.5 \scriptsize$\pm0.5$ & 84.7 \\
VREx$^\dagger$ & 86.0 \scriptsize$\pm1.6$ & 79.1 \scriptsize$\pm0.6$ & 96.9 \scriptsize$\pm0.5$ & 77.7 \scriptsize$\pm1.7$ & 84.9 \\
MLDG$^\dagger$ & 85.5 \scriptsize$\pm1.4$ & 80.1 \scriptsize$\pm1.7$ & 97.4 \scriptsize$\pm0.3$ & 76.6 \scriptsize$\pm1.1$ & 84.9 \\
ARM$^\dagger$ & 86.8 \scriptsize$\pm0.6$ & 76.8 \scriptsize$\pm0.5$ & 97.4 \scriptsize$\pm0.3$ & 79.3 \scriptsize$\pm1.2$ & 85.1 \\
RSC$^\dagger$ & 85.4 \scriptsize$\pm0.8$ & 79.7 \scriptsize$\pm1.8$ & 97.6 \scriptsize$\pm0.3$ & 78.2 \scriptsize$\pm1.2$ & 85.2 \\
Mixstyle$^\ddagger$ & 86.8 \scriptsize$\pm0.5$ & 79.0 \scriptsize$\pm1.4$ & 96.6 \scriptsize$\pm0.1$ & 78.5 \scriptsize$\pm2.3$ & 85.2 \\
ER & 87.5 & 79.3 & \textbf{98.3} & 76.3 & 85.3 \\
pAdaIN & 85.8 & 81.1 & 97.2 & 77.4 & 85.4 \\
ERM$^\dagger$ & 84.7 \scriptsize$\pm0.4$ & 80.8 \scriptsize$\pm0.6$ & 97.2 \scriptsize$\pm0.3$ & 79.3 \scriptsize$\pm1.0$ & 85.5 \\
EISNet & 86.6 & 81.5 & 97.1 & 78.1 & 85.8 \\
CORAL$^\dagger$ & 88.3 \scriptsize$\pm0.2$ & 80.0 \scriptsize$\pm0.5$ & 97.5 \scriptsize$\pm0.3$ & 78.8 \scriptsize$\pm1.3$ & 86.2 \\
SagNet$^\dagger$ & 87.4 \scriptsize$\pm1.0$ & 80.7 \scriptsize$\pm0.6$ & 97.1 \scriptsize$\pm0.1$ & 80.0 \scriptsize$\pm0.4$ & 86.3 \\
DSON & 87.0 & 80.6 & 96.0 & \textbf{82.9} & 86.6 \\
\midrule
Ours & \textbf{89.3} \scriptsize$\pm0.2$ & \textbf{83.4} \scriptsize$\pm0.6$ & 97.3 \scriptsize$\pm0.3$ & 82.5 \scriptsize$\pm0.5$ & \textbf{88.1} \\
\bottomrule
\end{tabular}
\end{table}

\subsection{VLCS}
\begin{table}[H]
\centering
\small
\renewcommand{\arraystretch}{1.1}
\caption{\small\textbf{Out-of-domain accuracies (\%) on} \data{VLCS}\textbf{.}}
\begin{tabular}{lllll|c}
\toprule
\textbf{Algorithm} & \textbf{C} & \textbf{L} & \textbf{S} & \textbf{V} & \textbf{Avg} \\
\midrule
GroupDRO$^\dagger$ & 97.3 \scriptsize$\pm0.3$ & 63.4 \scriptsize$\pm0.9$ & 69.5 \scriptsize$\pm0.8$ & 76.7 \scriptsize$\pm0.7$ & 76.7 \\
RSC$^\dagger$ & 97.9 \scriptsize$\pm0.1$ & 62.5 \scriptsize$\pm0.7$ & 72.3 \scriptsize$\pm1.2$ & 75.6 \scriptsize$\pm0.8$ & 77.1 \\
MLDG$^\dagger$ & 97.4 \scriptsize$\pm0.2$ & 65.2 \scriptsize$\pm0.7$ & 71.0 \scriptsize$\pm1.4$ & 75.3 \scriptsize$\pm1.0$ & 77.2 \\
MTL$^\dagger$ & 97.8 \scriptsize$\pm0.4$ & 64.3 \scriptsize$\pm0.3$ & 71.5 \scriptsize$\pm0.7$ & 75.3 \scriptsize$\pm1.7$ & 77.2 \\
ERM$^\ddagger$ & 98.0 \scriptsize$\pm0.3$ & 64.7 \scriptsize$\pm1.2$ & 71.4 \scriptsize$\pm1.2$ & 75.2 \scriptsize$\pm1.6$ & 77.3 \\
I-Mixup & 98.3 \scriptsize$\pm0.6$ & 64.8 \scriptsize$\pm1.0$ & 72.1 \scriptsize$\pm0.5$ & 74.3 \scriptsize$\pm0.8$ & 77.4 \\
ERM$^\dagger$ & 97.7 \scriptsize$\pm0.4$ & 64.3 \scriptsize$\pm0.9$ & 73.4 \scriptsize$\pm0.5$ & 74.6 \scriptsize$\pm1.3$ & 77.5 \\
MMD$^\dagger$ & 97.7 \scriptsize$\pm0.1$ & 64.0 \scriptsize$\pm1.1$ & 72.8 \scriptsize$\pm0.2$ & 75.3 \scriptsize$\pm3.3$ & 77.5 \\
CDANN$^\dagger$ & 97.1 \scriptsize$\pm0.3$ & 65.1 \scriptsize$\pm1.2$ & 70.7 \scriptsize$\pm0.8$ & 77.1 \scriptsize$\pm1.5$ & 77.5 \\
ARM$^\dagger$ & 98.7 \scriptsize$\pm0.2$ & 63.6 \scriptsize$\pm0.7$ & 71.3 \scriptsize$\pm1.2$ & 76.7 \scriptsize$\pm0.6$ & 77.6 \\
SagNet$^\dagger$ & 97.9 \scriptsize$\pm0.4$ & 64.5 \scriptsize$\pm0.5$ & 71.4 \scriptsize$\pm1.3$ & 77.5 \scriptsize$\pm0.5$ & 77.8 \\
Mixstyle$^\ddagger$ & 98.6 \scriptsize$\pm0.3$ & 64.5 \scriptsize$\pm1.1$ & 72.6 \scriptsize$\pm0.5$ & 75.7 \scriptsize$\pm1.7$ & 77.9 \\
VREx$^\dagger$ & 98.4 \scriptsize$\pm0.3$ & 64.4 \scriptsize$\pm1.4$ & 74.1 \scriptsize$\pm0.4$ & 76.2 \scriptsize$\pm1.3$ & 78.3 \\
IRM$^\dagger$ & 98.6 \scriptsize$\pm0.1$ & 64.9 \scriptsize$\pm0.9$ & 73.4 \scriptsize$\pm0.6$ & 77.3 \scriptsize$\pm0.9$ & 78.6 \\
DANN$^\dagger$ & \textbf{99.0} \scriptsize$\pm0.3$ & 65.1 \scriptsize$\pm1.4$ & 73.1 \scriptsize$\pm0.3$ & 77.2 \scriptsize$\pm0.6$ & 78.6 \\
CORAL$^\dagger$ & 98.3 \scriptsize$\pm0.1$ & \textbf{66.1} \scriptsize$\pm1.2$ & 73.4 \scriptsize$\pm0.3$ & 77.5 \scriptsize$\pm1.2$ & 78.8 \\
\midrule
Ours & 98.8 \scriptsize$\pm0.1$ & 63.3 \scriptsize$\pm0.3$ & \textbf{75.3} \scriptsize$\pm0.5$ & \textbf{79.2} \scriptsize$\pm0.6$ & \textbf{79.1} \\
\bottomrule
\end{tabular}
\end{table}

\subsection{OfficeHome}
\begin{table}[H]
\centering
\small
\renewcommand{\arraystretch}{1.1}
\caption{\small\textbf{Out-of-domain accuracies (\%) on} \data{OfficeHome}\textbf{.}}
\begin{tabular}{lllll|c}
\toprule
\textbf{Algorithm} & \textbf{A} & \textbf{C} & \textbf{P} & \textbf{R} & \textbf{Avg} \\
\midrule
Mixstyle$^\ddagger$ & 51.1 \scriptsize$\pm0.3$ & 53.2 \scriptsize$\pm0.4$ & 68.2 \scriptsize$\pm0.7$ & 69.2 \scriptsize$\pm0.6$ & 60.4 \\
IRM$^\dagger$ & 58.9 \scriptsize$\pm2.3$ & 52.2 \scriptsize$\pm1.6$ & 72.1 \scriptsize$\pm2.9$ & 74.0 \scriptsize$\pm2.5$ & 64.3 \\
ARM$^\dagger$ & 58.9 \scriptsize$\pm0.8$ & 51.0 \scriptsize$\pm0.5$ & 74.1 \scriptsize$\pm0.1$ & 75.2 \scriptsize$\pm0.3$ & 64.8 \\
RSC$^\dagger$ & 60.7 \scriptsize$\pm1.4$ & 51.4 \scriptsize$\pm0.3$ & 74.8 \scriptsize$\pm1.1$ & 75.1 \scriptsize$\pm1.3$ & 65.5 \\
CDANN$^\dagger$ & 61.5 \scriptsize$\pm1.4$ & 50.4 \scriptsize$\pm2.4$ & 74.4 \scriptsize$\pm0.9$ & 76.6 \scriptsize$\pm0.8$ & 65.7 \\
DANN$^\dagger$ & 59.9 \scriptsize$\pm1.3$ & 53.0 \scriptsize$\pm0.3$ & 73.6 \scriptsize$\pm0.7$ & 76.9 \scriptsize$\pm0.5$ & 65.9 \\
GroupDRO$^\dagger$ & 60.4 \scriptsize$\pm0.7$ & 52.7 \scriptsize$\pm1.0$ & 75.0 \scriptsize$\pm0.7$ & 76.0 \scriptsize$\pm0.7$ & 66.0 \\
MMD$^\dagger$ & 60.4 \scriptsize$\pm0.2$ & 53.3 \scriptsize$\pm0.3$ & 74.3 \scriptsize$\pm0.1$ & 77.4 \scriptsize$\pm0.6$ & 66.4 \\
MTL$^\dagger$ & 61.5 \scriptsize$\pm0.7$ & 52.4 \scriptsize$\pm0.6$ & 74.9 \scriptsize$\pm0.4$ & 76.8 \scriptsize$\pm0.4$ & 66.4 \\
VREx$^\dagger$ & 60.7 \scriptsize$\pm0.9$ & 53.0 \scriptsize$\pm0.9$ & 75.3 \scriptsize$\pm0.1$ & 76.6 \scriptsize$\pm0.5$ & 66.4 \\
ERM$^\dagger$ & 61.3 \scriptsize$\pm0.7$ & 52.4 \scriptsize$\pm0.3$ & 75.8 \scriptsize$\pm0.1$ & 76.6 \scriptsize$\pm0.3$ & 66.5 \\
MLDG$^\dagger$ & 61.5 \scriptsize$\pm0.9$ & 53.2 \scriptsize$\pm0.6$ & 75.0 \scriptsize$\pm1.2$ & 77.5 \scriptsize$\pm0.4$ & 66.8 \\
ERM$^\ddagger$ & 63.1 \scriptsize$\pm0.3$ & 51.9 \scriptsize$\pm0.4$ & 77.2 \scriptsize$\pm0.5$ & 78.1 \scriptsize$\pm0.2$ & 67.6 \\
I-Mixup & 62.4 \scriptsize$\pm0.8$ & 54.8 \scriptsize$\pm0.6$ & 76.9 \scriptsize$\pm0.3$ & 78.3 \scriptsize$\pm0.2$ & 68.1 \\
SagNet$^\dagger$ & 63.4 \scriptsize$\pm0.2$ & 54.8 \scriptsize$\pm0.4$ & 75.8 \scriptsize$\pm0.4$ & 78.3 \scriptsize$\pm0.3$ & 68.1 \\
CORAL$^\dagger$ & 65.3 \scriptsize$\pm0.4$ & 54.4 \scriptsize$\pm0.5$ & 76.5 \scriptsize$\pm0.1$ & 78.4 \scriptsize$\pm0.5$ & 68.7 \\
\midrule
Ours & \textbf{66.1} \scriptsize$\pm0.4$ & \textbf{57.7} \scriptsize$\pm0.4$ & \textbf{78.4} \scriptsize$\pm0.1$ & \textbf{80.2} \scriptsize$\pm0.2$ & \textbf{70.6} \\
\bottomrule
\end{tabular}
\end{table}

\subsection{TerraIncognita}
\begin{table}[H]
\centering
\small
\renewcommand{\arraystretch}{1.1}
\caption{\small\textbf{Out-of-domain accuracies (\%) on} \data{TerraIncognita}\textbf{.}}
\begin{tabular}{lllll|c}
\toprule
\textbf{Algorithm} & \textbf{L100} & \textbf{L38} & \textbf{L43} & \textbf{L46} & \textbf{Avg} \\
\midrule
MMD$^\dagger$ & 41.9 \scriptsize$\pm3.0$ & 34.8 \scriptsize$\pm1.0$ & 57.0 \scriptsize$\pm1.9$ & 35.2 \scriptsize$\pm1.8$ & 42.2 \\
GroupDRO$^\dagger$ & 41.2 \scriptsize$\pm0.7$ & 38.6 \scriptsize$\pm2.1$ & 56.7 \scriptsize$\pm0.9$ & 36.4 \scriptsize$\pm2.1$ & 43.2 \\
Mixstyle$^\ddagger$ & 54.3 \scriptsize$\pm1.1$ & 34.1 \scriptsize$\pm1.1$ & 55.9 \scriptsize$\pm1.1$ & 31.7 \scriptsize$\pm2.1$ & 44.0 \\
ARM$^\dagger$ & 49.3 \scriptsize$\pm0.7$ & 38.3 \scriptsize$\pm2.4$ & 55.8 \scriptsize$\pm0.8$ & 38.7 \scriptsize$\pm1.3$ & 45.5 \\
MTL$^\dagger$ & 49.3 \scriptsize$\pm1.2$ & 39.6 \scriptsize$\pm6.3$ & 55.6 \scriptsize$\pm1.1$ & 37.8 \scriptsize$\pm0.8$ & 45.6 \\
CDANN$^\dagger$ & 47.0 \scriptsize$\pm1.9$ & 41.3 \scriptsize$\pm4.8$ & 54.9 \scriptsize$\pm1.7$ & 39.8 \scriptsize$\pm2.3$ & 45.8 \\
ERM$^\dagger$ & 49.8 \scriptsize$\pm4.4$ & 42.1 \scriptsize$\pm1.4$ & 56.9 \scriptsize$\pm1.8$ & 35.7 \scriptsize$\pm3.9$ & 46.1 \\
VREx$^\dagger$ & 48.2 \scriptsize$\pm4.3$ & 41.7 \scriptsize$\pm1.3$ & 56.8 \scriptsize$\pm0.8$ & 38.7 \scriptsize$\pm3.1$ & 46.4 \\
RSC$^\dagger$ & 50.2 \scriptsize$\pm2.2$ & 39.2 \scriptsize$\pm1.4$ & 56.3 \scriptsize$\pm1.4$ & \textbf{40.8} \scriptsize$\pm0.6$ & 46.6 \\
DANN$^\dagger$ & 51.1 \scriptsize$\pm3.5$ & 40.6 \scriptsize$\pm0.6$ & 57.4 \scriptsize$\pm0.5$ & 37.7 \scriptsize$\pm1.8$ & 46.7 \\
IRM$^\dagger$ & 54.6 \scriptsize$\pm1.3$ & 39.8 \scriptsize$\pm1.9$ & 56.2 \scriptsize$\pm1.8$ & 39.6 \scriptsize$\pm0.8$ & 47.6 \\
CORAL$^\dagger$ & 51.6 \scriptsize$\pm2.4$ & 42.2 \scriptsize$\pm1.0$ & 57.0 \scriptsize$\pm1.0$ & 39.8 \scriptsize$\pm2.9$ & 47.7 \\
MLDG$^\dagger$ & 54.2 \scriptsize$\pm3.0$ & 44.3 \scriptsize$\pm1.1$ & 55.6 \scriptsize$\pm0.3$ & 36.9 \scriptsize$\pm2.2$ & 47.8 \\
I-Mixup & \textbf{59.6} \scriptsize$\pm2.0$ & 42.2 \scriptsize$\pm1.4$ & 55.9 \scriptsize$\pm0.8$ & 33.9 \scriptsize$\pm1.4$ & 47.9 \\
SagNet$^\dagger$ & 53.0 \scriptsize$\pm2.9$ & 43.0 \scriptsize$\pm2.5$ & 57.9 \scriptsize$\pm0.6$ & 40.4 \scriptsize$\pm1.3$ & 48.6 \\
ERM$^\ddagger$ & 54.3 \scriptsize$\pm0.4$ & 42.5 \scriptsize$\pm0.7$ & 55.6 \scriptsize$\pm0.3$ & 38.8 \scriptsize$\pm2.5$ & 47.8 \\
\midrule
Ours & 55.4 \scriptsize$\pm0.0$ & \textbf{44.9} \scriptsize$\pm1.1$ & \textbf{59.7} \scriptsize$\pm0.4$ & 39.9 \scriptsize$\pm0.2$ & \textbf{50.0} \\
\bottomrule
\end{tabular}
\end{table}

\subsection{DomainNet}
\begin{table}[H]
\centering
\small
\renewcommand{\arraystretch}{1.1}
\caption{\small\textbf{Out-of-domain accuracies (\%) on} \data{DomainNet}\textbf{.}}
\begin{tabular}{lllllll|c}
\toprule
\textbf{Algorithm} & \textbf{clip} & \textbf{info} & \textbf{paint} & \textbf{quick} & \textbf{real} & \textbf{sketch} & \textbf{Avg} \\
\midrule
MMD$^\dagger$ & 32.1 \scriptsize$\pm13.3$ & 11.0 \scriptsize$\pm4.6$ & 26.8 \scriptsize$\pm11.3$ & 8.7 \scriptsize$\pm2.1$ & 32.7 \scriptsize$\pm13.8$ & 28.9 \scriptsize$\pm11.9$ & 23.4 \\
GroupDRO$^\dagger$ & 47.2 \scriptsize$\pm0.5$ & 17.5 \scriptsize$\pm0.4$ & 33.8 \scriptsize$\pm0.5$ & 9.3 \scriptsize$\pm0.3$ & 51.6 \scriptsize$\pm0.4$ & 40.1 \scriptsize$\pm0.6$ & 33.3 \\
VREx$^\dagger$ & 47.3 \scriptsize$\pm3.5$ & 16.0 \scriptsize$\pm1.5$ & 35.8 \scriptsize$\pm4.6$ & 10.9 \scriptsize$\pm0.3$ & 49.6 \scriptsize$\pm4.9$ & 42.0 \scriptsize$\pm3.0$ & 33.6 \\
IRM$^\dagger$ & 48.5 \scriptsize$\pm2.8$ & 15.0 \scriptsize$\pm1.5$ & 38.3 \scriptsize$\pm4.3$ & 10.9 \scriptsize$\pm0.5$ & 48.2 \scriptsize$\pm5.2$ & 42.3 \scriptsize$\pm3.1$ & 33.9 \\
Mixstyle$^\ddagger$ & 51.9 \scriptsize$\pm0.4$ & 13.3 \scriptsize$\pm0.2$ & 37.0 \scriptsize$\pm0.5$ & 12.3 \scriptsize$\pm0.1$ & 46.1 \scriptsize$\pm0.3$ & 43.4 \scriptsize$\pm0.4$ & 34.0 \\
ARM$^\dagger$ & 49.7 \scriptsize$\pm0.3$ & 16.3 \scriptsize$\pm0.5$ & 40.9 \scriptsize$\pm1.1$ & 9.4 \scriptsize$\pm0.1$ & 53.4 \scriptsize$\pm0.4$ & 43.5 \scriptsize$\pm0.4$ & 35.5 \\
CDANN$^\dagger$ & 54.6 \scriptsize$\pm0.4$ & 17.3 \scriptsize$\pm0.1$ & 43.7 \scriptsize$\pm0.9$ & 12.1 \scriptsize$\pm0.7$ & 56.2 \scriptsize$\pm0.4$ & 45.9 \scriptsize$\pm0.5$ & 38.3 \\
DANN$^\dagger$ & 53.1 \scriptsize$\pm0.2$ & 18.3 \scriptsize$\pm0.1$ & 44.2 \scriptsize$\pm0.7$ & 11.8 \scriptsize$\pm0.1$ & 55.5 \scriptsize$\pm0.4$ & 46.8 \scriptsize$\pm0.6$ & 38.3 \\
RSC$^\dagger$ & 55.0 \scriptsize$\pm1.2$ & 18.3 \scriptsize$\pm0.5$ & 44.4 \scriptsize$\pm0.6$ & 12.2 \scriptsize$\pm0.2$ & 55.7 \scriptsize$\pm0.7$ & 47.8 \scriptsize$\pm0.9$ & 38.9 \\
I-Mixup & 55.7 \scriptsize$\pm0.3$ & 18.5 \scriptsize$\pm0.5$ & 44.3 \scriptsize$\pm0.5$ & 12.5 \scriptsize$\pm0.4$ & 55.8 \scriptsize$\pm0.3$ & 48.2 \scriptsize$\pm0.5$ & 39.2 \\
SagNet$^\dagger$ & 57.7 \scriptsize$\pm0.3$ & 19.0 \scriptsize$\pm0.2$ & 45.3 \scriptsize$\pm0.3$ & 12.7 \scriptsize$\pm0.5$ & 58.1 \scriptsize$\pm0.5$ & 48.8 \scriptsize$\pm0.2$ & 40.3 \\
MTL$^\dagger$ & 57.9 \scriptsize$\pm0.5$ & 18.5 \scriptsize$\pm0.4$ & 46.0 \scriptsize$\pm0.1$ & 12.5 \scriptsize$\pm0.1$ & 59.5 \scriptsize$\pm0.3$ & 49.2 \scriptsize$\pm0.1$ & 40.6 \\
ERM$^\dagger$ & 58.1 \scriptsize$\pm0.3$ & 18.8 \scriptsize$\pm0.3$ & 46.7 \scriptsize$\pm0.3$ & 12.2 \scriptsize$\pm0.4$ & 59.6 \scriptsize$\pm0.1$ & 49.8 \scriptsize$\pm0.4$ & 40.9 \\
MLDG$^\dagger$ & 59.1 \scriptsize$\pm0.2$ & 19.1 \scriptsize$\pm0.3$ & 45.8 \scriptsize$\pm0.7$ & 13.4 \scriptsize$\pm0.3$ & 59.6 \scriptsize$\pm0.2$ & 50.2 \scriptsize$\pm0.4$ & 41.2 \\
CORAL$^\dagger$ & 59.2 \scriptsize$\pm0.1$ & 19.7 \scriptsize$\pm0.2$ & 46.6 \scriptsize$\pm0.3$ & 13.4 \scriptsize$\pm0.4$ & 59.8 \scriptsize$\pm0.2$ & 50.1 \scriptsize$\pm0.6$ & 41.5 \\
MetaReg & 59.8 & \textbf{25.6} & 50.2 & 11.5 & 64.6 & 50.1 & 43.6 \\
DMG & 65.2 & 22.2 & 50.0 & 15.7 & 59.6 & 49.0 & 43.6 \\
ERM$^\ddagger$ & 63.0 \scriptsize$\pm0.2$ & 21.2 \scriptsize$\pm0.2$ & 50.1 \scriptsize$\pm0.4$ & 13.9 \scriptsize$\pm0.5$ & 63.7 \scriptsize$\pm0.2$ & 52.0 \scriptsize$\pm0.5$ & 44.0 \\
\midrule
Ours & \textbf{66.0} \scriptsize$\pm0.1$ & 22.4 \scriptsize$\pm0.3$ & \textbf{53.5} \scriptsize$\pm0.1$ & \textbf{16.1} \scriptsize$\pm0.2$ & \textbf{65.8} \scriptsize$\pm0.4$ & \textbf{55.5} \scriptsize$\pm0.3$ & \textbf{46.5} \\
\bottomrule
\end{tabular}
\end{table}

\section{Assets}

In this section, we discuss about licenses, copyrights, and ethical issues of our assets, such as code and datasets. 

\subsection{Code}

Our work is built upon DomainBed~\cite{gulrajani2020domainbed}\footnote{\url{https://github.com/facebookresearch/DomainBed}}, which is released under the MIT license.

\subsection{Datasets}

While we use public datasets only, we track how the datasets were built to discuss licenses, copyrights, and potential ethical issues.
For \data{DomainNet}~\cite{peng2019domainnet} and \data{OfficeHome}~\cite{venkateswara2017officehome}, we use the datasets for non-profit academic research only following their fair use notice. \data{TerraIncognita}~\cite{beery2018terraincognita} is a subset of Caltech Camera Traps (CCT) dataset, distributed under the Community Data License Agreement (CDLA) license.
\data{PACS}~\cite{li2017pacs} and \data{VLCS}~\cite{fang2013vlcs} datasets have images collected from the web and we could not find any statements about licenses, copyrights, or whether consent was obtained.
%Further, we could not find whether they get authorization of the assets or not, even though both datasets have person class.
Considering that both datasets contain person class and images of people, there may be potential ethical issues.

\section{Reproducibility}

To provide details of our algorithm and guarantee reproducibility, we provide the source code\footnote{\url{https://github.com/khanrc/swad}} publicly. The code also specifies detailed environments, dependencies, how to download datasets, and instructions to reproduce the main results (Table 1 and 2 in the main text).

\subsection{Infrastructures}

Every experiment is conducted on a single NVIDIA Tesla P40 or V100, Python 3.8.6, PyTorch 1.7.0, Torchvision 0.8.1, and CUDA 9.2.

\subsection{Runtime Analysis}

The total runtime varies depending on datasets and the moment detected to overfit. It takes about 4 hours for \data{PACS} and \data{VLCS}, 8 hours for \data{OfficeHome}, 8.5 hours for \data{TerraIncognita}, and 56 hours for \data{DomainNet} on average, when using a single NVIDIA Tesla P40 GPU. Each experiment includes the leave-one-out cross-validations for all domains in each dataset.
%That is, each experiment has the maximum number of iterations $T \times D$, where $T$ and $D$ indicates the number of iterations for each target domain and the number of domains, respectively.

\subsection{Complexity Analysis}

The only additional time overhead incurs from stochastic weights selection, which requires further evaluations.
To analyze the overhead, let the forward time $t_f$, backward time $t_b$, training and validation split ratio $r=|X^{train}|/|X^{valid}|$, total in-domain samples $n$, and evaluation frequency $v$ that indicates how many evaluations are conducted for each epoch. For conciseness, we assume $t=t_f=t_b$ and do not consider early stopping.

For one epoch, training time is $2tnr/(r+1)$, and evaluation time is $vtn/(r+1)$. The total runtime for one epoch is $tn(2r+v)/(r+1)$. Final overhead ratio is $(2r+v)/(2r+v_b)$ where $v_b$ is the evaluation frequency of a baseline. In our main experiments, we use $r=4$. Compared to the default parameters of DomainBed~\cite{gulrajani2020domainbed}, we use $v=2v_b$ for \data{DomainNet}, $v=6v_b$ for \data{VLCS}, and $v=3v_b$ for the others. Then, the total runtime of our algorithm takes from 1.07 (\data{PACS}) to 1.27 (\data{DomainNet)} times more than the ERM baseline. In practice, it can be improved by conducting approximated evaluations using sub-sampled validation set.
%stochastic evaluation approximation via validation set sampling.

In terms of memory complexity, our method does not require additional GPU memory. Instead, we leverage CPU memory to minimize training time overhead, which takes up to $\max (N, M)$ times more than the baseline.

\end{document}